\documentclass[journal]{IEEEtran}
\newtheorem{Semi-supervised ABOF}{Definition}
\usepackage{algorithm}
\usepackage{algorithmic}
\usepackage{graphicx}
\usepackage{epstopdf}
\usepackage{float}
\usepackage{subfigure}
\usepackage{multicol}
\usepackage{multirow}
\usepackage{wrapfig}
\usepackage{url}
\usepackage{indentfirst}
\usepackage{amssymb}
\usepackage{amsmath}
\usepackage{delarray}
\usepackage{tabularx}
\usepackage{mathrsfs}
\usepackage{amsmath}
\usepackage{amsthm}
\usepackage{bm}
\usepackage{booktabs}
\usepackage{hyperref}
\usepackage{tikz}
\usepackage{textcomp}
\usepackage[T1]{fontenc} % optional
\usepackage[english]{babel}
\hypersetup{
	colorlinks,
	linkcolor=red,
	anchorcolor=blue,
	citecolor=green
}
\usepackage{caption}
\renewcommand{\footnoterule}{  
	\kern -3pt
	\hrule  width 0.15\textwidth height 1pt
	\kern 2pt
}
\usepackage{fancyhdr}
\usepackage{caption}
\renewcommand{\footnoterule}{  
	\kern -3pt
	\hrule  width 0.15\textwidth height 1pt
	\kern 2pt
}
\makeatletter
\def\hlinew#1{
	\noalign{\ifnum0=`}\fi\hrule \@height #1 \futurelet
	\reserved@a\@xhline}
\makeatother

\begin{document}

%\author{\IEEEauthorblockN{***,	
%	} \\
%	\IEEEauthorblockA
%	{Department of Computing, Hong Kong Polytechnic University, Hong Kong, China}
%}

\author{\IEEEauthorblockN{Sitong Mao,
		Jiaxin Chen,
		Xiao Shen,
		Fu-lai Chung~\IEEEmembership{Member,~IEEE}
	   } \\
	\IEEEauthorblockA
	{Department of Computing, The Hong Kong Polytechnic University, Hong Kong, China}
}

\title{Deep Adversarial Domain Adaptation Based on Multi-layer Joint Kernelized Distance}
\maketitle

\begin{abstract}
Domain adaptation refers to the learning scenario that a model learned from the source data is applied on the target data which have the same categories but different distribution. While it has been widely applied, the distribution discrepancy between source data and target data can substantially affect the adaptation performance. The problem has been recently addressed by employing adversarial learning and distinctive adaptation performance has been reported. In this paper, a deep adversarial domain adaptation model based on a multi-layer joint kernelized distance metric is proposed. By utilizing the abstract features extracted from deep networks, the multi-layer joint kernelized distance (MJKD) between the $j$th target data predicted as the $m$th category and all the source data of the $m'$th category is computed. Base on MJKD, a class-balanced selection strategy is utilized in each category to select target data that are most likely to be classified correctly and treat them as labeled data using their pseudo labels. Then an adversarial architecture is used to draw the newly generated labeled training data and the remaining target data close to each other. In this way, the target data itself provide valuable information to enhance the domain adaptation. An analysis of the proposed method is also given and the experimental results demonstrate that the proposed method can achieve a better performance than a number of state-of-the-art methods.	
\end{abstract}


\section{Introduction}
The importance of domain adaptation has been explored in a series of applications, e.g., information retrieval as in cross domain recommendation~\cite{farseev2017cross} and cross network influence maximization~\cite{Shen:2017:LCI:3077136.3080646}~\cite{shen2019cross}, computational biology~\cite{liu2008evigan}, natural language processing~\cite{mcclosky2006reranking}~\cite{daume2009frustratingly}, and computer vision~\cite{saenko2010adapting}~\cite{long2015learning}~\cite{long2016deep}~\cite{gong2012geodesic}~\cite{weston2012deep}~\cite{pan2011domain}~\cite{tzeng2014deep}. Domain adaptation refers to the learning scenario that adapts a model  to the unlabeled or a few labeled target data by borrowing information from labeled source data from different but related domains. Under the setting of domain adaptation, despite consisting of the same categories, the source data and the target data are typically distributed differently which is referred as domain shift. For instance, in the scenario of visual domain adaptation, the distribution can be substantially affected by angle transformation, illumination, or occlusion. However, machine learning models that work well rely on the assumption that the training set and the test set are drawn from the same feature space and the same distribution, which is not valid in domain adaptation settings because of the domain shift. Hence, to obtain favorable performance on the target dataset whose distribution is different from the source data, one may need to recollect labeled training data and then retrain the models on the extended dataset. However, it is prohibitively expensive or even impossible to collect more training data with label information. Hence, it is important to develop domain adaptation methods that can borrow prior knowledge to compensate for the unavailable or insufficient labels of the target data.

To combat the performance degradation in target domain arising from domain shift, previous works have explored approaches in various directions. The common ones are feature augmentation~\cite{daume2009frustratingly}~\cite{duan2012learning}~\cite{gopalan2014unsupervised}, feature transformation~\cite{saenko2010adapting}~\cite{kulis2011you}, and domain re-sampling~\cite{gong2013connecting}. Although these methods have made prominent progress, their shallow architectures prevent them from achieving more desirable performance. Recently, deep neural networks have been proved having strong ability of learning transferable features~\cite{donahue2014decaf}, and this depicts the potential of empowering domain adaptation with deep learning.

Parameters in deep neural networks eventually transit from general to task-specific as the layer goes higher/deeper. The transferability of each particular layer in a deep neural network has been quantified in~\cite{yosinski2014transferable} which shows that parameters from lower-layers are applicable to both source and target tasks, while the transferability declines in higher/deeper layers. Therefore, the network pre-trained on source dataset is not likely to be discriminative enough when it is applied to the target dataset directly. Inspired by such characteristic of deep neural networks, some hierarchical approaches have been proposed to boost the generalization performance by reducing the domain discrepancy in higher layers~\cite{long2015learning}~\cite{long2016deep}~\cite{chen2012marginalized}~\cite{Chopra2013DLIDDL}, or utilizing self-training method to map the source data and the target data closer iteratively~\cite{mao2018deep}~\cite{Shen:2017:LCI:3077136.3080646}. These deep learning based methods can significantly outperform those approaches using shallow architectures.

Most recently, embedding adversarial learning architecture in the deep neural networks has achieved impressive performance by mapping the targe data and the source data closer to each other in a two-player manner~\cite{ganin2016domain}. In this kind of approaches, a discriminator is trained to distinguish source data and target data by taking features extracted from a deep neural network as inputs, while the deep neural network is tuned to confuse the discriminator. Instead of artificially defining a distribution discrepancy metric, the adversarial architecture leverages the power of deep network structure, namely the discriminator, to help mapping the source and target data close in an effective way. In addition, its performance can be enhanced by setting the joint features of the predicted probabilities and the extracted deep features as the input of the discriminator~\cite{long2018conditional}~\cite{pei2018multi}. 

In this paper, a target data select-and-adapt strategy is leveraged to further enhance the adversarial learning performance. In the previous conditional adversarial learning architecture, the discriminator tries to distinguish the joint features (tensor-product $\mathbf{f}\otimes\mathbf{p}$ of the deep features $\mathbf{f}$ and the predicted results $\mathbf{\tilde{y}}$) of the source data and the target data. According to the definition of the joint distribution $p(\mathbf{f},\mathbf{\tilde{y}})$, the predicted results may influence the adaptation performance significantly. If the class information given by $\mathbf{\tilde{y}}$ is correct, then $\mathbf{f}$ will be mapped close to its ground truth category. Thus, a method that can improve the capability of the source classifier on classifying target data will improve the performance of adversarial learning. The proposed method contains two steps: \textbf{1)} First, a model ``$M_0$'' is finetuned on the labeled source data using classic deep neural network architecture (e.g., AlexNet, etc) pretrained on some large datasets (e.g., ImageNet). Then, a metric named ``MJKD'' is proposed to measure how likely a target data is correctly classified using the deep features extracted by ``$M_0$''. Here a class-balanced selection strategy is leveraged to avoid poor performance~\cite{buda2018systematic}~\cite{he2009learning}, which means that the number of the target selected from each category is the same. \textbf{2)} After integrating the target data selected by ``MJKD'' to the source data with their pseudo labels, the integrated dataset is used as the labeled training data. In this way, the selected target data can provide valuable information itself to help to enhance the diversity of the features of each category, which can help to map more test data close to their ground truth categories in the adversarial learning process. As far as we know, the proposed method is the first one to use the selected target data in a supervised manner in the adversarial learning process. It demonstrates that the target data should get more attention in adversarial domain adaptation learning, which is critical to advance the technology of unsupervised domain adaptation. It contrasts with the previous approaches to map the target data close to only the source data adversarially. The further value of this work is that it can show the value of information provided by the target data, which can provide a new direction of thinking about the domain adaptation problem. The main contributions of this paper can be summarized as follows: 
\begin{itemize}
	\item A deep layered joint distance metric that can be effectively used to rank the target data for proper inclusion by the training data is proposed.    
	\item An adversarial network model empowered by updating the labeled training set with class-balanced target data is developed to attain state-of-the-art adaptation performance. 
	\item Instead of only being supervised by the source data, the proposed method leverages the information provided by target data itself.
	\item An analysis of the proposed method is provided to deepen the understanding of deep adversarial domain adaptation. 
\end{itemize}

In the next section, we review some representative and recent works related to domain adaptation. After that, our proposed approach and some preliminaries are introduced in detail and then an analysis of the proposed method is given. We finally evaluate our approach and show the comparative results before concluding the paper.

\section{Related Work}
Previous domain adaptation approaches based on shallow architectures can be roughly grouped into the following types: 1) In~\cite{daume2009frustratingly}, Daum\'{e} III proposed a feature augmentation-based method that maps features to an augmented space by simply copying the original feature vectors to a domain-specific portion and a domain-generic portion respectively. As an extension of this general idea, some manifold-based~\cite{gopalan2014unsupervised}~\cite{gopalan2011domain}~\cite{shrivastava2014unsupervised}~\cite{zheng2012grassmann} and kernel-based~\cite{gong2012geodesic}~\cite{gong2014learning} data augmentation approaches have been proposed. 2) Another direction is to learn a transformation under which the source and target distributions can be represented closer~\cite{saenko2010adapting}~\cite{blitzer2011domain}. 3) One additional approach is to make labeled source instances that are most similar to target data carry more weights~\cite{liu2008evigan}~\cite{huang2007correcting}~\cite{mansour2009domain}~\cite{sugiyama2007covariate}~\cite{gong2013connecting}. Here the similarities can be estimated by various methods such as the kernel mean matching (KMM) procedure~\cite{huang2007correcting}. Despite the appreciable improvement made by these methods, they are still limited by the shallow architecture which cannot effectively learn representative features and hence  their domain-specific variability is suppressed.

Deep neural networks have gained much attention in many applications recently for its distinctive power in learning more robust features that are invariant to the differences between tasks~\cite{bengio2013representation}, and its lower layer features can be generalized to almost any tasks directly~\cite{yosinski2014transferable}. However, the distribution discrepancy between domains cannot be effectively minimized in the higher layers. This inspires exploration of domain adaptation approaches based on deep neural networks. Some previous works transfer deep learned features to the target networks by reusing parameters of mid-layers pretrained on the source data~\cite{yang2017learning}~\cite{qian2016learning}~\cite{oquab2014learning}. Another popular direction is to minimize the distribution discrepancy of higher layers by integrating a manually defined statistical metric into the loss function. Some works based on this kind of approach mainly consider the distribution changes of the features $P(X)$, i.e., marginal distribution. Long \textit{et al.}~\cite{long2015learning} proposed a Deep Adaptation Network (DAN) which incorporates multi-kernel maximum mean discrepancy (MK-MMD) of the highest few layers as a regularizer in the CNN loss function. Also in~\cite{tzeng2015simultaneous}, the proposed CNN architecture combines domain confusion and softmax cross-entropy losses, which only correct the shifts in the marginal distributions. In this approach, the MK-MMD is computed without considering the labels. However, it is not clear that under what conditions the approximately same marginal distributions ($P^S(\mathcal{T}(X))\approx P^T(\mathcal{T}(X))$) can imply similar conditional distributions ($P^S(Y|\mathcal{T}(X))\approx P^T(Y|\mathcal{T}(X))$).

To address the shortcoming of merely relying on correcting the marginal domain shifts, Gong et al.~\cite{gong2016domain} aim to find conditional transferable components that are invariant across different domains. In~\cite{long2016deep}, Long et al. have recently proposed a ``joint distribution discrepancy'' (JDD) metric. By bounding the JDD together with the cross-entropy loss of the source data, the conditional distribution discrepancy can be reduced. To compute JDD, the distribution over the class labels produced by the pre-trained CNN model is used. 

Most recently, adversarial learning has been adopted to deal with the domain adaptation problem and remarkable performance has been reported. As demonstrated in~\cite{long2018conditional}, the probability vectors predicted by the deep neural networks can provide useful adaptation information, so the tensor products of the deep features and the probability vectors are sent to the discriminator as inputs to bound the conditional distribution discrepancy. The discriminator acts to distinguish which domain the data comes from, while the deep neural networks try to learn features that can confuse the discriminator. Thus, a deep model that iteratively maps the target and source data closer can be obtained. In~\cite{pei2018multi}, the tensor product is further split into vectors, each of which represents the component of the data in a certain category. This method is essentially similar to that of~\cite{long2018conditional}. Although integrating the probability vectors in the adversarial networks can improve the adaptation, misclassified target data is still an obstacle for further improvement of the adversarial learning process because when the probability vectors are combined, the target data tend to be mapped closer to the category it is classified to, i.e., the class with larger probability. 

The deep domain adaptation model proposed in this paper attempts to alleviate the mis-prediction problem of target labels arising from domain shifts in the supervised training part of the adversarial architecture. A multi-layer joint kernelized distance (MJKD) is proposed to identify the target samples that are most likely to be correctly classified. The new metric effectively selects a same number of correctly predicted target data for each category through which a more accurate prediction can be obtained. Thus, the selected target data can be integrated to carry out the subsequent adversarial learning process effectively. This can provide a direction for the future research of domain adaptation which will exploit the potential value of the target data itself.

\begin{figure*}
	\centering
	\includegraphics[width=0.9\linewidth,height=0.5\linewidth]{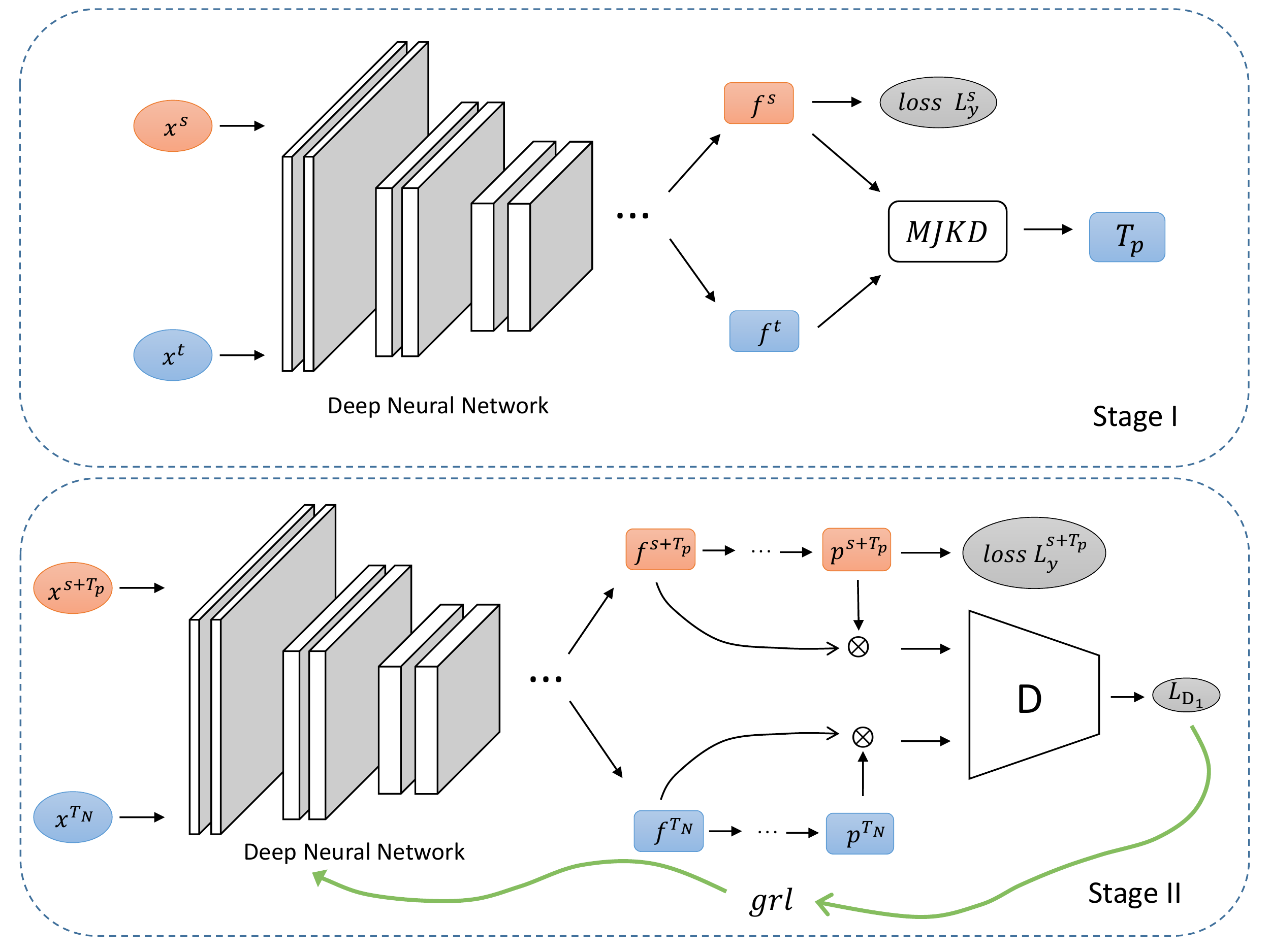}
	\caption{An illustration of the proposed deep adversarial domain adaptation. In the first stage, $x^s$ denotes the source data and $x^t$ denotes the target data. $f^s$ and $f^t$ are the deep features of source data and target data extracted by the deep neural network respectively. $T_p$ is the the set of selected target data which are more likely to be correctly classified. In the second stage, the architecture proposed in CDAN~\cite{long2018conditional} is used. $x^{s+T_p}$ is the updated labeled training data and $x^{T_N}$ is the updated unlabeled data. $f^{s+T_p}$ and $f^{T_N}$ are the deep features. $p^{s+T_p}$ and $p^{T_N}$ are the probabilities distributed over each category predicted by the deep neural network. The symbol \textcircled{$\times$} denotes the tensor product of the deep features and the probability vectors. $D$ is the discriminator. $grl$ is the gradient reversal layer~\cite{ganin2015unsupervised}.}
	\label{fig:emd} 
\end{figure*}

\section{Methodology}
In the standard setting of unsupervised domain adaptation, labeled source dataset $\mathcal{S}=\{X^{S},Y^{S}\}$ and un-labeled target data $\mathcal{T}=\{X^{T}\}$ are given. Here $X^S=\{x^s_i\},i=1,2,\dots,n_s$, where $n_s$ is the number of samples in the source domain. Similarly, $X^T=\{x^t_i\},i=1,2,\dots,n_t$. And $Y^S=\{y^s_i\},i=1,2,\dots,n_s,y^s_i\in\{1,2,\dots,c\}$ denotes the label of the source data, where $c$ is the number of categories. 

In the following, the MJKD is firstly introduced and then its use to carry out target data selection is described. The basic architecture of the deep adversarial domain adaptation model is subsequently presented. Finally, the integrated procedure is given.

\subsection{Preliminary: Deep Adversarial Domain Adaptation Architecture}
Hereinafter, $D$ is used to denote the Discriminator and $G$ is used to denote the Generator. As shown in Figure~\ref{fig:emd}, a deep neural network is employed for $G$, and the deep features generated by it will be sent to the discriminator as inputs. The goal of the discriminator is to distinguish the source data from the target data, while the deep neural network of $G$ tries to extract features that can confuse the discriminator, i.e., making it hard to tell which domain the input data belong to. 

$G$ is conditioned on two error terms: the source classifier loss $\mathcal{L}_y$ and the adversarial training loss. The minimization of the first error term can be described by eq.~\ref{eq:source_classify}. 
\begin{equation}
\min\limits_G E(G) = \mathbb{E}_{(x^s, y^s)\sim p_{s}(x^s,y^s)}[\mathcal{L}(G(x^s),y^s)]
\label{eq:source_classify}
\end{equation}
On the other hand, the adversarial training process is given by eq.~\ref{eq:adversarial}. Here, $D(\cdot)$ denotes the probability of a data comes from the source domain. $G(\cdot)$ is the deep features extracted by the deep neural networks. $G(x^s), x^s\sim p_{s}(x^s)$ denotes the deep features of source data generated by the deep neural network, which is a convolutional neural network (CNN) in this work and $G(x^t), x^t\sim p_{t}(x^t)$ are the deep features generated for the target data. As expressed in eq.~\ref{eq:adversarial}, the training process of the adversarial domain adaptation model is a two-player game between the discriminator $D$ and the generator $G$. The discriminator tries to distinguish the source data from the target one by maximizing the expectation of $D(x^s)$ and minimize the expectation of $D(x^t)$, while the generator $G$ is trained to confuse the discriminator by minimizing the expectation of $D(G(x^s))$ and maximizing the expectation of $D(G(x^t))$, i.e., 
\begin{equation}
\begin{aligned}
\min\limits_G \max\limits_D V(G,D) & = \mathbb{E}_{x^s\sim p_{s}(x^s)}[log(D(G(x^s)))] \\
& +\mathbb{E}_{x^t\sim p_{t}(x^t)}[log(1-D(G(x^t)))].
\end{aligned}
\label{eq:adversarial}
\end{equation}

As demonstrated in~\cite{long2018conditional}, the probability vectors $P$ predicted by the network can provide significant information for adaptation. Here $P = [p_1,\; p_2,\; \dots,\; p_c ]$ where $p_m$ denotes the probability of this data belong to the $m$th category. Thus, following~\cite{long2018conditional}, the features sent to the discriminator are the tensor products of the deep features from CNN $G^{f}(\cdot)$ and the probability vectors associated with the category information $G^{p}(\cdot)$ (cf. the \textcircled{$\times$} in Figure~\ref{fig:emd}). Then the error function for adversarial learning becomes~\cite{long2018conditional}:
\begin{equation}
\begin{aligned}
\min\limits_G \max\limits_D & V(G,D) = \\
& \mathbb{E}_{x^{s}\sim p_{s}(x^{s})}[logD(G^{f}(x^{s}) \otimes G^{p}(x^{s}))] \\
+ & \mathbb{E}_{x^{t}\sim p_{t}(x^t)}[log(1-D(G^{f}(x^{t}) \otimes G^{p}(x^t)))] \\
= & \mathbb{E}_{(f,\tilde{y})\sim G_{s}}[logD(f \otimes \tilde{y})] \\
+ & \mathbb{E}_{(f,\tilde{y})\sim G_{t}}[log(1-D(f \otimes \tilde{y})].
\end{aligned}
\label{eq:crossproduct_ad}
\end{equation}
Here $p_{s}(x^s)$ denotes the distribution of the source data, and $p_{t}(x^t)$ denotes the distribution of the target data. $G_s$ and $G_t$ denote the joint distribution of features and probabilities of source data and target data respectively.

In the next subsection, the multi-layer joint kernelized distance is explained in detail.

\subsection{MJKD Enhanced Domain Adaptation}
Given the source and target deep features ${x^{s\ell}_{im}},{x^{t\ell}_{jm}}$ extracted from different layers $\ell=\ell_1,\dots,\ell_2$ of a deep neural network pretrained on the source data, with $m=1,2,\dots,c$ denoting the ground truth label of the source data and the predicted label of the target data, $x^{s\ell}_{im}$ denotes the deep feature of the $i$th source data of the $m$th category extracted from layer $\ell$, and ${x^{t\ell}_{jm}}$ denotes the deep feature of the $j$th target data of the $m$th category extracted from layer $\ell$. 
\subsubsection{Multi-layer Joint Kernelized Distance}
Many previous works measure the discrepancy between two distributions with respect to the deep features in the Reproducing Kernel Hilbert Space $\mathcal{H}_k$, where the inner product is defined as $<\phi(x),\phi(x')>=k(x,x')$ and $k(\cdot,\cdot)$ is the kernel function. Gaussian kernel $e^{-\frac{\parallel x_i - x_j\parallel ^2}{\gamma}}$ has been frequently used and was also employed in our experiments. In order to choose the parameter $\gamma$ of the kernel function automatically, a multi-kernel approach was proposed in~\cite{long2015learning}. For example, if five values of $\gamma$ are used in the kernel function, namely $\gamma_0 = avg \sum_{i,j}\parallel x^s_{im}-x^t_{jm}\parallel^2$, $\gamma_1 = \gamma_0 \times 2$, $\gamma_2 = \gamma_0 \times 4$, $\gamma_3 = \gamma_0 / 2$, $\gamma_4 = \gamma_0 / 4$, then the final result is the average of these kernel functions using different value of $\gamma$. In the proposed method, instead of calculating the distance between all the target data and the source data, we consider each target data as an individual distribution.
Consequently, the distance between a target data $x^{t}_{jm}$ and a source category $m'=1,\dots,c$ is defined as:
\begin{equation}
\begin{aligned}
d^2_k & (x^{t}_{jm},x^s_{m'}) = \\
& \lVert \mathbb{E}[\otimes_{\ell=\ell_1}^{\ell_2}\phi^{\ell}(x^{t\ell}_{jm})]-\mathbb{E}[\otimes_{\ell=\ell_1}^{\ell_2}\phi^{\ell}(x^{s\ell}_{m'})]\rVert^2_{\mathcal{H}^{\ell}_k}.
\end{aligned}
\label{eq:1}
\end{equation}
Here,  $\otimes_{\ell=\ell_1}^{\ell_2}\phi^{\ell}(x^\ell)=\phi^{\ell_1}(x^{\ell_1})\otimes\dots\otimes\phi^{\ell_2}(x^{\ell_2})$ is the joint embedding feature maps of different layers. The inner product in this joint embedding Reproducing Hilbert Space is defined as 
\begin{equation}
<\otimes_{\ell=\ell_1}^{\ell_2}\phi^{\ell}(x^\ell),\otimes_{\ell=\ell_1}^{\ell_2}\phi^{\ell}(x'^\ell)>=\prod\limits_{\ell=\ell_1}^{\ell_2}k^\ell (x^\ell,x'^\ell).
\label{eq:2}
\end{equation}
Thus, a multi-layer joint kernelized distance (MJKD) between $x^{t}_{jm}$ and source category $m'$ is proposed as
\begin{equation}
\begin{aligned}
& d^2_k (x^{t}_{jm},x^s_{m'}) & =\quad & \prod\limits_{\ell=\ell_1}^{\ell_2}k^\ell (x^{t\ell}_{jm},x^{t\ell}_{jm})+ \\
& & & \frac{1}{(n^s_{m'})^2}\sum\limits^{n^s_{m'}}_{i,k=1}\prod\limits_{\ell=\ell_1}^{\ell_2}k^\ell (x^{s\ell}_{im'},x^{s\ell}_{km'})- \\
& & & \frac{2}{n^s_{m'}}\sum\limits^{n^s_{m'}}_{i=1}\prod\limits_{\ell=\ell_1}^{\ell_2}k^\ell(x^{t\ell}_{jm},x^{s\ell}_{im'}). %\\
%& & =\quad & 2-\frac{2}{n^s_{m'}}\sum\limits^{n^s_{m'}}_{i=1}\prod\limits_{\ell=\ell_1}^{\ell_2}k^\ell(x^{t\ell}_{jm},x^{s\ell}_{im'}).
\end{aligned}
\label{eq:MJKD}
\end{equation}

\subsubsection{Class-balanced Data Selection}
It has been shown that whether the training data is class-balanced is significant for the performance of a model. Models trained on unbalanced data may perform poorly for weakly represented categories which contain relatively few examples~\cite{buda2018systematic}~\cite{he2009learning}. Thus, based on the metric MJKD, we further propose to select the same number of target data in each category according to their pseudo labels and then incorporate them to the corresponding source category. 

Given the target data predicted to belong to the $m$th category ${x^{t}_{jm}}, j=1,2,\dots,n^t_m$, the MJKD between each target data $x^t_{jm}$ and the $m'$ source category $d^2_k (x^{t}_{jm},x^s_{m'})$ can be computed according to eq.~\ref{eq:MJKD}. Then, the relative distance of $x^{t}_{jm}$ is defined as
\begin{equation}
R(x^{t}_{jm})=\sum\limits_{m'=1}^{c}\frac{d^2_k (x^{t}_{jm},x^s_{m})}{d^2_k (x^{t}_{jm},x^s_{m'})}.
\label{eq:4}
\end{equation}
Intuitively, the correctly predicted target data $x^{t}_{jm}$ will be close to $x^s_{m}$ and far from $x^s_{m'},m\neq m'$. So the smaller $R(x^{t}_{jm})$ is, the more likely $x^{t}_{jm}$ belongs to the $m$th category. To carry out the class-balanced selection, we select top $k$ target data that are most likely to be correctly predicted in each category, and add them to the supervised training set with their pseudo-labels. Here, $k$ could be selected according to the amount of data in the target set. For larger target set, we could choose larger $k$. 
%In this paper, the value $k$ is chosen according to the following rule:
%\begin{equation}
%\left\{
%\begin{array}{lr}
%n_p = \dfrac{n^t}{4} \\
%k = \dfrac{n_p}{c}
%\end{array}
%\right.
%\label{eq:value_k}
%\end{equation}
%Here $n_p$ denotes the total number of target data added to the supervised training set. According to eq.~\ref{eq:value_k}, $n_p$ is set as a quarter of the total number of target data. Then, $k$ is chosen by making these $n_p$ data equally distributed in each category. 
This paper simply proposes an one-off select-and-adapt procedure. In our experiments, the total number of selected target data $c \times k$ is set to be a quarter of the amount of all the target data and evenly adapt them to the $c$ source categories. A sensitivity analysis of the selection percentage is also given. 

With the selected target data integrated with the source domain, an adversarial learning process can be straightforwardly implemented. In the next subsection, the overall training process integrating MKJD and the adversarial domain adaptation is demonstrated.

\subsection{Overall Training Process}
%The selected target data will have the same domain labels as the source data. The adversarial learning process is shown as below:
%\begin{equation}
%\begin{aligned}
%	\min\limits_G \max\limits_D\mathcal{L} & = \mathbb{E}_{x\sim p_{data}(x^s)}[log(D(G(x)))] \\
%	& +\mathbb{E}_{z\sim p_{data}(x^t)}[log(1-D(G(z)))].
%\end{aligned}
%\label{eq:integrate_ad}
%\end{equation}

The proposed deep adversarial domain adaptation model based on MJKD has been described in Algorithm~\ref{algo} and illustrated in Figure.~\ref{fig:emd}. As indicated, by fine-tuning on a deep model (e.g., AlexNet~\cite{krizhevsky2012imagenet}) trained on the ImageNet using the labeled source data, a pre-trained initial network model $M_0$ can be obtained. The pre-trained model $M_0$ is then used to classify all the unlabeled data of the target domain. The predicted target label, together with the labels of source data, the source features $x^{s\ell}$ and the target features $x^{t\ell}$ extracted by the deep neural network are then used to compute the MJKD. Then by computing the relative distance using MJKD, the correctly predicted ranking for each individual category (label) can be obtained. Based on the ranking information, it can be decided that which target samples are deemed as correctly labeled data $\mathcal{T}^p$ using class-balanced strategy. Then, $\mathcal{T}^p$ can be integrated to the source samples with their pseudo-labels to form the updated labeled training set. This updated labeled training set and the remaining unlabeled target data are then used to train the adversarial network whose architecture is proposed in CDAN~\cite{long2018conditional}. To train the discriminator of the adversarial network, the domain labels of the selected target data are set the same as that of the source data. At the same time, the selected auxiliary target data are removed from the unlabeled target sample set. The remaining target data are then assigned different domain labels from that of the source domain in the following adversarial training. Upon finishing the adversarial training process in Step 9, the classification accuracy for all the target data are produced.
{\renewcommand\baselinestretch{0.8}\selectfont
	\renewcommand{\algorithmicrequire}{\textbf{Input:}}
	\renewcommand{\algorithmicensure}{\textbf{Procedure:}}
	\begin{algorithm}
		\caption{Deep Adversarial Domain Adaptation Based on MJKD}
		\begin{algorithmic}[1]
			\REQUIRE
			\STATE Source dataset $\mathcal{S}=\{X^{\mathcal{S}},Y^{\mathcal{S}}\}$ \\		
			\STATE Target dataset $\mathcal{T}=\{X^{\mathcal{T}}\}$ \\
			\STATE Given pre-trained model $M$
			\ENSURE ~~\\
			\STATE Get the initial model $\mathcal{M}_0$ by fine-tuning from $M$ on $\mathcal{S}$; \\
			\STATE Apply $\mathcal{M}_0$ on $\mathcal{T}$; \\
			\STATE Select correctly predicted target data $\{\mathcal{T}^p\}$ based on the MJKD measures (cf. eq.\ref{eq:1}\texttildelow\ref{eq:4}) using class-balanced strategy; \\
			\STATE Update the labeled and unlabeled training sets as \\
			$\mathcal{S}=\mathcal{S}\bigcup\{\mathcal{T}^p\}$; 
			$\mathcal{T}=\mathcal{T}-\{\mathcal{T}^p\}$; \\
			%\STATE Update the network: $\mathcal{M}_{p+1}=Finetune(\mathcal{M}_p,\mathcal{S})$; \\
			\STATE Apply the adversarial learning to the updated datasets:\\
			$
			\begin{aligned}
			\min\limits_G \max\limits_D & V(G,D) = \\
			& \mathbb{E}_{x_{s}\sim p_{s}(x^{s})}[logD(G^{f}(x^{s}) \otimes G^{p}(x^{s}))] \\
			+ & \mathbb{E}_{x^{t}\sim p_{t}(x^t)}[log(1-D(G^{f}(x^{t}) \otimes G^{p}(x^t)))].
			\end{aligned}
			$
			\renewcommand{\algorithmicensure}{\textbf{Output:}}
			\ENSURE ~~\\
			\STATE Classification results of target data.
		\end{algorithmic}\label{algo}
	\end{algorithm}
	\par}

\section{Analysis}
In this section, we first provide the justifications of the select-and-adapt strategy described in Section III. B by presenting the empirical observations. Then a theoretical driven analysis is given to explain how the proposed method can influence the adversarial learning process.

\subsection{Empirical Observations}  
To evaluate the target data integrating method proposed in Section III.B, we compare the performance of several state-of-the-art methods (TCA~\cite{pan2011domain}, GFK~\cite{gong2012geodesic}, AlexNet~\cite{krizhevsky2012imagenet}, LapCNN~\cite{weston2012deep}, DDC~\cite{tzeng2014deep}, DAN~\cite{long2015learning}, JAN~\cite{long2016deep} and JAN-A~\cite{long2016deep}) with it, denoted as MJKD. The prediction accuracies of CNN on Office-31 dataset are reported in Table~\ref{tab:self_training}. The results show that adding the selected data with their pseudo-labels during the supervised training process can significantly enhance the classification performance of the target data.

Note that the experimental results reported in this subsection was got under the case that MJKD has not involved the adversarial training part, i.e., step 8 of Algorithm~\ref{algo}. After step 7, the CNN was just retrained on the updated training dataset. Table~\ref{tab:self_training} shows that by integrating the selected target data into the source domain, the performance of classifying the target dataset can be significantly improved with respect to using the model pre-trained on the source data only, i.e., AlexNet in Table~\ref{tab:self_training}. When compared with a few recently proposed deep learning methods, MJKD, even without the subsequent adversarial training, has already demonstrated its competitive advantages. In some tasks, MJKD can even achieve the best results (e.g., in task $W\rightarrow A$ and $D\rightarrow A$).

\begin{table*}[ht]
	\centering
	\caption{Comparison of accuracy on Office-31 dataset under conventional unsupervised domain adaptation settings.}%~\cite{long2015learning, long2016deep}
	\begin{tabular}{p{3.5cm}<{\centering}|p{1.5cm}<{\centering}p{1.5cm}<{\centering}p{1.5cm}<{\centering}p{1.5cm}<{\centering}p{1.5cm}<{\centering}p{1.5cm}<{\centering}}
		\hlinew{1.5pt}
		Method & W$\rightarrow$D & W$\rightarrow$A & D$\rightarrow$W & D$\rightarrow$A & A$\rightarrow$W & A$\rightarrow$D \\
		\hline
		\hline
		TCA~\cite{pan2011domain} & 95.2 & 50.9 & 93.2 & 51.6 & 61.0 & 60.8 \\
		GFK~\cite{gong2012geodesic} & 95.0 & 48.1 & 95.6 & 52.4 & 60.4 & 60.6 \\
		AlexNet~\cite{krizhevsky2012imagenet} & 99.0 & 49.8 & 95.1 & 51.1 & 61.6 & 63.8 \\
		LapCNN~\cite{weston2012deep} & 99.1 & 48.2 & 94.7 & 51.6 & 60.4 & 63.1 \\
		DDC~\cite{tzeng2014deep} & 98.5  & 52.2 & 95.0 & 52.1 & 61.8 & 64.4 \\
		%RevGrad~\cite{ganin2015unsupervised} & 1846.653607  & 244572 & 1 & 1 & 1 & 1 \\
		DAN~\cite{long2015learning} & 99.0 & 53.1 & 96.0 & 54.0 & 68.5 & 67.0 \\
		JAN~\cite{long2016deep} & 99.5 & 55.0 & \textbf{96.6} & \textbf{58.3} & 74.9 & 71.8 \\
		JAN-A~\cite{long2016deep} & \textbf{99.6} & 56.3 & \textbf{96.6} & 57.5 & \textbf{75.2} & \textbf{72.8} \\
		\hline
		\hline
		MJKD & 99.4 & \textbf{56.4} & 96.0 & \textbf{58.3} & 70.2 & 70.2 \\
		\hlinew{1.5pt}
	\end{tabular}
	\label{tab:self_training}
\end{table*}

\subsection{Theoretical Observations}
As mentioned in Section III, the probability vectors predicted by the generator can provide important information during the adversarial learning process. In this section, how the probabilities influence the performance of the adversarial network is analyzed.

Given the objective function $V(G,D)$ in eq.~\ref{eq:crossproduct_ad} and following the proof of ``Proposition 1'' in~\cite{goodfellow2014generative}, \textit{for any fixed $G$, the optimal discriminator $D$ in eq.~\ref{eq:crossproduct_ad} is }
%\begin{equation}
%D^{*}_{G}(x \otimes y) = \frac{p_{s}(y|x) p_{s}(x)}{p_{s}(y|x) p_{s}(x) + p_{t}(y|x) p_{t}(x)}
%\end{equation}
\begin{equation}
D^{*}_{G}(f \otimes \tilde{y}) = \frac{G_{s}(f, \tilde{y})}{G_{s}(f,\tilde{y}) + G_{t}(f,\tilde{y})}
\label{eq:op_D}
\end{equation}
where $G_{s}(f, \tilde{y})$ denotes the joint distribution of the deep features and predicted probabilities over each class of the source domain while $G_{t}(f, \tilde{y})$ is that of the target domain. The deep features are denoted by $f$ and the probability vectors are denoted by $\tilde{y}$. Here $\tilde{y} = [\tilde{y}_1,\; \tilde{y}_2,\; \dots,\; \tilde{y}_c]$ where $\tilde{y}_m$ denotes the probability of this data belong to the $m$th category. 
\begin{proof}	
Given any fixed generator $G$, the discriminator $D$ is trained to maximize the value function $V(G,D)$:
%\begin{equation}
%\begin{aligned}
%\min\limits_G \max\limits_D & V(G,D) \\
%= & \int_{x_{s}} p_{s}(x^{s})[logD(G^{f}(x^{s}) \otimes G^{p}(x^{s}))] \\
%+ & \int_{x^{t}} p_{t}(x^t)[log(1-D(G^{f}(x^{t}) \otimes G^{p}(x^t)))] \\
%= & \int_{x} p_{s}(x) p_{s}(y|x)logD(x \otimes y) \\
%+ & p_{t}(x) p_{t}(y|x)log(1-D(x \otimes y)) d_x \\
%\end{aligned}
%\label{eq:optimal_D}
%\end{equation}
\begin{equation}
\begin{aligned}
\min\limits_G \max\limits_D & V(G,D) \\
= & \int_{x^{s}} p_{s}(x^{s})[logD(G^{f}(x^{s}) \otimes G^{p}(x^{s}))]d_{x_s} \\
+ & \int_{x^{t}} p_{t}(x^t)[log(1-D(G^{f}(x^{t}) \otimes G^{p}(x^t)))]d_{x_t} \\
= & \int_{f}\int_{\tilde{y}} G_{s}(f,\tilde{y}) logD(f \otimes \tilde{y}) \\
+ & G_{t}(f,\tilde{y}) log(1-D(f \otimes \tilde{y})) d_f d_{\tilde{y}} \\
\end{aligned}
\label{eq:optimal_D}
\end{equation}
where $G_{s}(f,\tilde{y}) = (G^{f}(x^s),G^{p}(x^s))_{x^s \sim p_{s}(x^s)}$ and $G_{t}(f,\tilde{y}) = (G^{f}(x^t),G^{p}(x^t))_{x^t \sim p_{t}(x^t)}$. Eq.~\ref{eq:optimal_D} has the same form as function $y \rightarrow a\, log(y) + b\,log(1 - y), (a,b) \in \mathbb{R}^2 \setminus \{0,0\}$, which achieves its maximum at $\frac{a}{a + b} \in [0,1]$. So similarly, given $G$ fixed, the optimal $D$ that makes $V(G,D)$ achieve its maximum can be obtained as in eq.~\ref{eq:op_D}. 
\end{proof}
Then, by substituting the optimal discriminator $D^{*}_{G}$ into eq.~\ref{eq:crossproduct_ad}, the training criterion for $G$ is to minimize

%\begin{equation}
%\begin{aligned}
%& V(G,D^{*}_{G}) \\
%& = \mathbb{E}_{x^{s}\sim p_{s}(x^{s})}[logD^{*}_{G}(G^{f}(x^{s}) \otimes G^{p}(x^{s}))] \\
%& + \mathbb{E}_{x^{t}\sim p_{t}(x^{t})}[log(1-D^{*}_{G}(G^{f}(x^{t}) \otimes G^{p}(x^t)))] \\
%& = \mathbb{E}_{(x,y) \sim G_{s}(x,y)}[logD^{*}(x \otimes y)] \\
%& + \mathbb{E}_{(x,y) \sim G_{t}(x,y)}[log(1-D^{*}_{G}(x \otimes y))] \\
%& = \mathbb{E}_{(x,y) \sim G_{s}(x,y)}[log\frac{p_{s}(x) p_{s}(y|x)}{p_{s}(x) p_{s}(y|x) + p_{t}(x) p_{t}(y|x)}] \\
%& + \mathbb{E}_{(x,y) \sim G_{t}(x,y)}[log\frac{p_{t}(x) p_{t}(y|x)}{p_{s}(x) p_{s}(y|x) + p_{t}(x) p_{t}(y|x)}] \\
%\end{aligned}
%\label{eq:min_G}
%\end{equation}
\begin{equation}
\begin{aligned}
& V(G,D^{*}_{G}) \\
& = \mathbb{E}_{x^{s}\sim p_{s}(x^{s})}[logD^{*}_{G}(G^{f}(x^{s}) \otimes G^{p}(x^{s}))] \\
& + \mathbb{E}_{x^{t}\sim p_{t}(x^{t})}[log(1-D^{*}_{G}(G^{f}(x^{t}) \otimes G^{p}(x^t)))] \\
& = \mathbb{E}_{(f,\tilde{y}) \sim G_{s}(f,\tilde{y})}[logD^{*}(f \otimes \tilde{y})] \\
& + \mathbb{E}_{(f,\tilde{y}) \sim G_{t}(f,\tilde{y})}[log(1-D^{*}_{G}(f \otimes \tilde{y}))] \\
& = \mathbb{E}_{(f,\tilde{y}) \sim G_{s}(f,\tilde{y})}[log\frac{G_{s}(f,\tilde{y})}{G_{s}(f,\tilde{y}) + G_{t}(f,\tilde{y})}] \\
& + \mathbb{E}_{(f,\tilde{y}) \sim G_{t}(f,\tilde{y})}[log\frac{G_{t}(f,\tilde{y})}{G_{s}(f,\tilde{y}) + G_{t}(f,\tilde{y})}] \\
\end{aligned}
\label{eq:min_G}
\end{equation}
According to~\cite{goodfellow2014generative}, it is straightforward to induce that eq.~\ref{eq:min_G} can be reformulated to
\begin{equation}
V(G,D^{*}_{G}) = -log(4) + 2\cdot JSD(G_{s}(f,\tilde{y})\parallel G_{t}(f,\tilde{y}))
\end{equation}
We can see that when $G_{s}(f,\tilde{y}) = G_{t}(f,\tilde{y})$, the global minimum can be achieved as the Jensen-Shannon divergence (JSD) between two distributions is always non-negative and equals to zero iff they are exactly the same. To sum up, in this adversarial architecture, the deep neural network $G$ tends to generate equally distributed probability-feature joint outputs for the target and source data
%\begin{equation}
%p_{s}(x)p_{s}(y|x)= p_{t}(x) p_{t}(y|x).
%\label{eq:distribution}
%\end{equation} 
\begin{equation}
\begin{aligned}
p(G^{f}(x^s))p(G^{p}(x^s)|G^{f}(x^s)) = \\
p(G^{f}(x^t))p(G^{p}(x^t)|G^{f}(x^t)).
\end{aligned}
\label{eq:distribution}
\end{equation} 
From eq.~\ref{eq:distribution}, it can be inferred that if the distribution of probability vectors predicted for source and target data are similar, i.e., $p(G^{p}(x^s)|G^{f}(x^s))\approx p(G^{p}(x^t)|G^{f}(x^t))$, their features will tend to be mapped close to each other, i.e., $p(G^{f}(x^s))\approx p(G^{f}(x^t))$. And it can be inferred that if a target data $t_i$ is classified to the same category as source data $s_j$, then $p(G^{f}(s_j))\approx G^{f}(t_i))$ as the probability vectors which indicate the same category have similar distribution to each other. So, in order to achieve eq.~\ref{eq:distribution}, the generator discussed here tends to map target and source data close if they are classified into the same category. Thus, it can be inferred that if the accuracy of $p(G^{p}(x^t)|G^{f}(x^t))$ can be enhanced, more target data will be mapped to the data spaces corresponding to their ground truth labels, which can further improve the accuracy.

\begin{figure}[h]
	\centering
	\subfigure{
		\includegraphics[width=0.35in,height=0.35in]{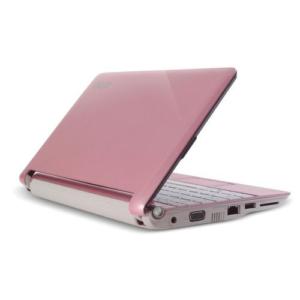}
		\includegraphics[width=0.35in,height=0.35in]{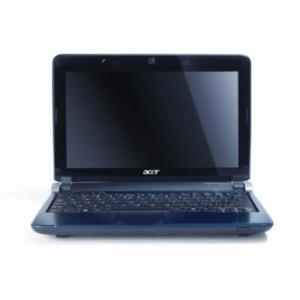}
		\includegraphics[width=0.35in,height=0.35in]{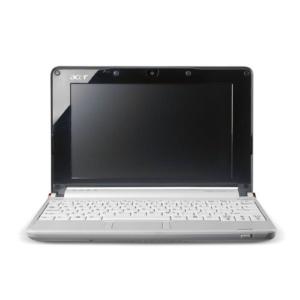}
		\includegraphics[width=0.35in,height=0.35in]{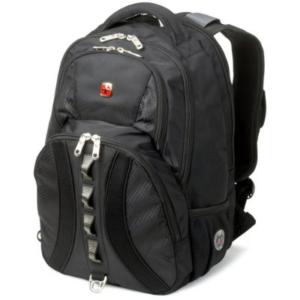}
		\includegraphics[width=0.35in,height=0.35in]{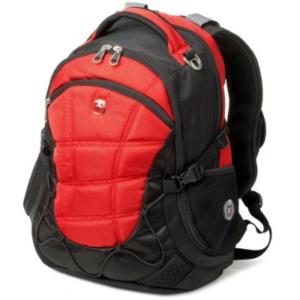}
		\includegraphics[width=0.35in,height=0.35in]{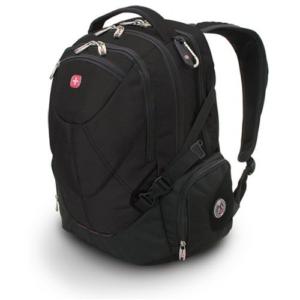}
		\includegraphics[width=0.35in,height=0.35in]{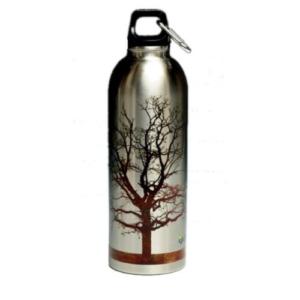}
		\includegraphics[width=0.35in,height=0.35in]{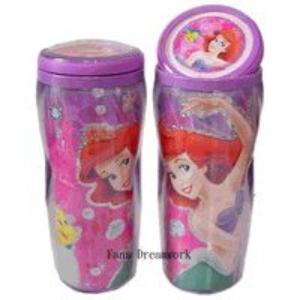}
	}\\
	\subfigure{
		\includegraphics[width=0.35in,height=0.35in]{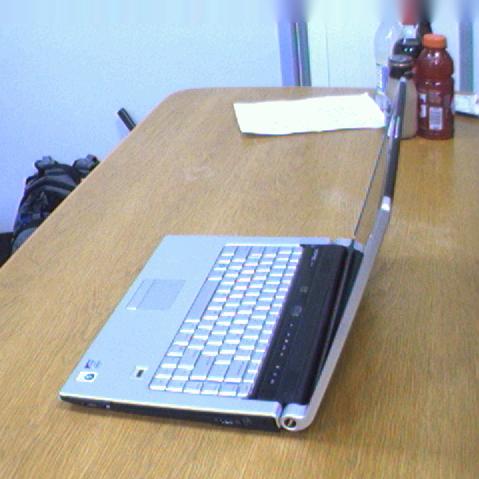}
		\includegraphics[width=0.35in,height=0.35in]{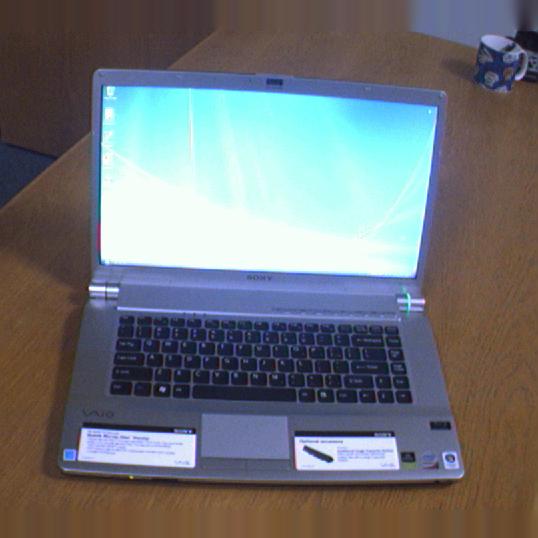}
		\includegraphics[width=0.35in,height=0.35in]{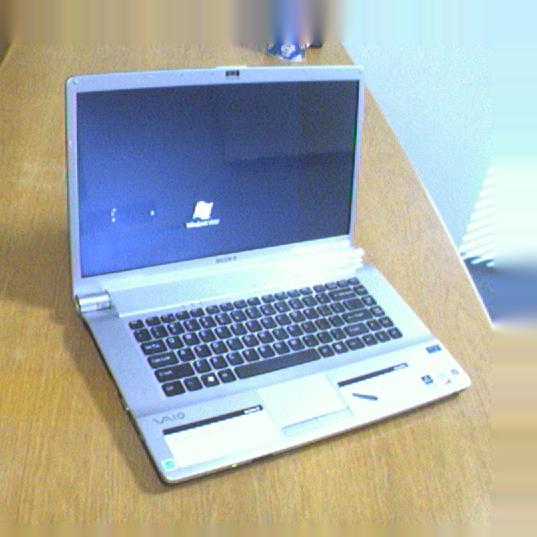}
		\includegraphics[width=0.35in,height=0.35in]{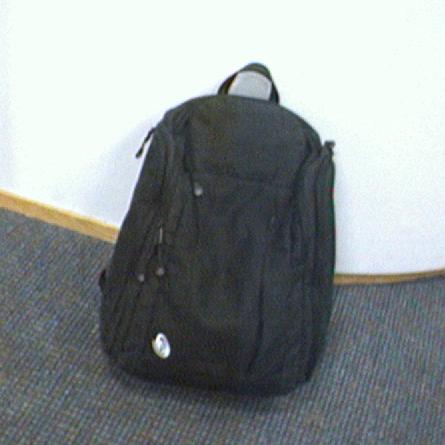}
		\includegraphics[width=0.35in,height=0.35in]{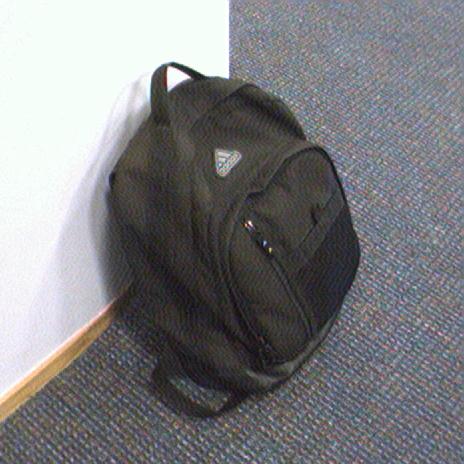}
		\includegraphics[width=0.35in,height=0.35in]{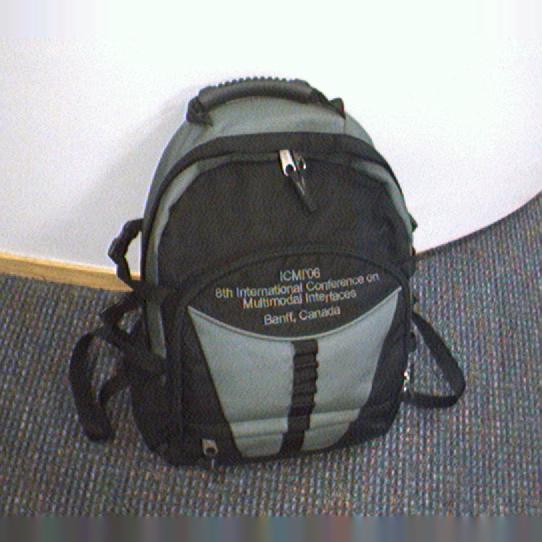}
		\includegraphics[width=0.35in,height=0.35in]{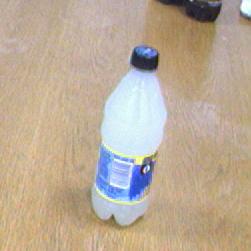}
		\includegraphics[width=0.35in,height=0.35in]{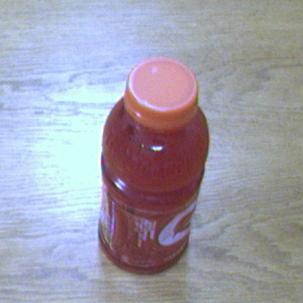}
	}\addtocounter{subfigure}{-2}\\
	\subfigure[]{
		\includegraphics[width=0.35in,height=0.35in]{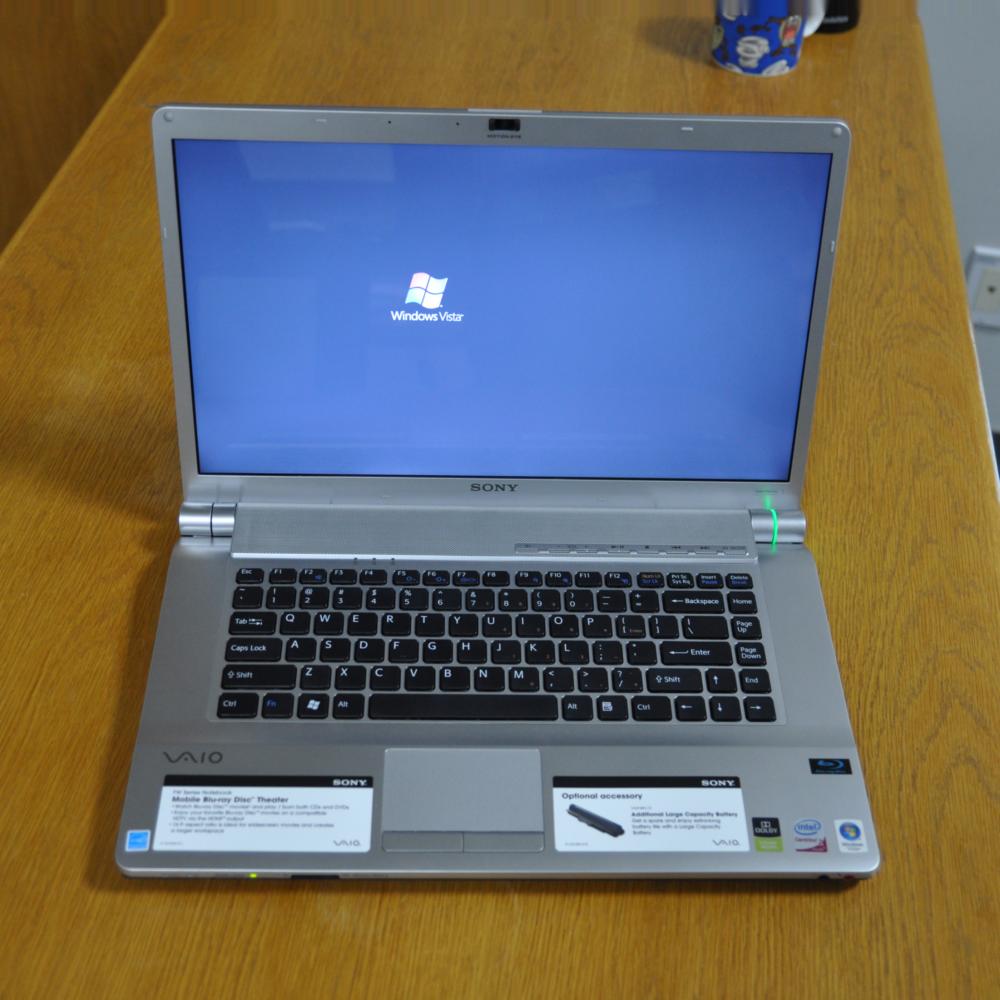}
		\includegraphics[width=0.35in,height=0.35in]{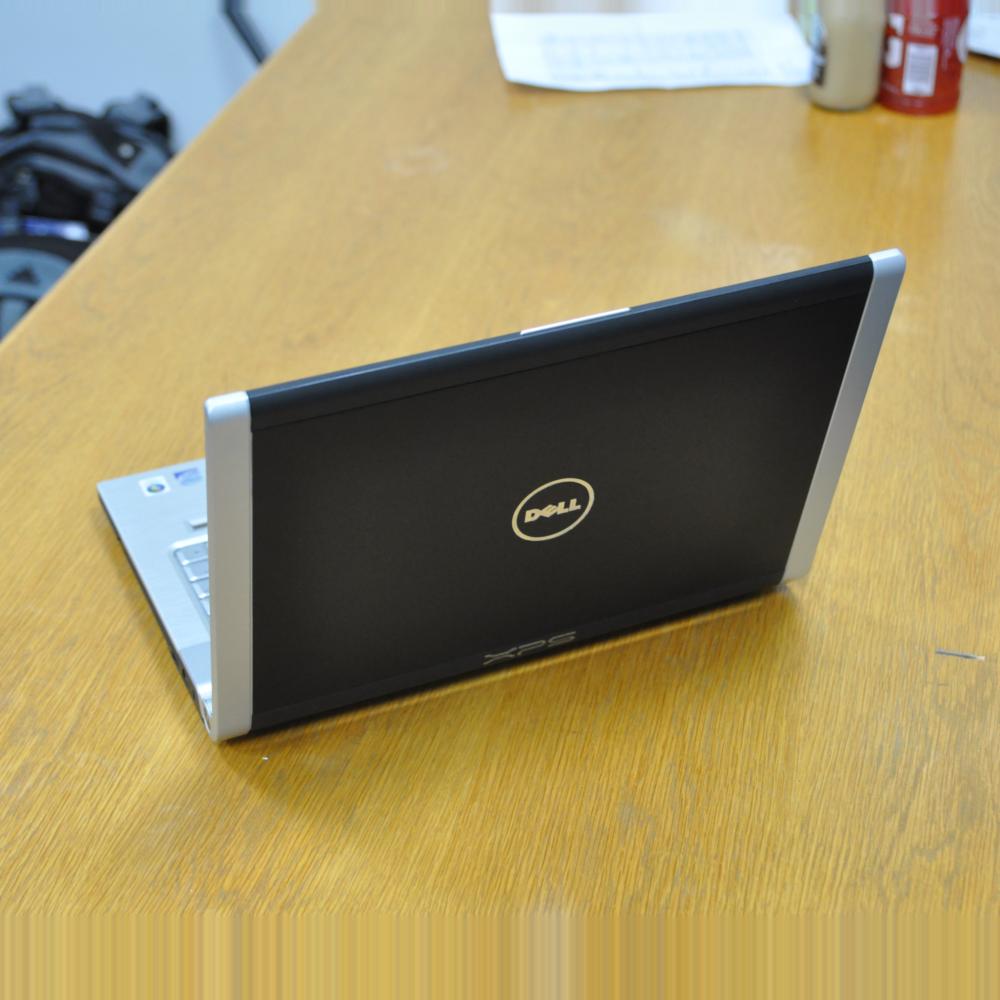}
		\includegraphics[width=0.35in,height=0.35in]{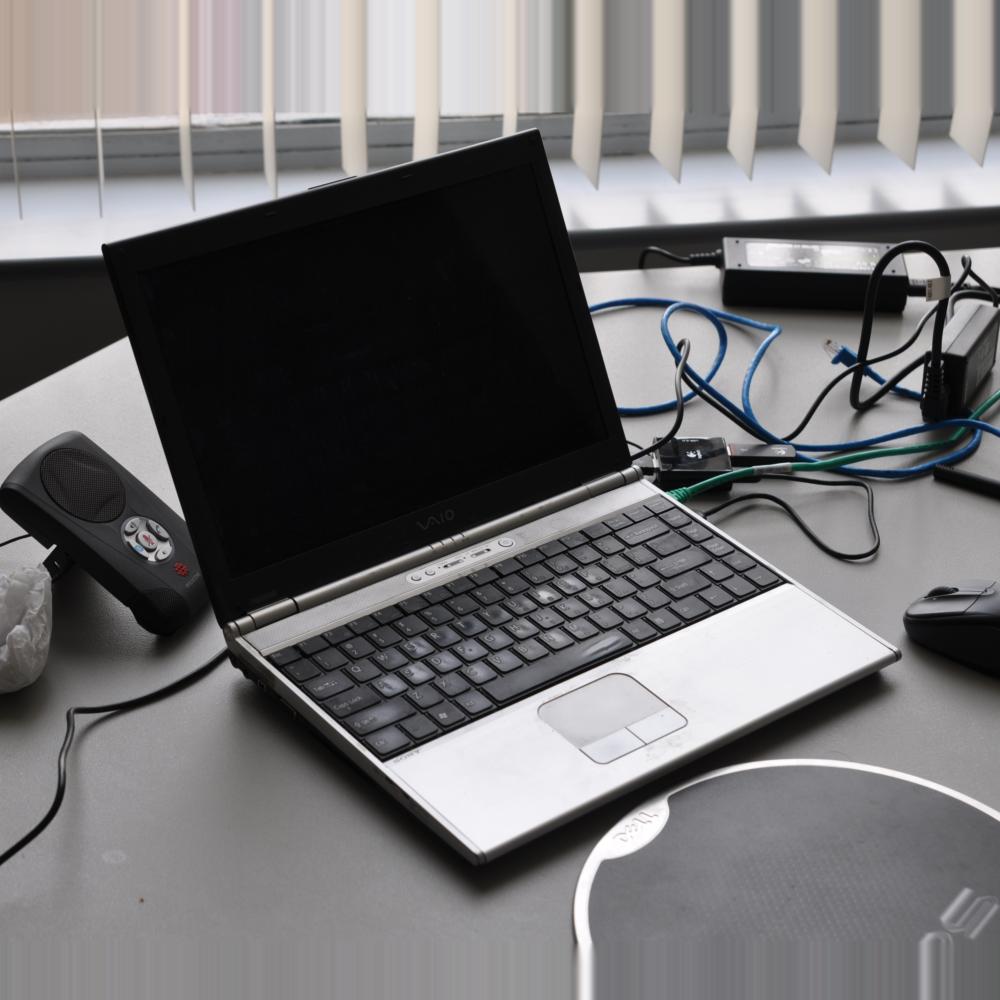}
		\includegraphics[width=0.35in,height=0.35in]{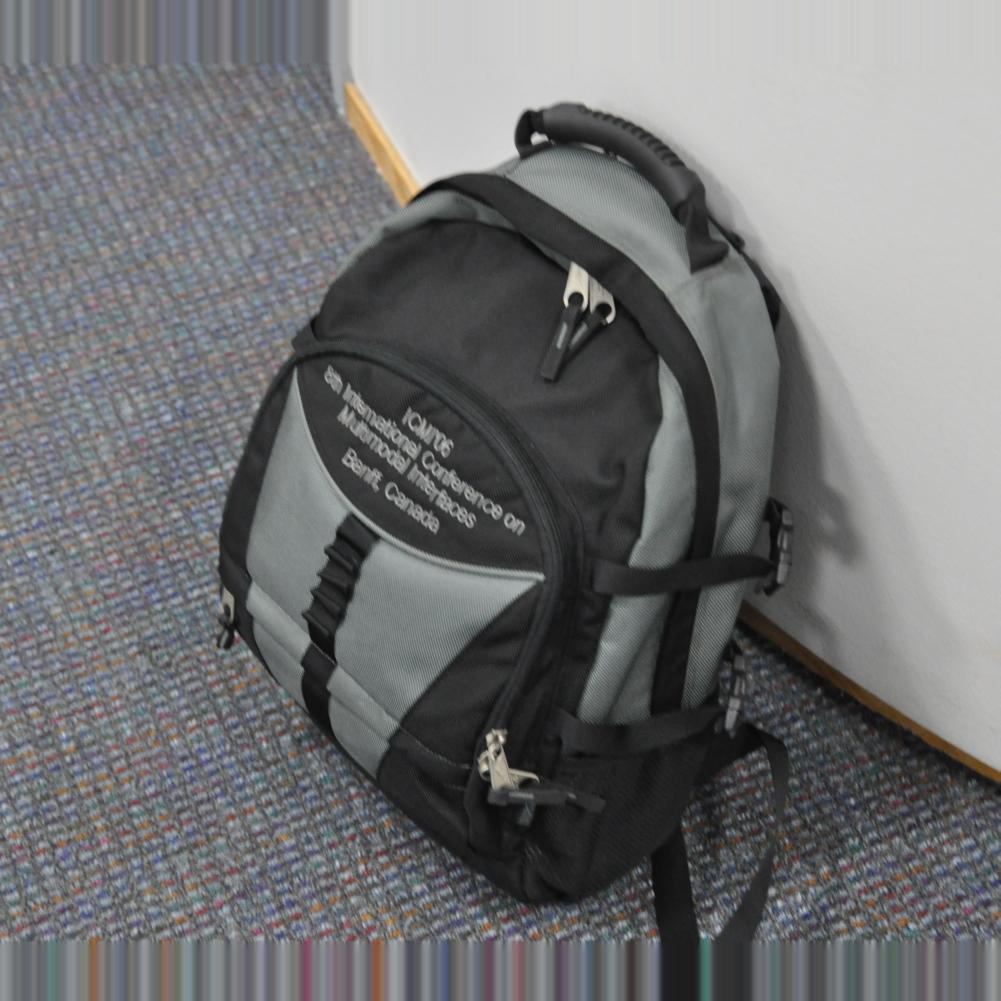}
		\includegraphics[width=0.35in,height=0.35in]{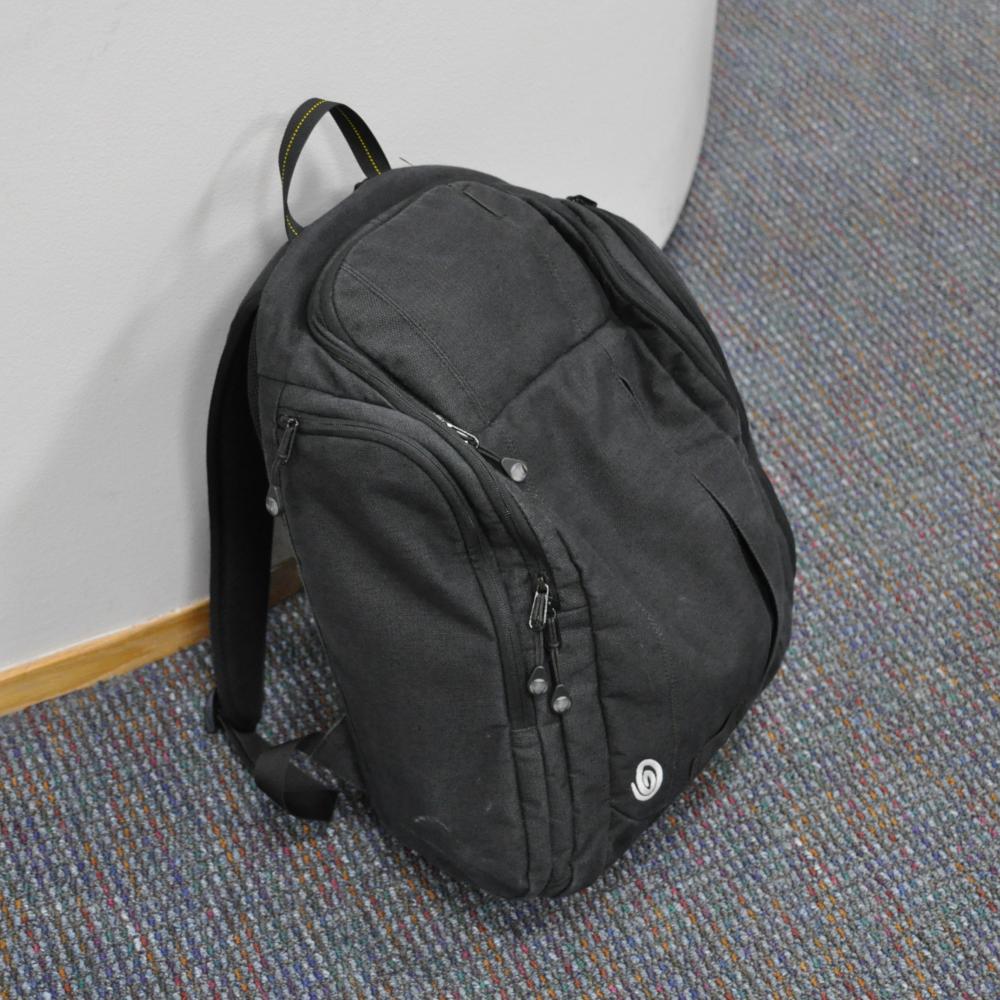}
		\includegraphics[width=0.35in,height=0.35in]{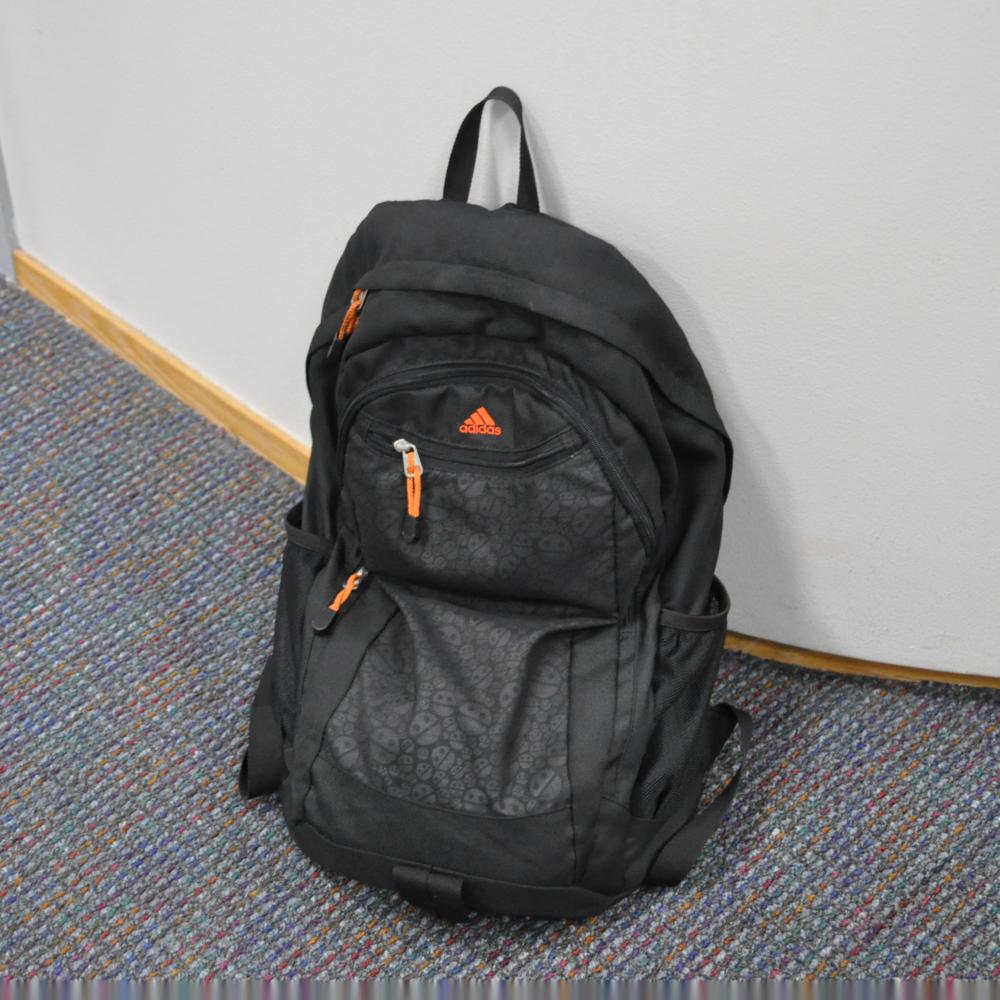}
		\includegraphics[width=0.35in,height=0.35in]{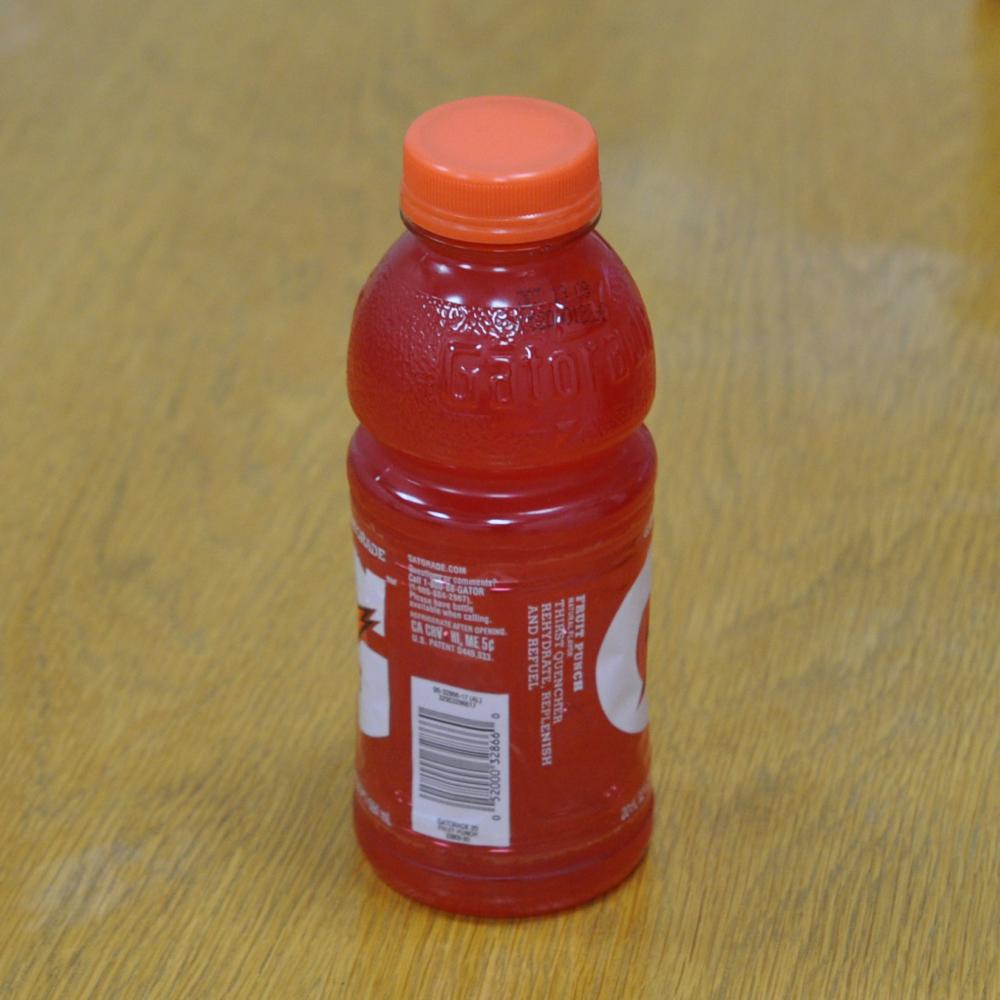}
		\includegraphics[width=0.35in,height=0.35in]{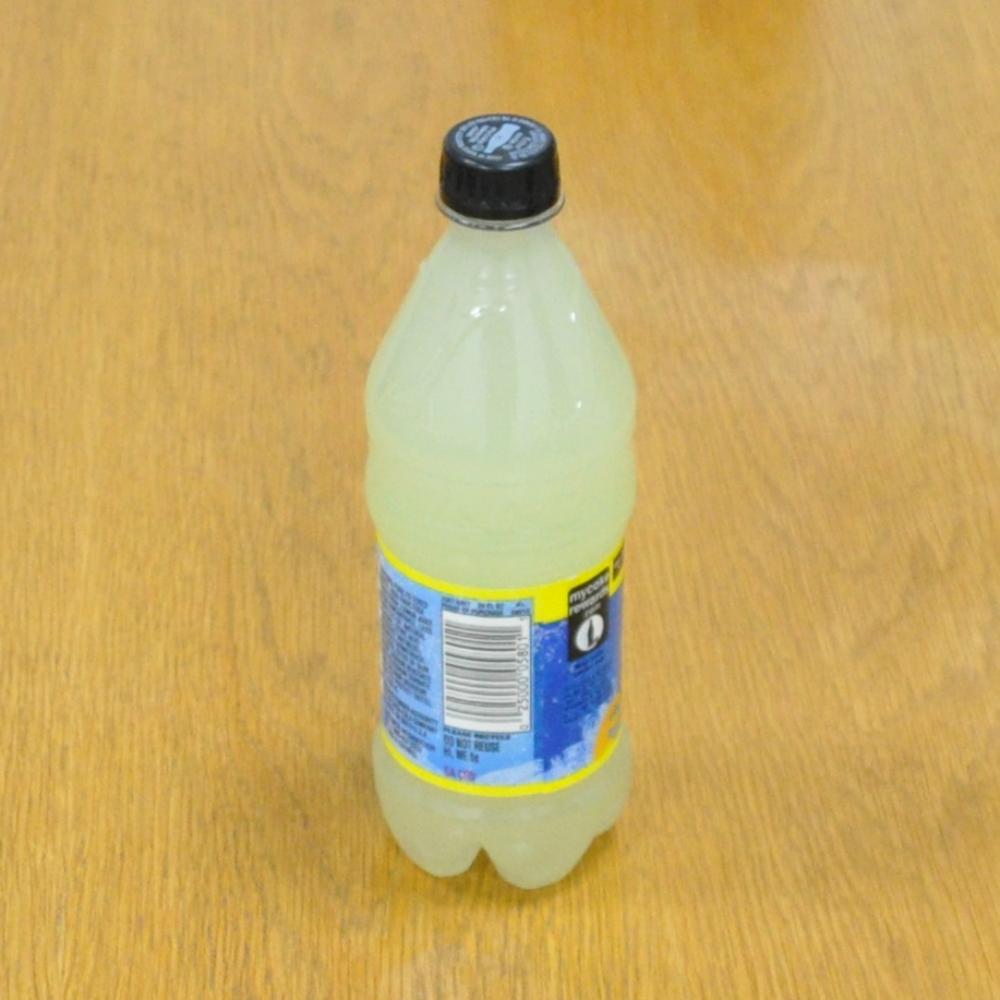}
	}\\
	\subfigure{
		\includegraphics[width=0.35in,height=0.35in]{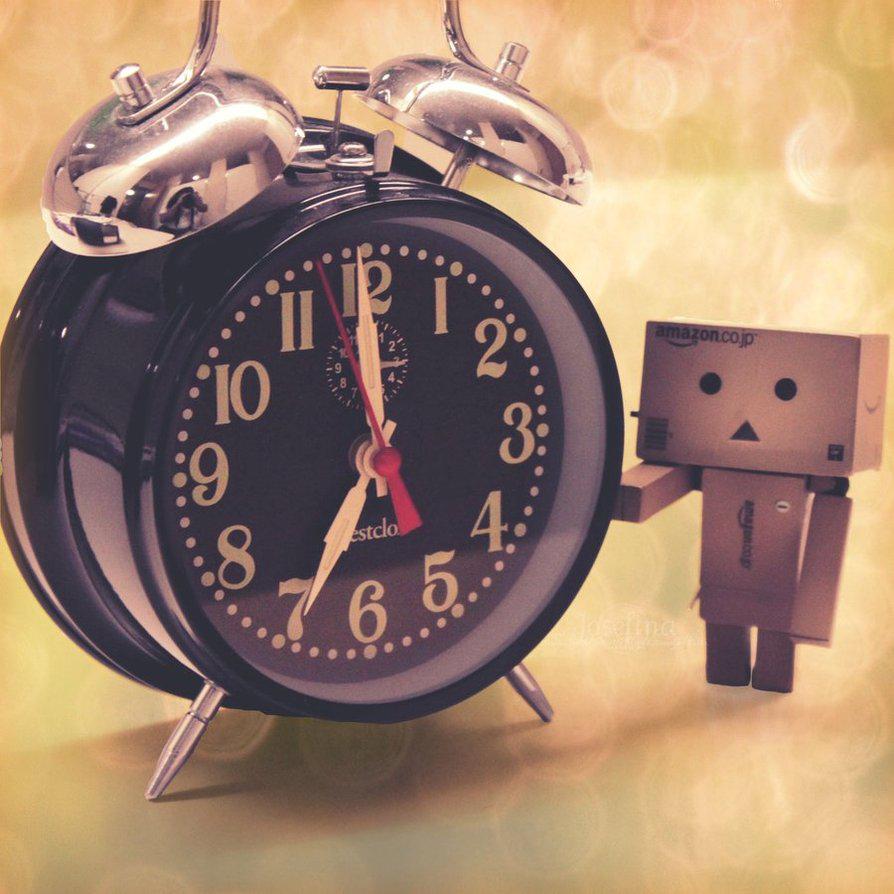}
		\includegraphics[width=0.35in,height=0.35in]{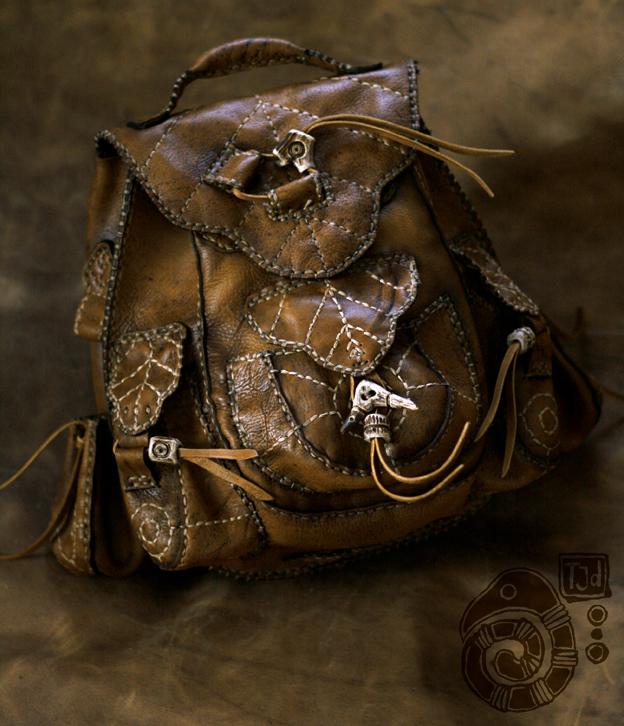}
		\includegraphics[width=0.35in,height=0.35in]{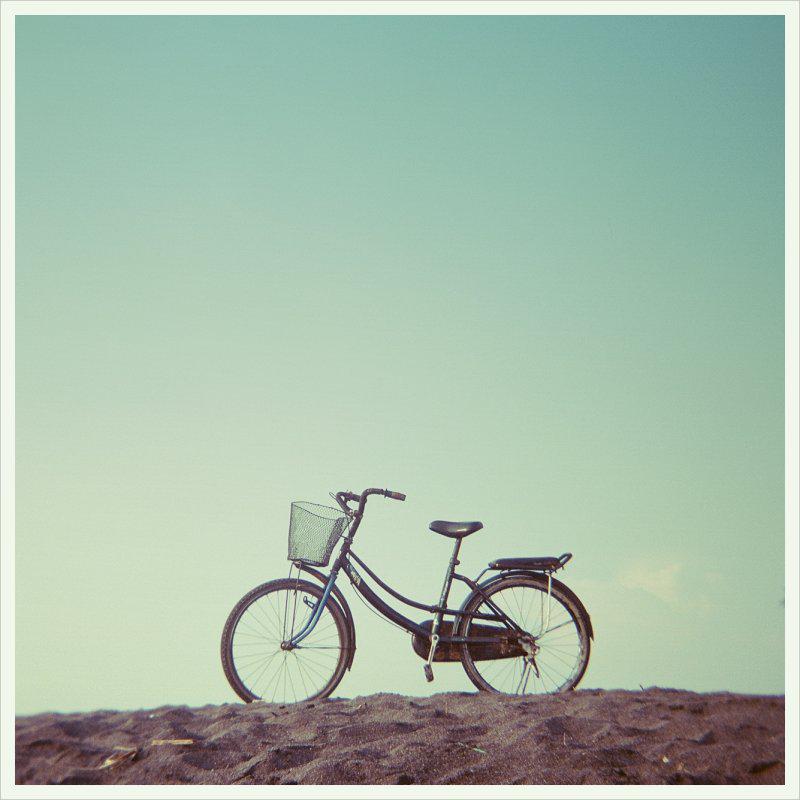}
		\includegraphics[width=0.35in,height=0.35in]{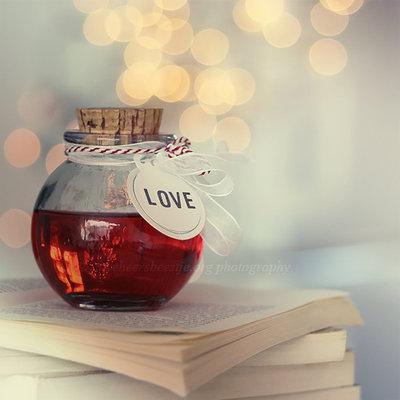}
		\includegraphics[width=0.35in,height=0.35in]{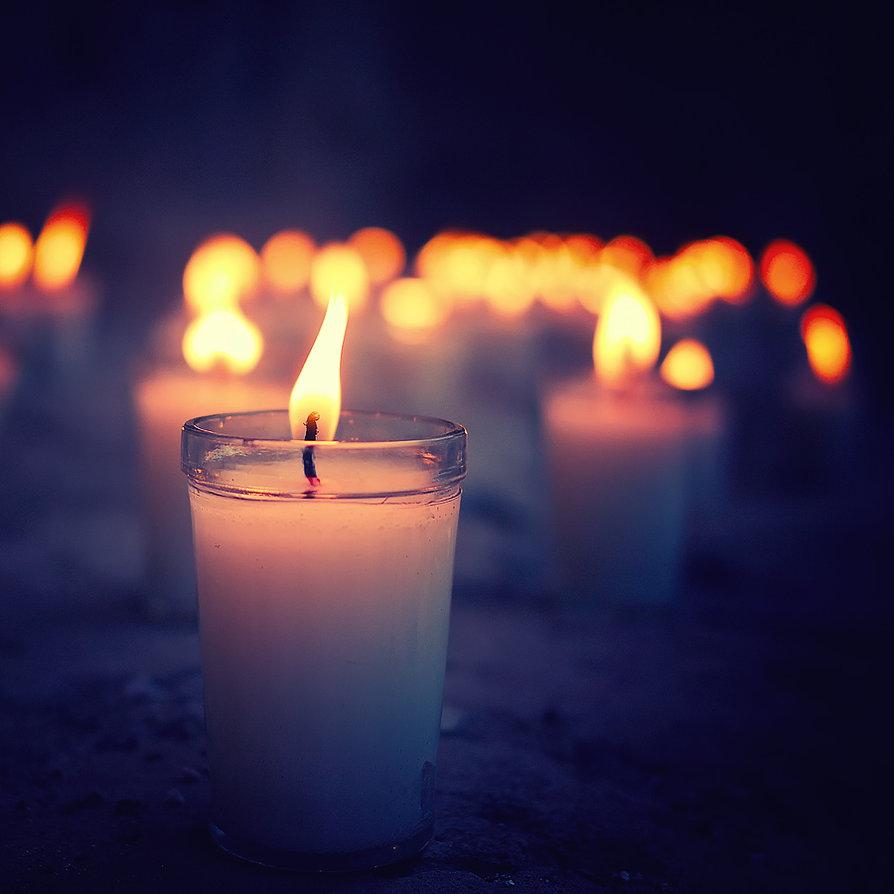}
		\includegraphics[width=0.35in,height=0.35in]{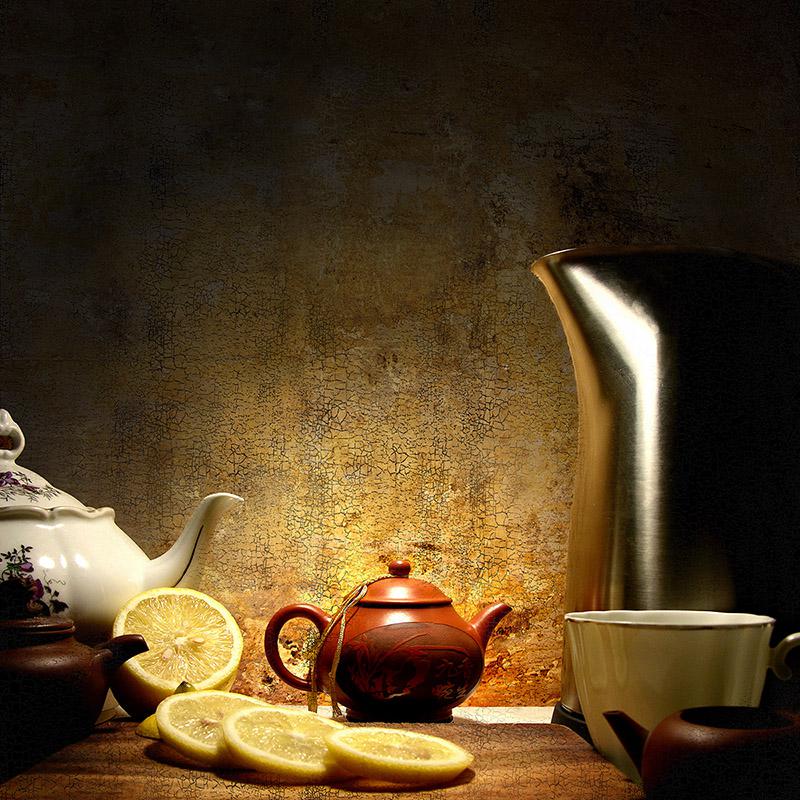}
		\includegraphics[width=0.35in,height=0.35in]{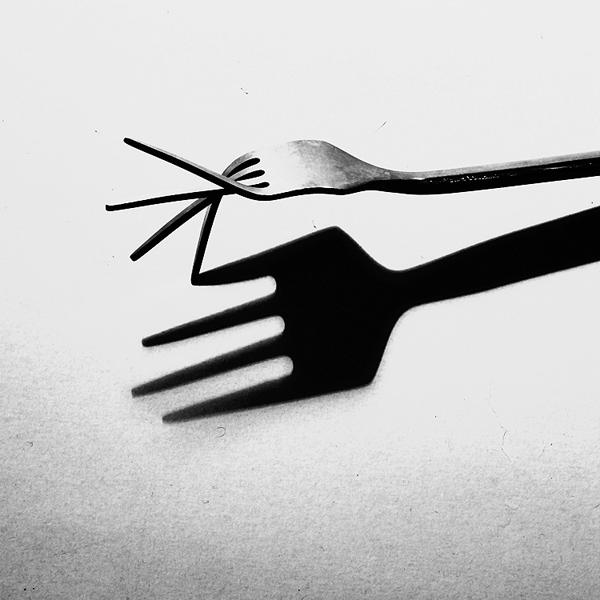}
		\includegraphics[width=0.35in,height=0.35in]{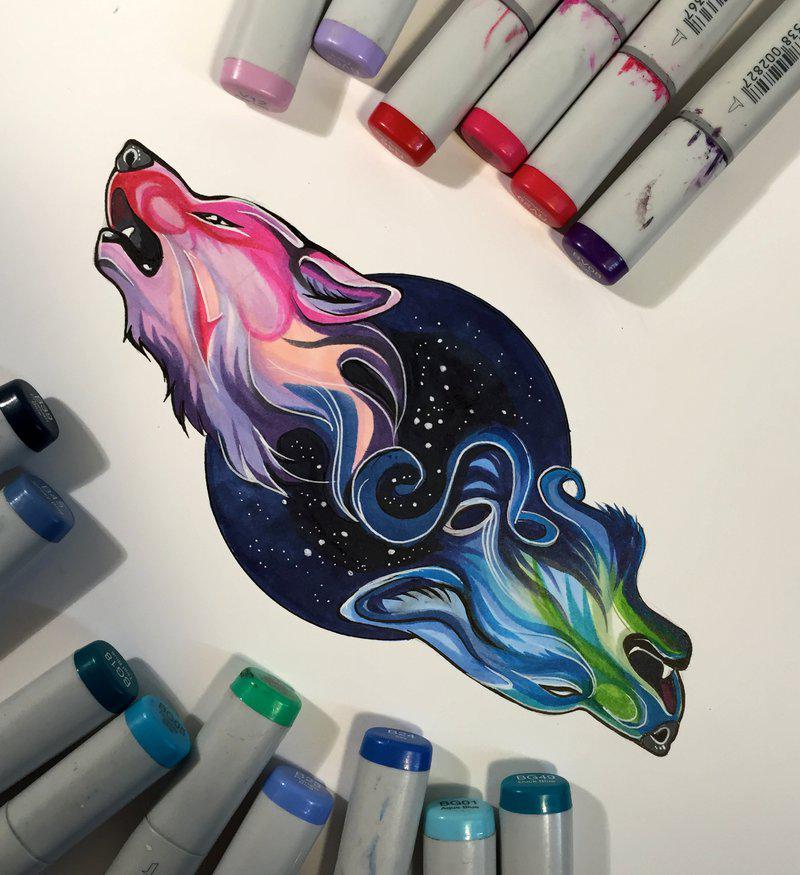}
	}\\
	\subfigure{
		\includegraphics[width=0.35in,height=0.35in]{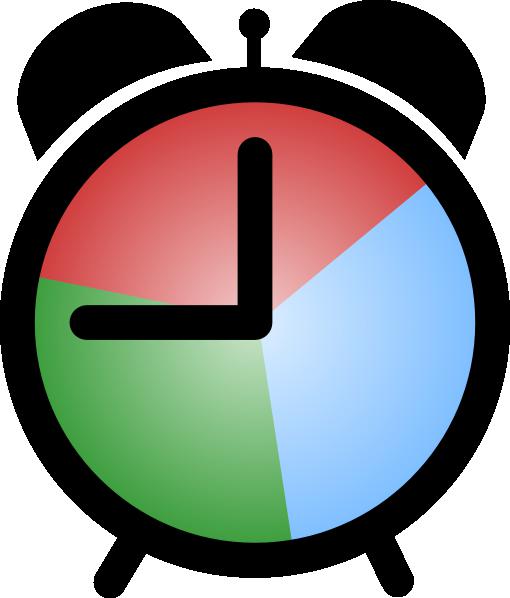}
		\includegraphics[width=0.35in,height=0.35in]{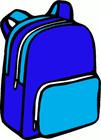}
		\includegraphics[width=0.35in,height=0.35in]{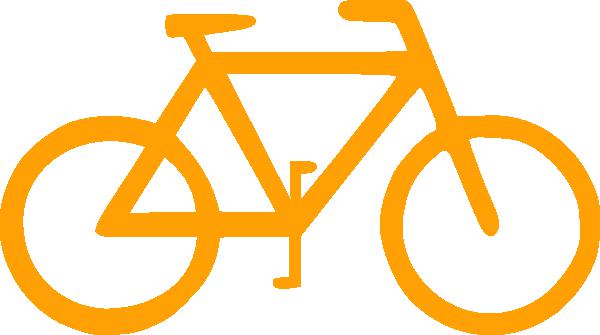}
		\includegraphics[width=0.35in,height=0.35in]{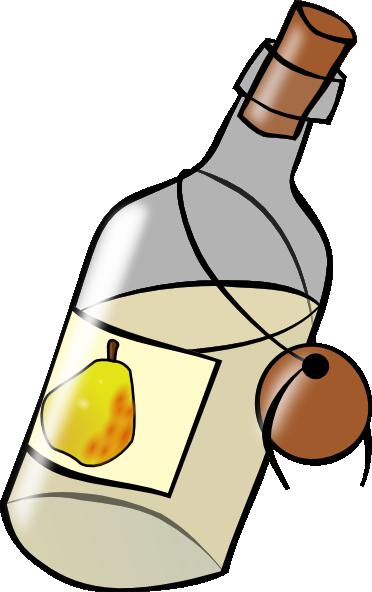}
		\includegraphics[width=0.35in,height=0.35in]{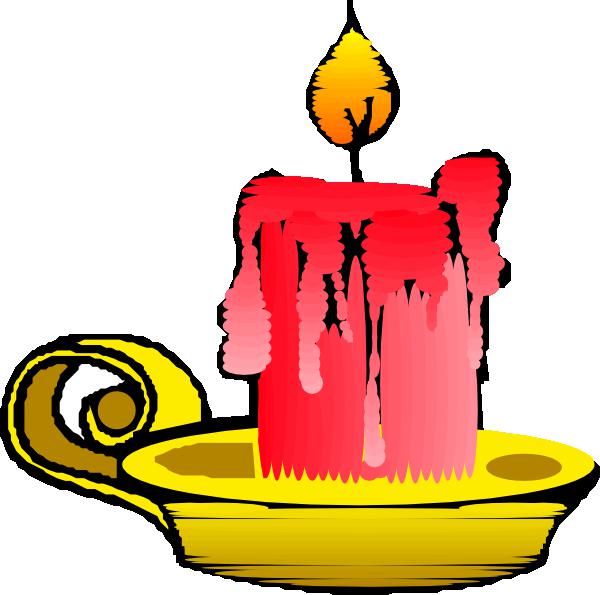}
		\includegraphics[width=0.35in,height=0.35in]{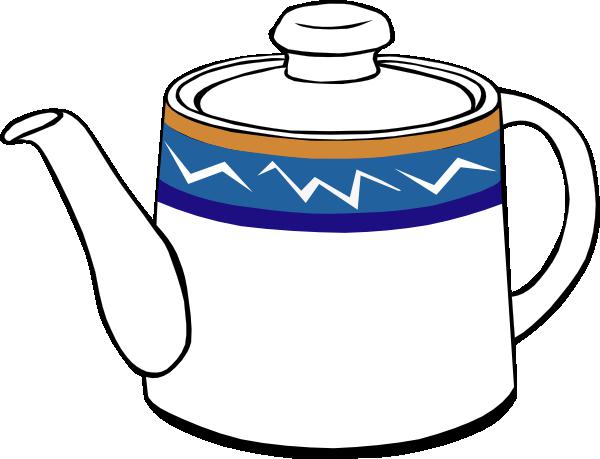}
		\includegraphics[width=0.35in,height=0.35in]{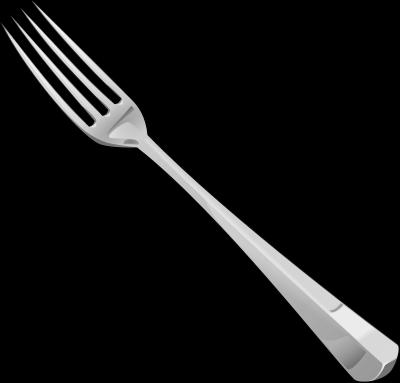}
		\includegraphics[width=0.35in,height=0.35in]{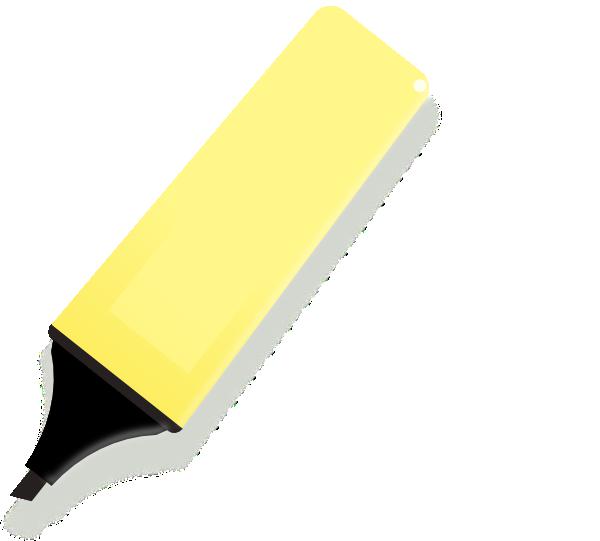}
	}\\
	\subfigure{
		\includegraphics[width=0.35in,height=0.35in]{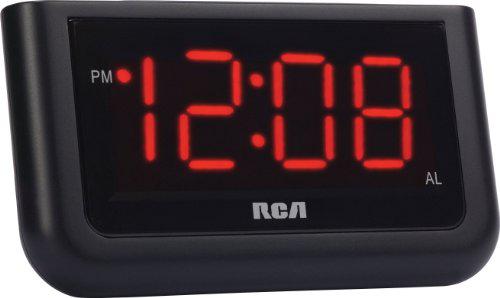}
		\includegraphics[width=0.35in,height=0.35in]{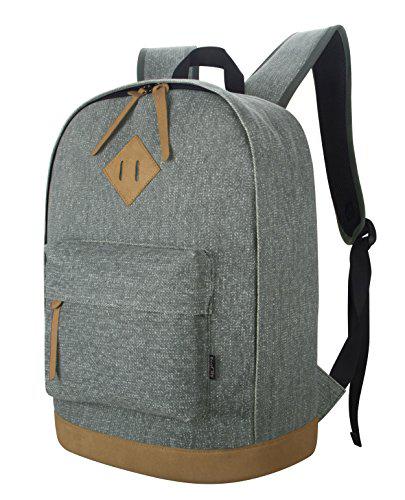}
		\includegraphics[width=0.35in,height=0.35in]{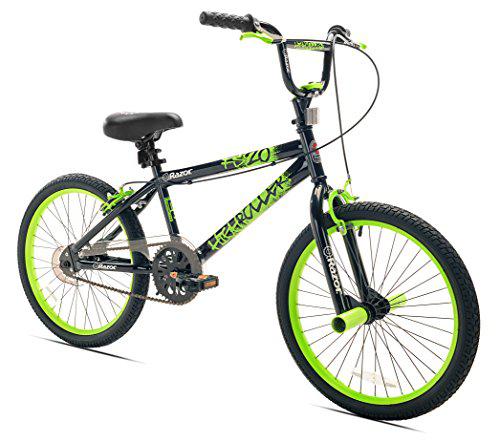}
		\includegraphics[width=0.35in,height=0.35in]{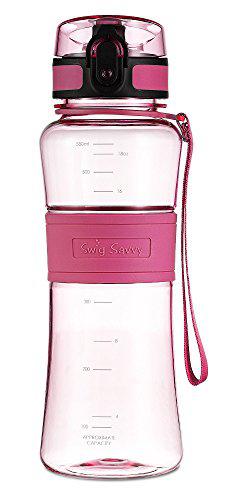}
		\includegraphics[width=0.35in,height=0.35in]{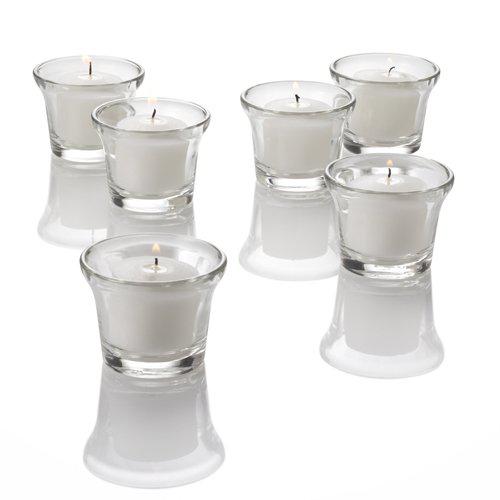}
		\includegraphics[width=0.35in,height=0.35in]{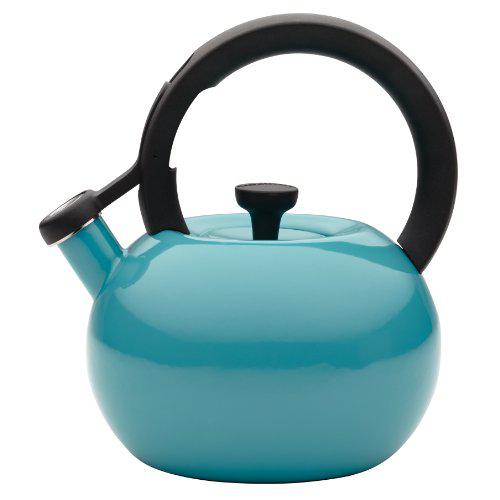}
		\includegraphics[width=0.35in,height=0.35in]{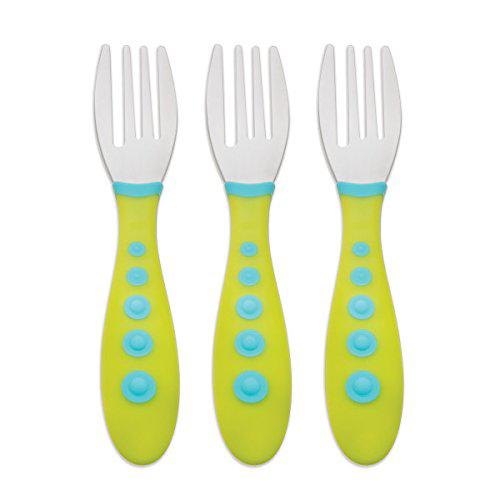}
		\includegraphics[width=0.35in,height=0.35in]{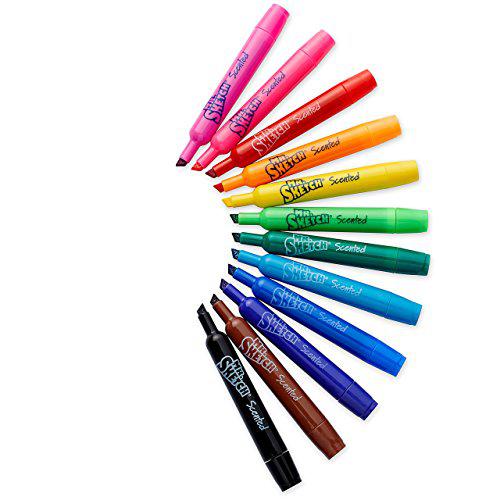}
	}\addtocounter{subfigure}{-3}\\
	\subfigure[]{
		\includegraphics[width=0.35in,height=0.35in]{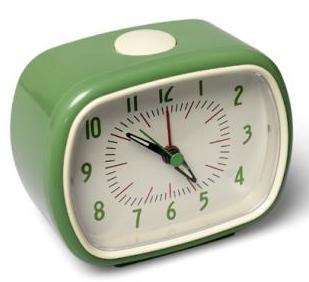}
		\includegraphics[width=0.35in,height=0.35in]{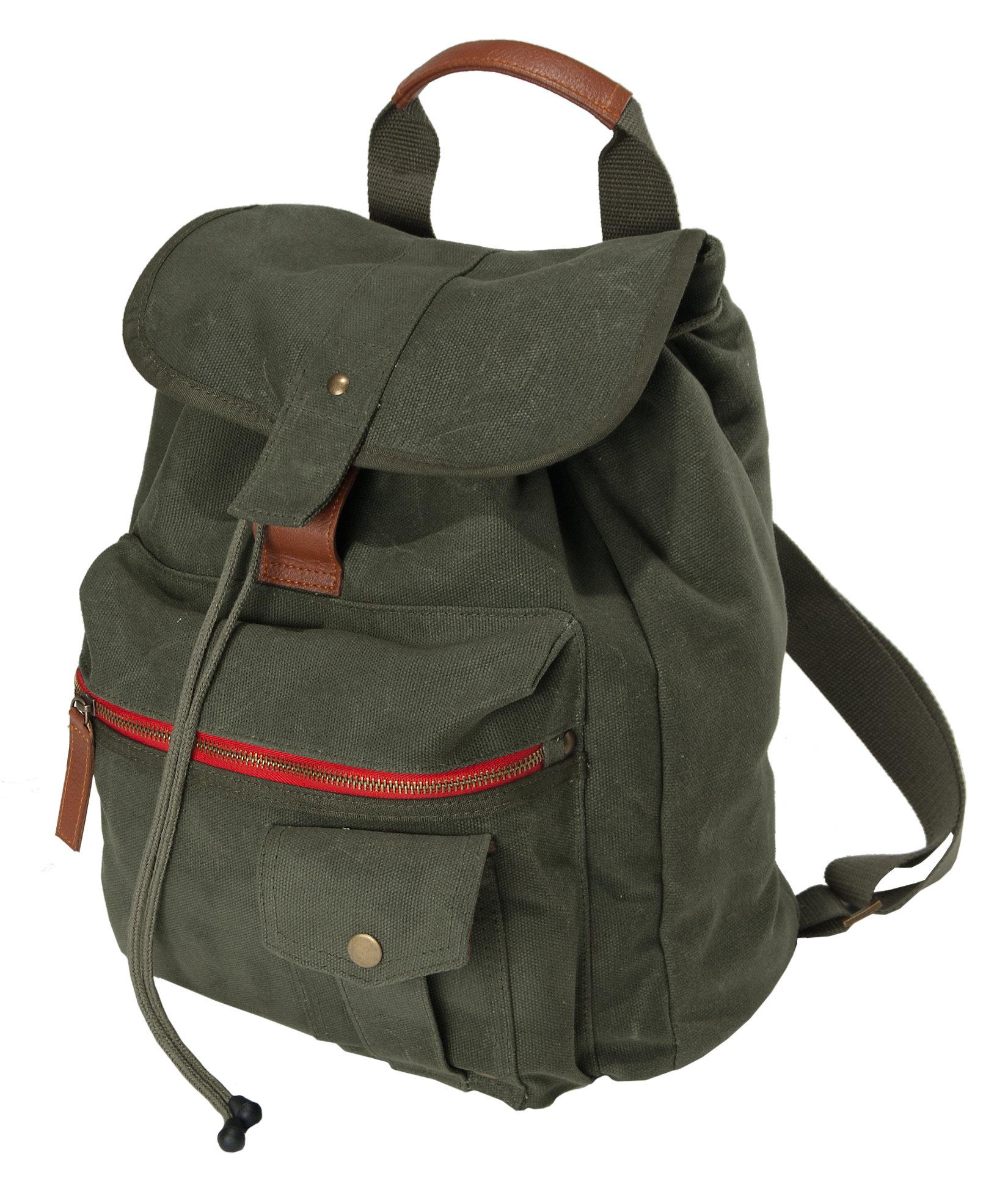}
		\includegraphics[width=0.35in,height=0.35in]{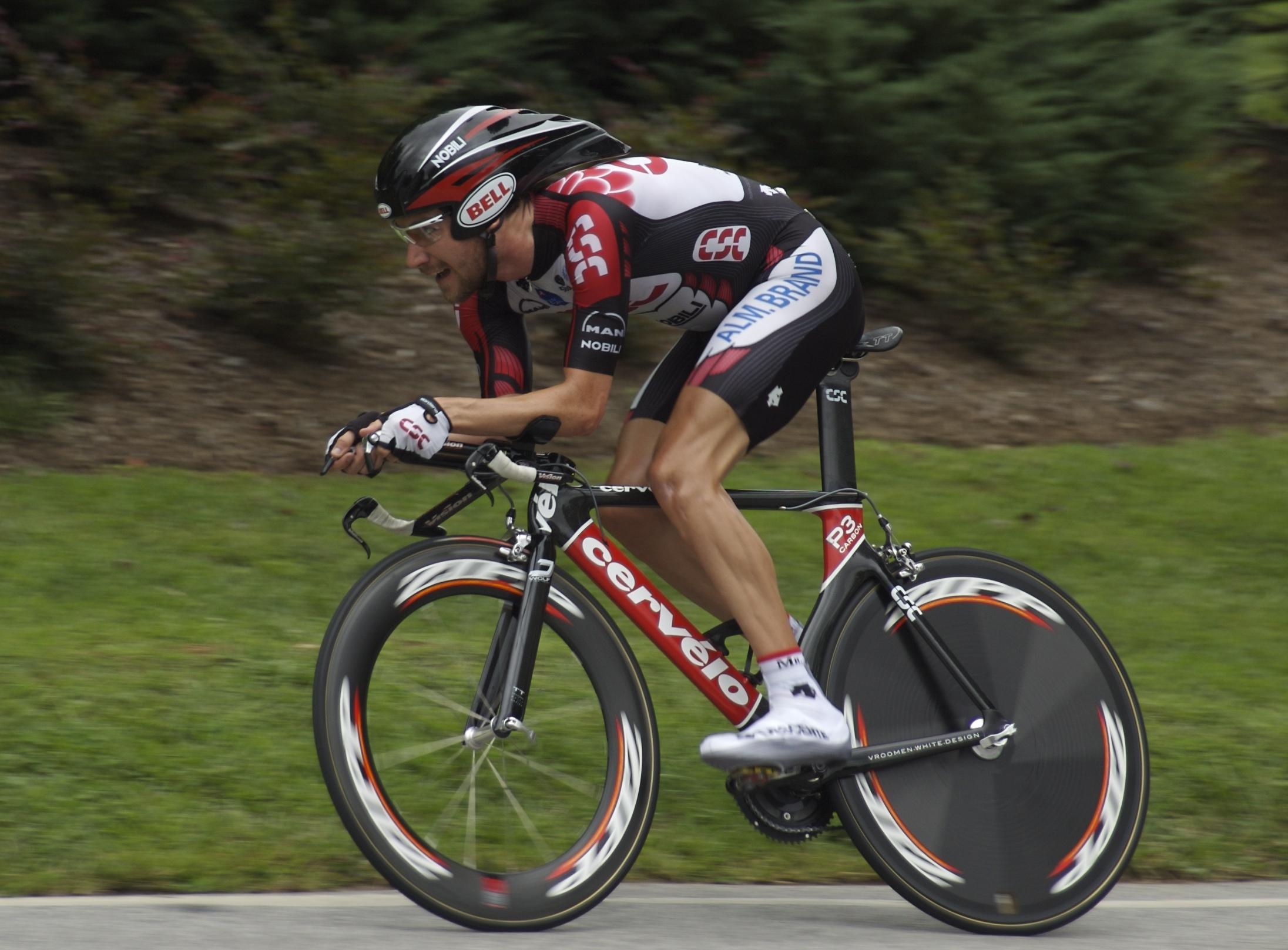}
		\includegraphics[width=0.35in,height=0.35in]{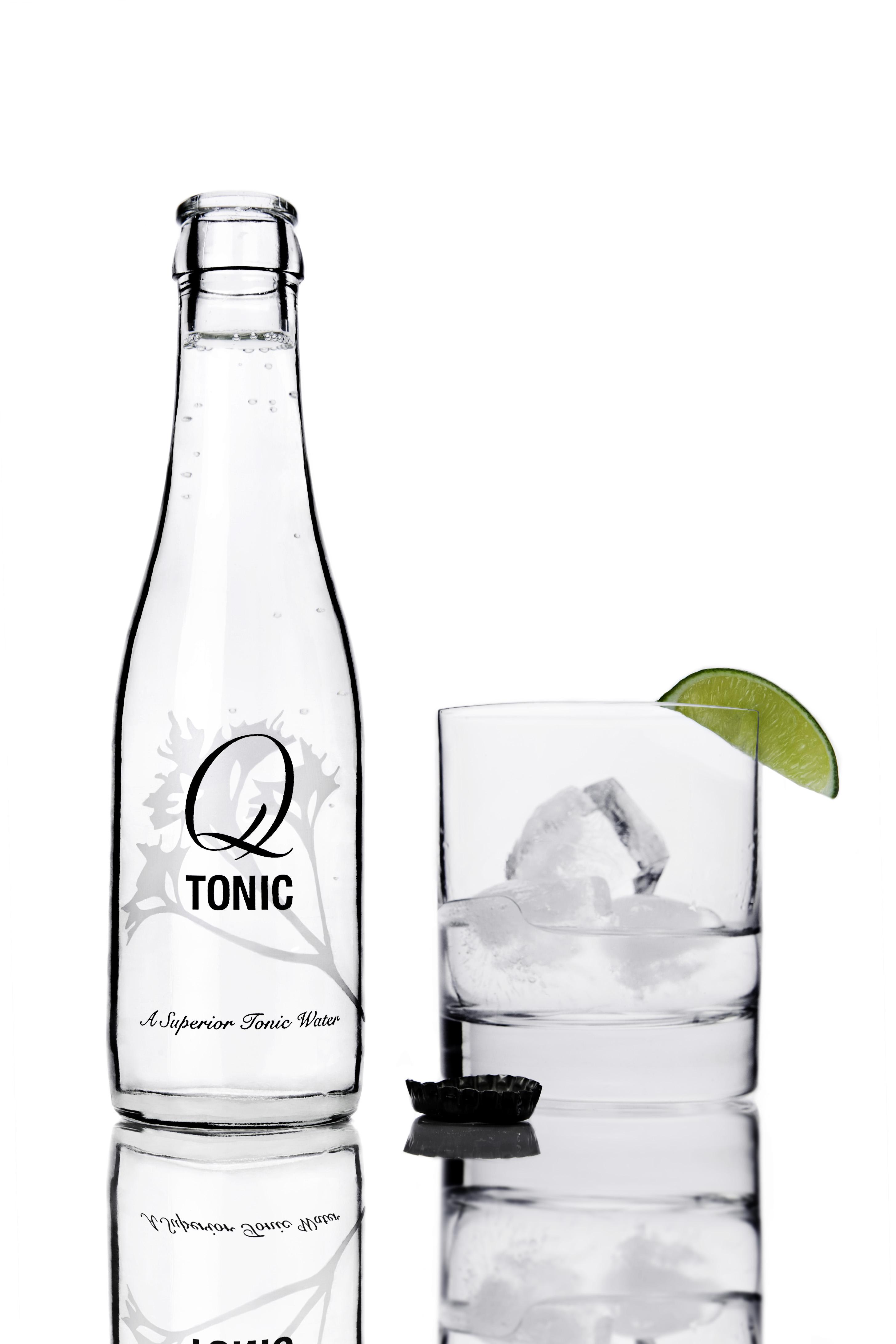}
		\includegraphics[width=0.35in,height=0.35in]{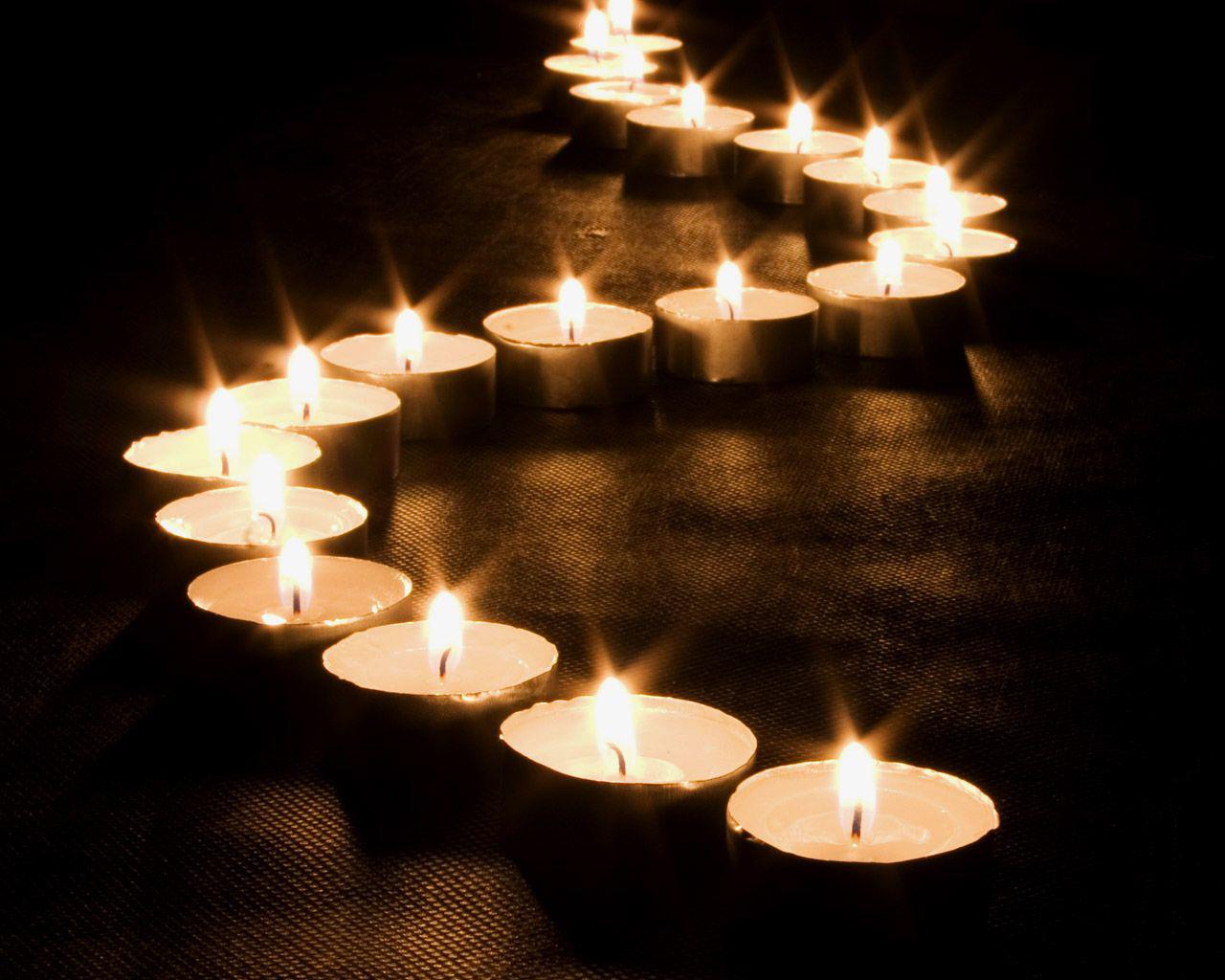}
		\includegraphics[width=0.35in,height=0.35in]{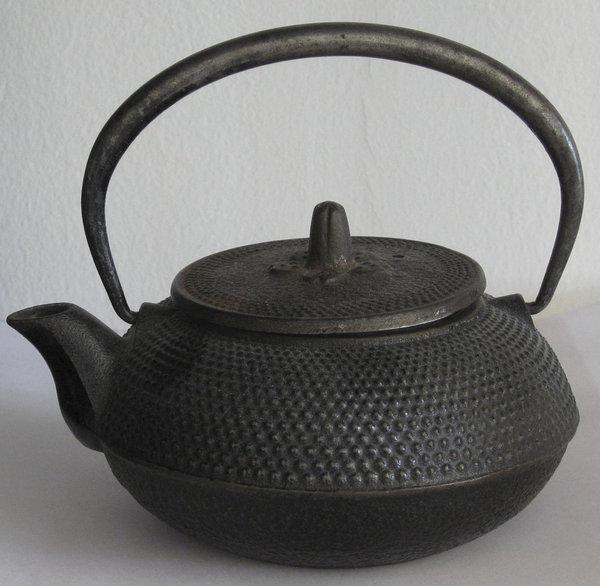}
		\includegraphics[width=0.35in,height=0.35in]{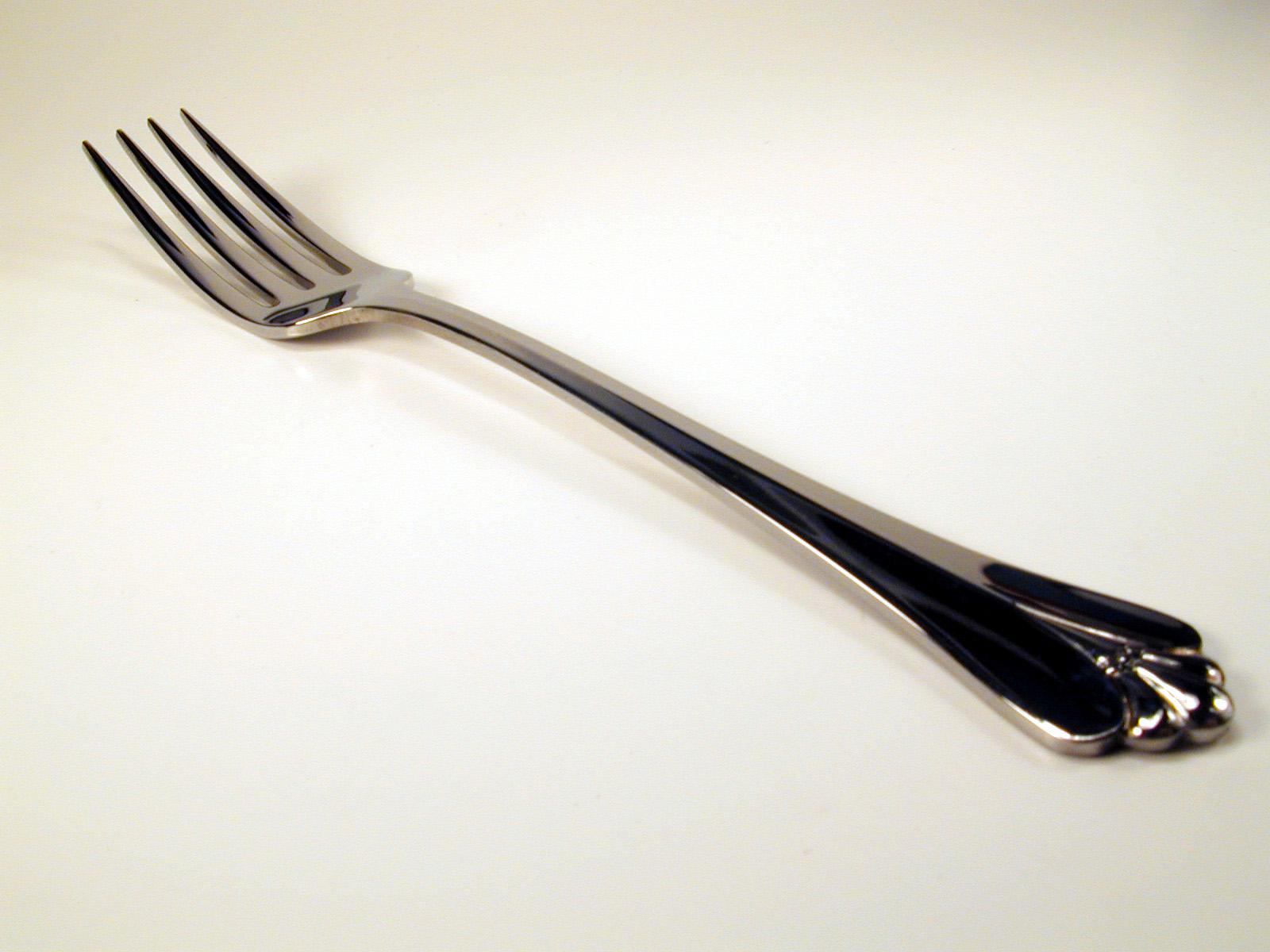}
		\includegraphics[width=0.35in,height=0.35in]{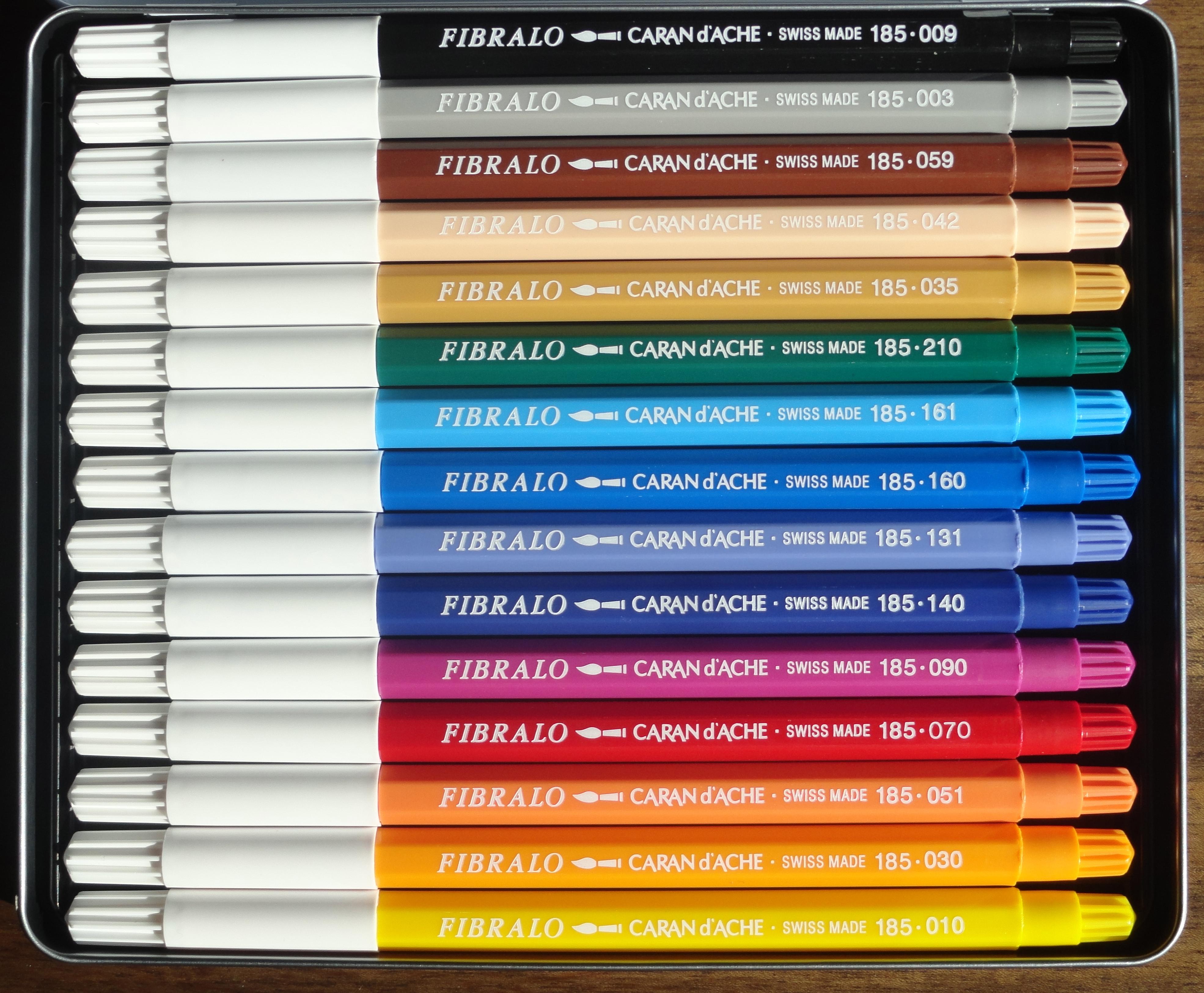}
	}
	\caption{Examples of images from different domains. In panel (a), images of the top row are from Amazon, images in the middle belong to domain Webcam, and the bottom row are images from domain DSLR. Examples of images from Office-Home dataset are shown in panel (b). The images are from Artistic, Clipart, Product, and Real World domain respectively from top to the bottom.}
	\label{fig:office}
\end{figure}

\begin{figure*}
	\begin{center}
		\subfigure[Initial source($A$) features]{
			\includegraphics[width=1.5in,height=1.4in]{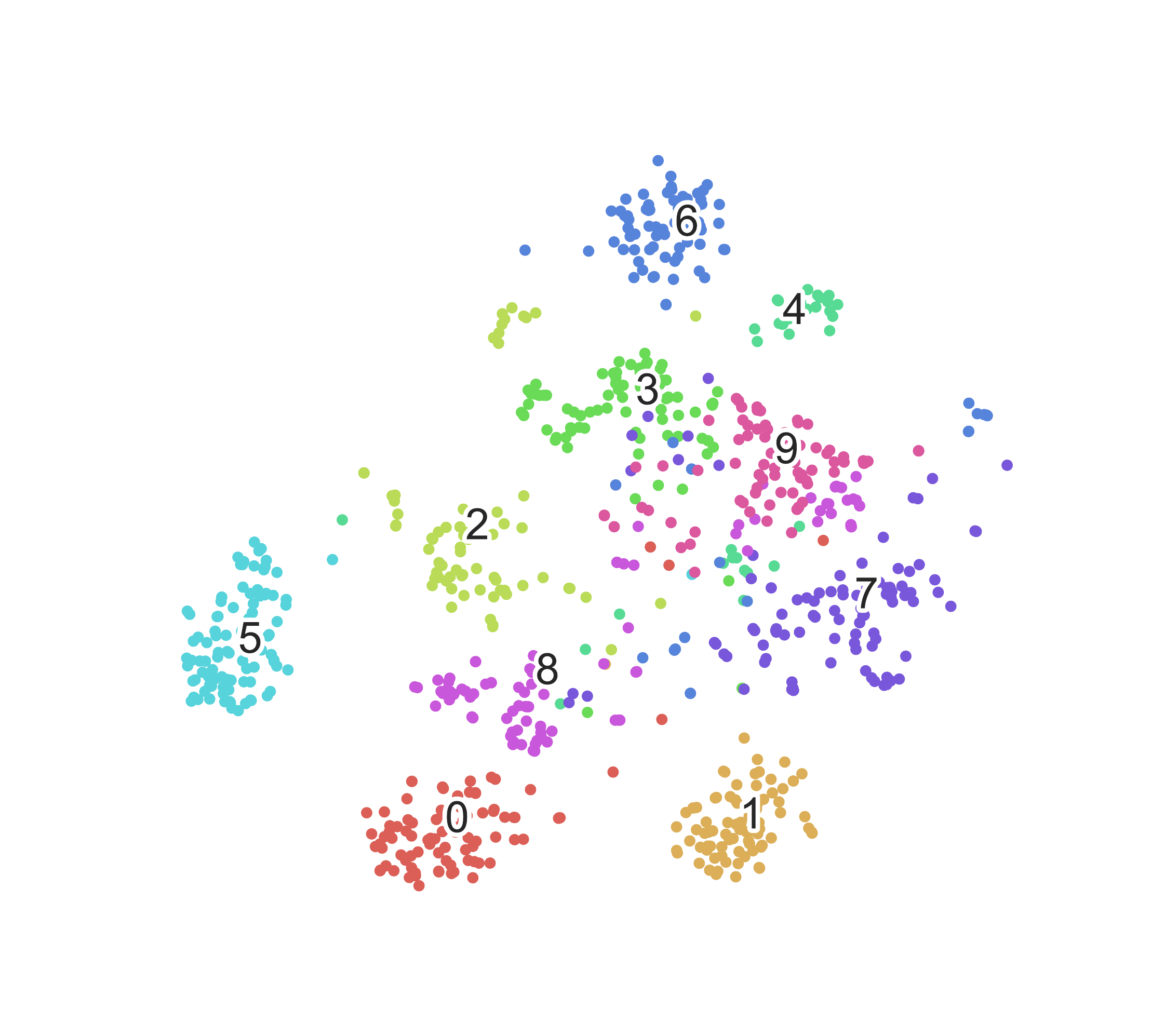}
		}	
		\subfigure[Initial target($D$) features]{
			\includegraphics[width=1.5in,height=1.4in]{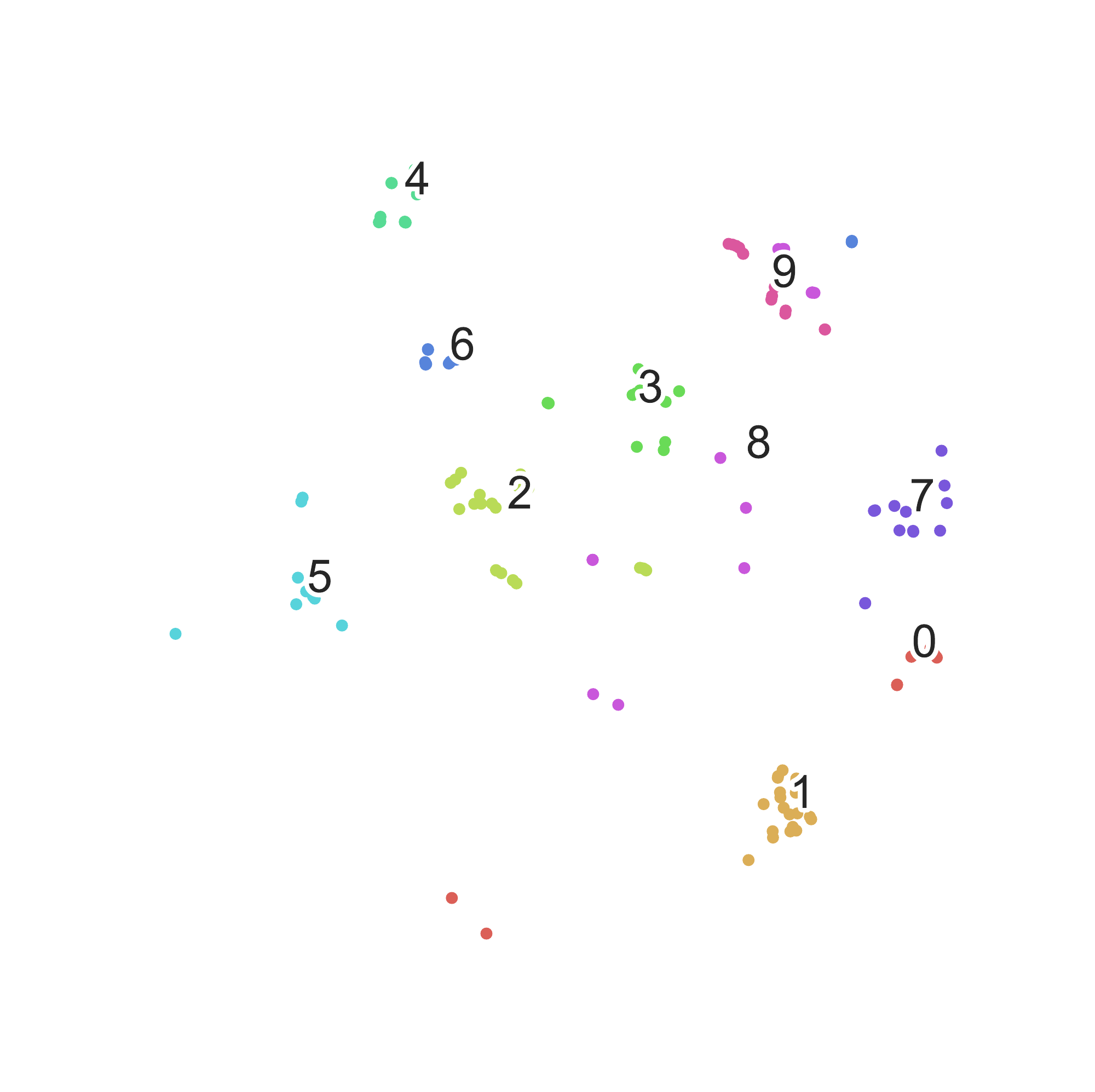}
		}
		\subfigure[Final source($A$) features]{
			\includegraphics[width=1.5in,height=1.4in]{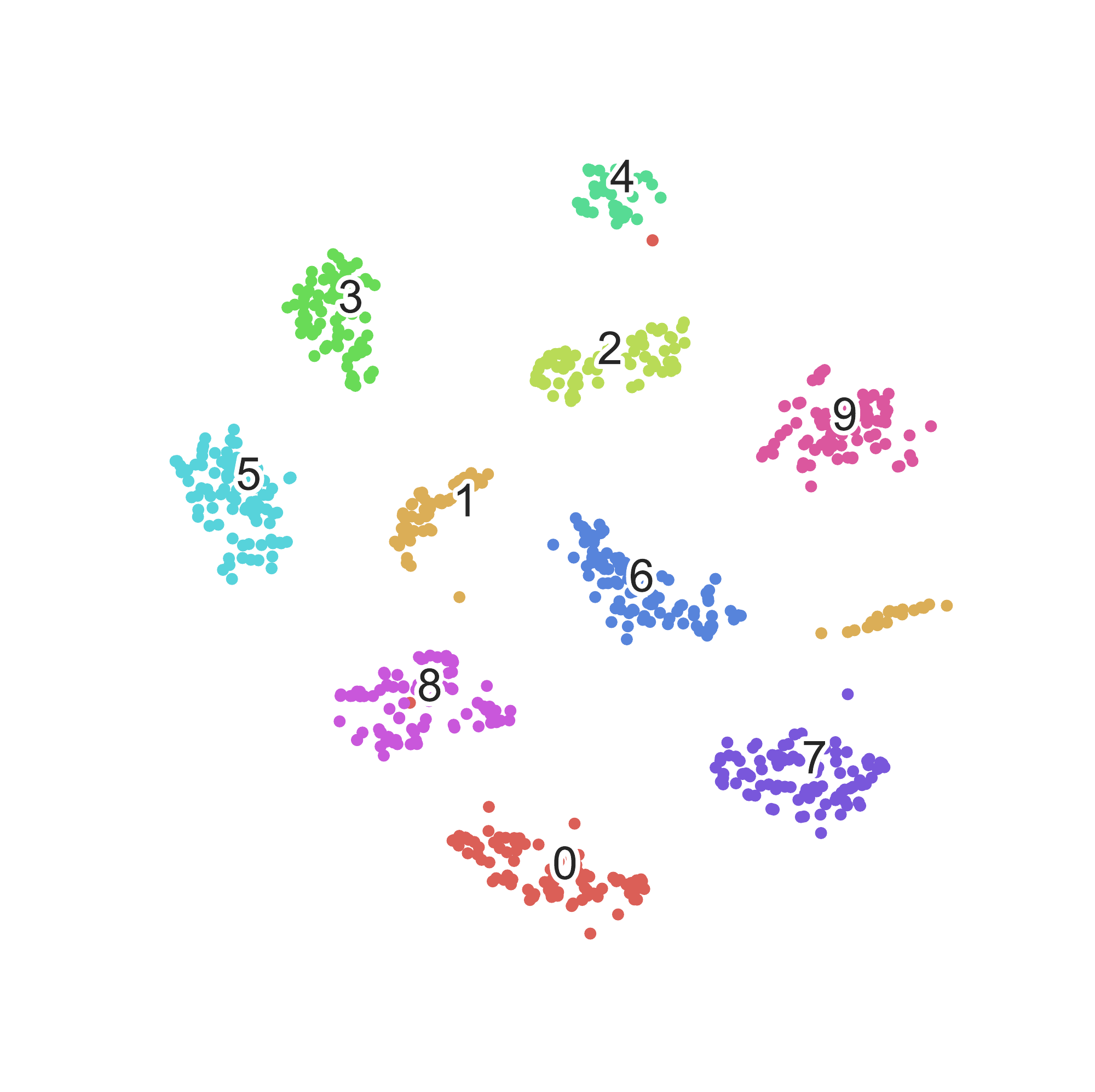}
		}	
		\subfigure[Final target($D$) features]{
			\includegraphics[width=1.5in,height=1.4in]{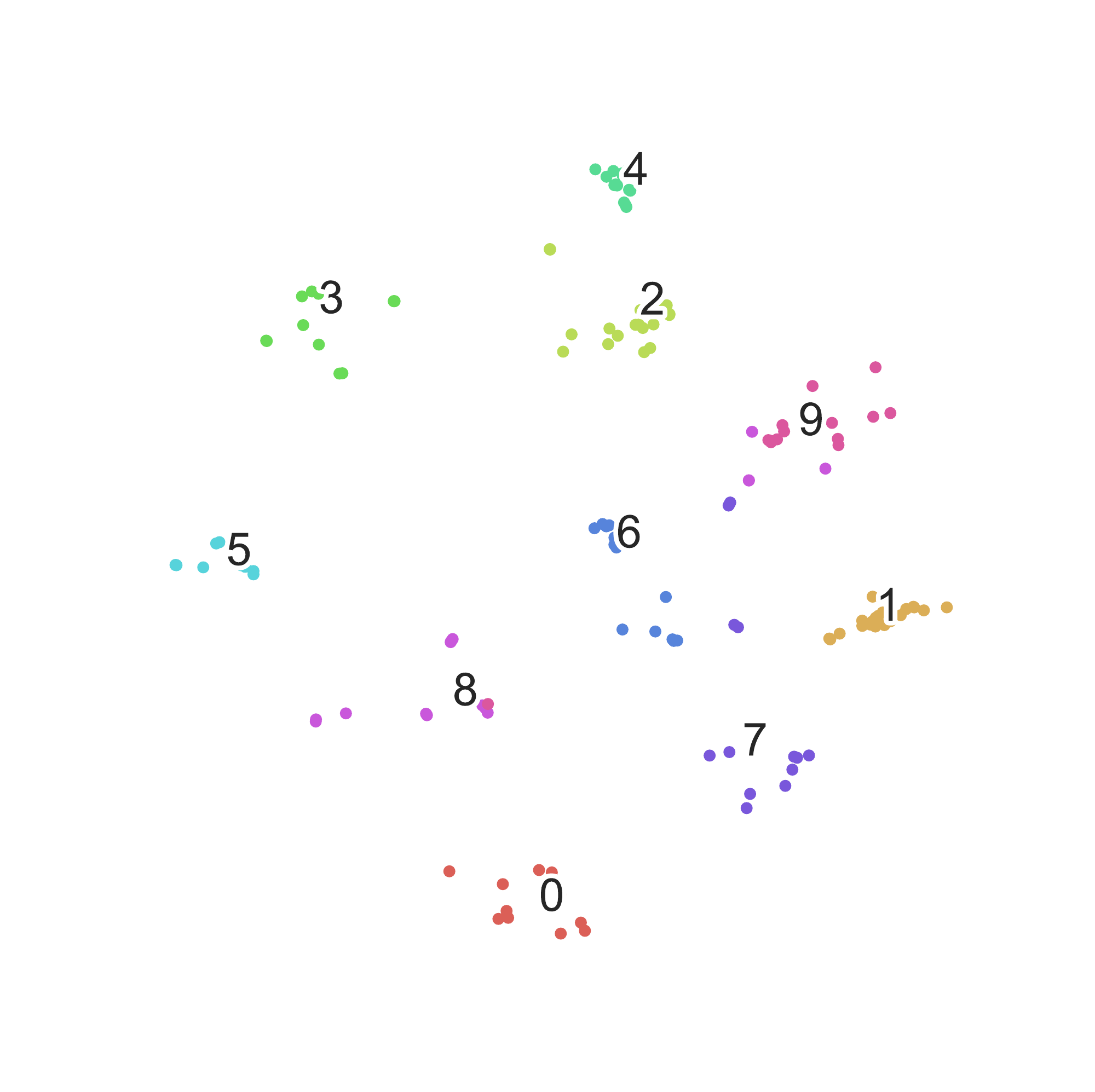}
		} \\	
		\subfigure[Initial source($W$) features]{
			\includegraphics[width=1.5in,height=1.4in]{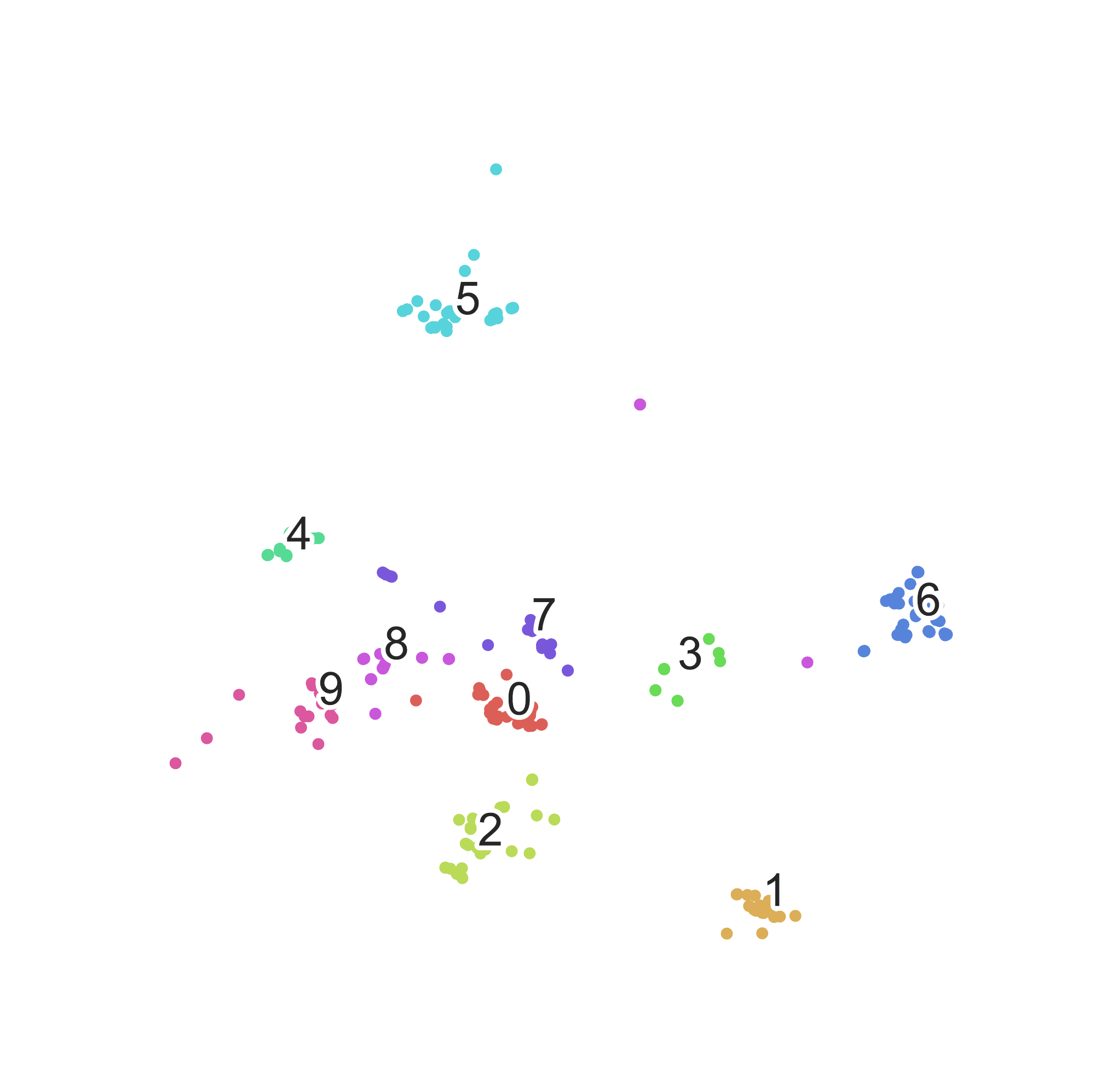}
		}	
		\subfigure[Initial target($A$) features]{
			\includegraphics[width=1.5in,height=1.4in]{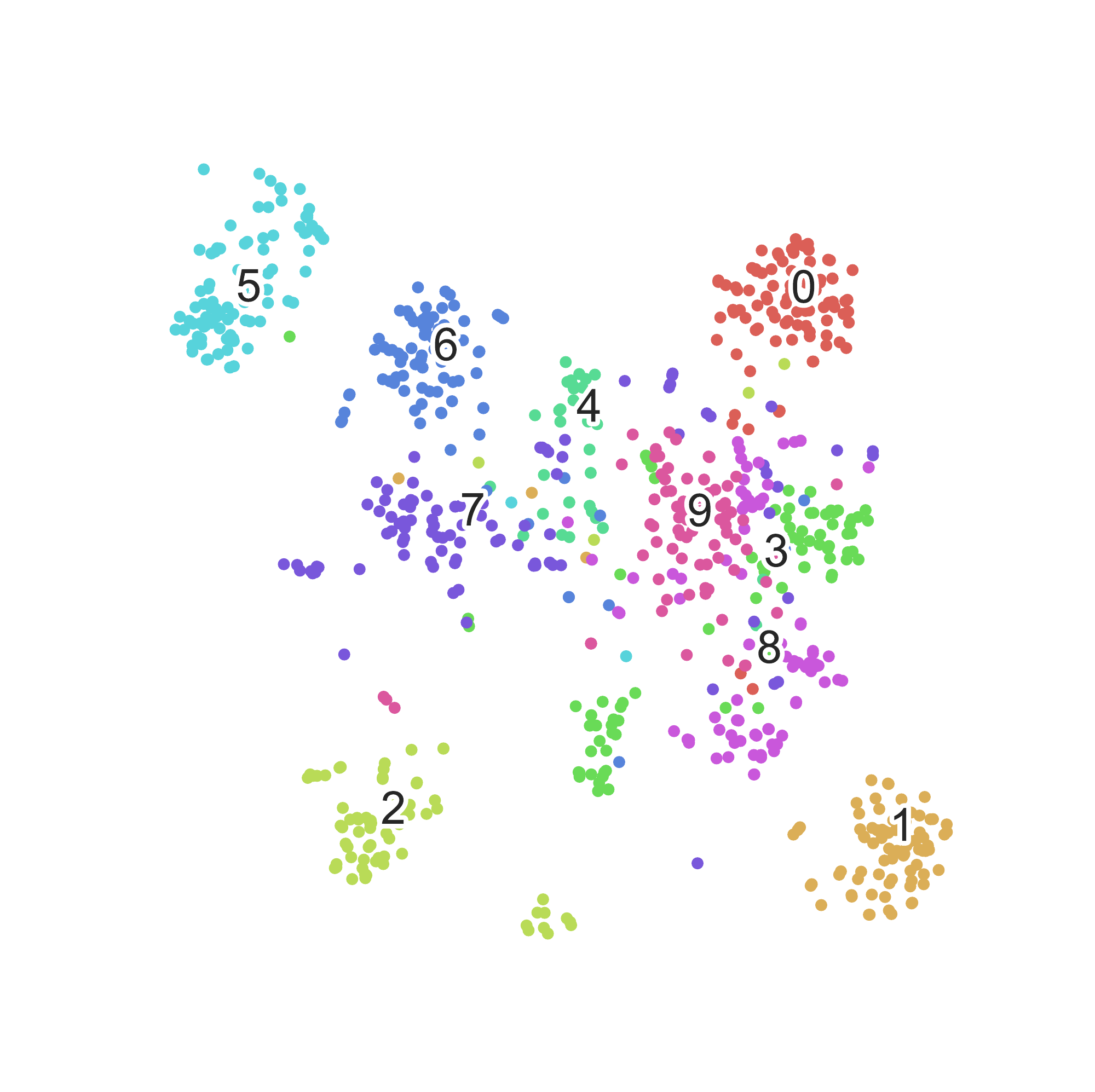}
		}
		\subfigure[Final source($W$) features]{
			\includegraphics[width=1.5in,height=1.4in]{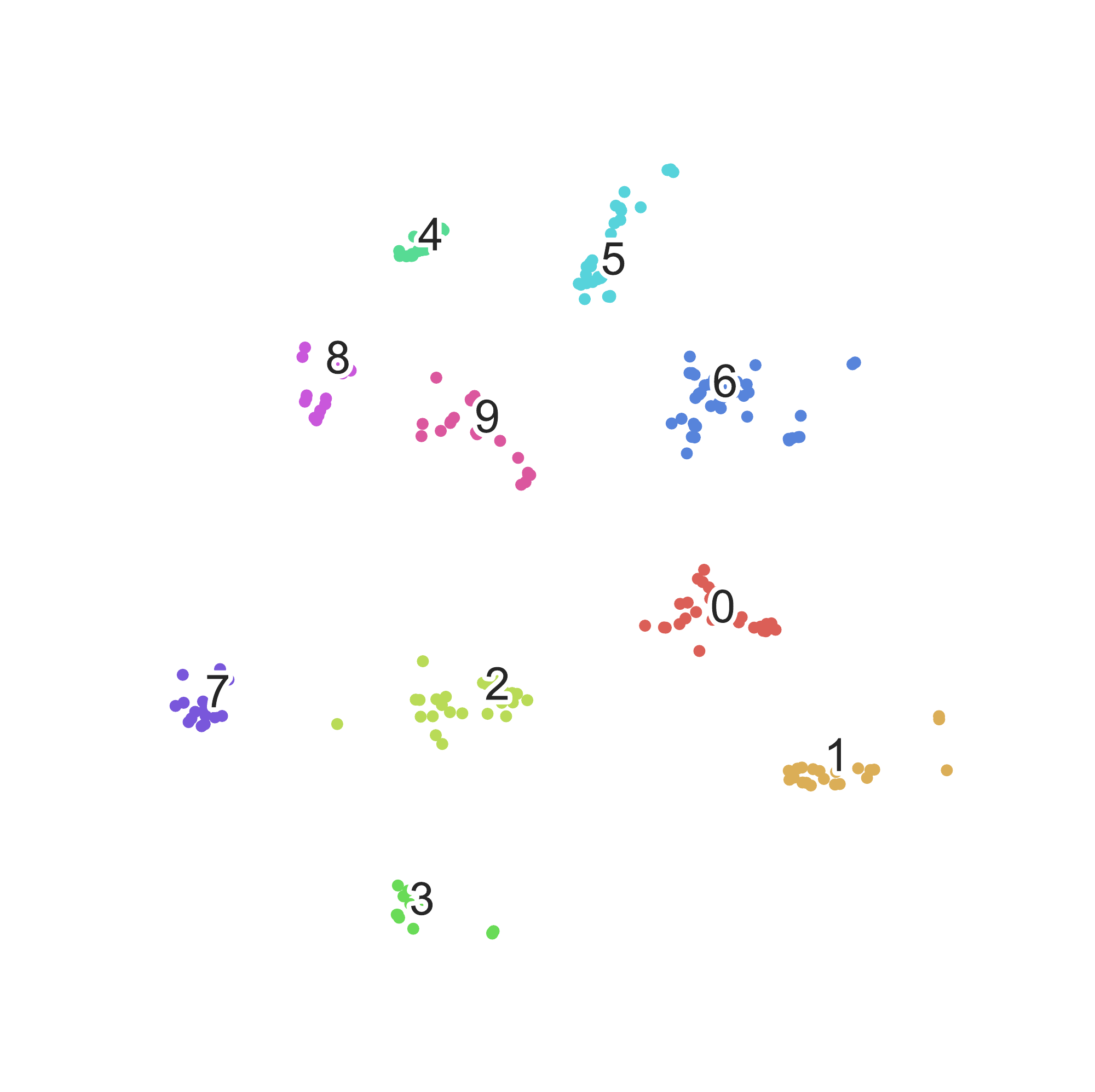}
		}	
		\subfigure[Final target($A$) features]{
			\includegraphics[width=1.5in,height=1.4in]{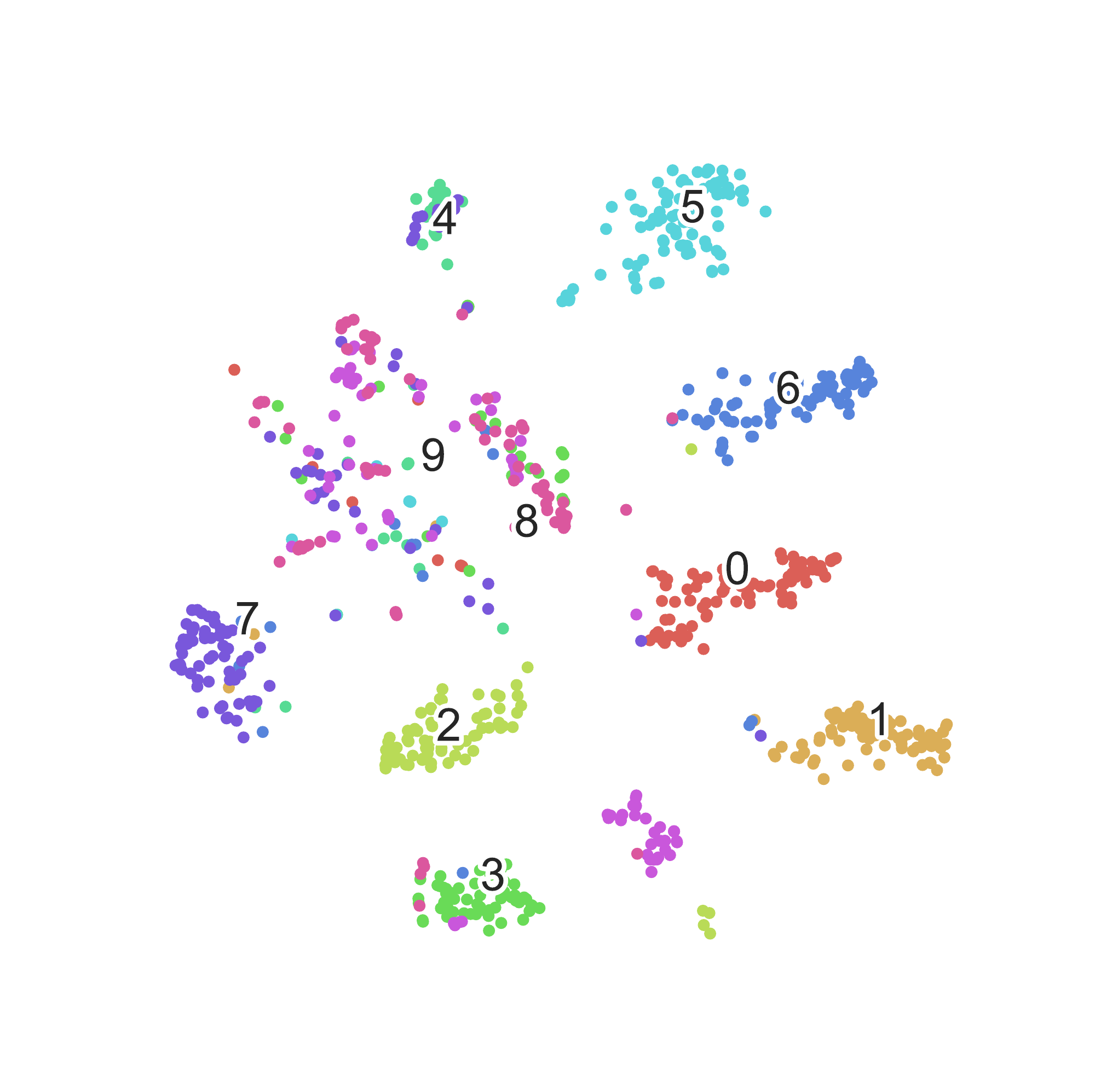}
		} \\
		\subfigure[Initial source($Pr$) features]{
			\includegraphics[width=1.5in,height=1.4in]{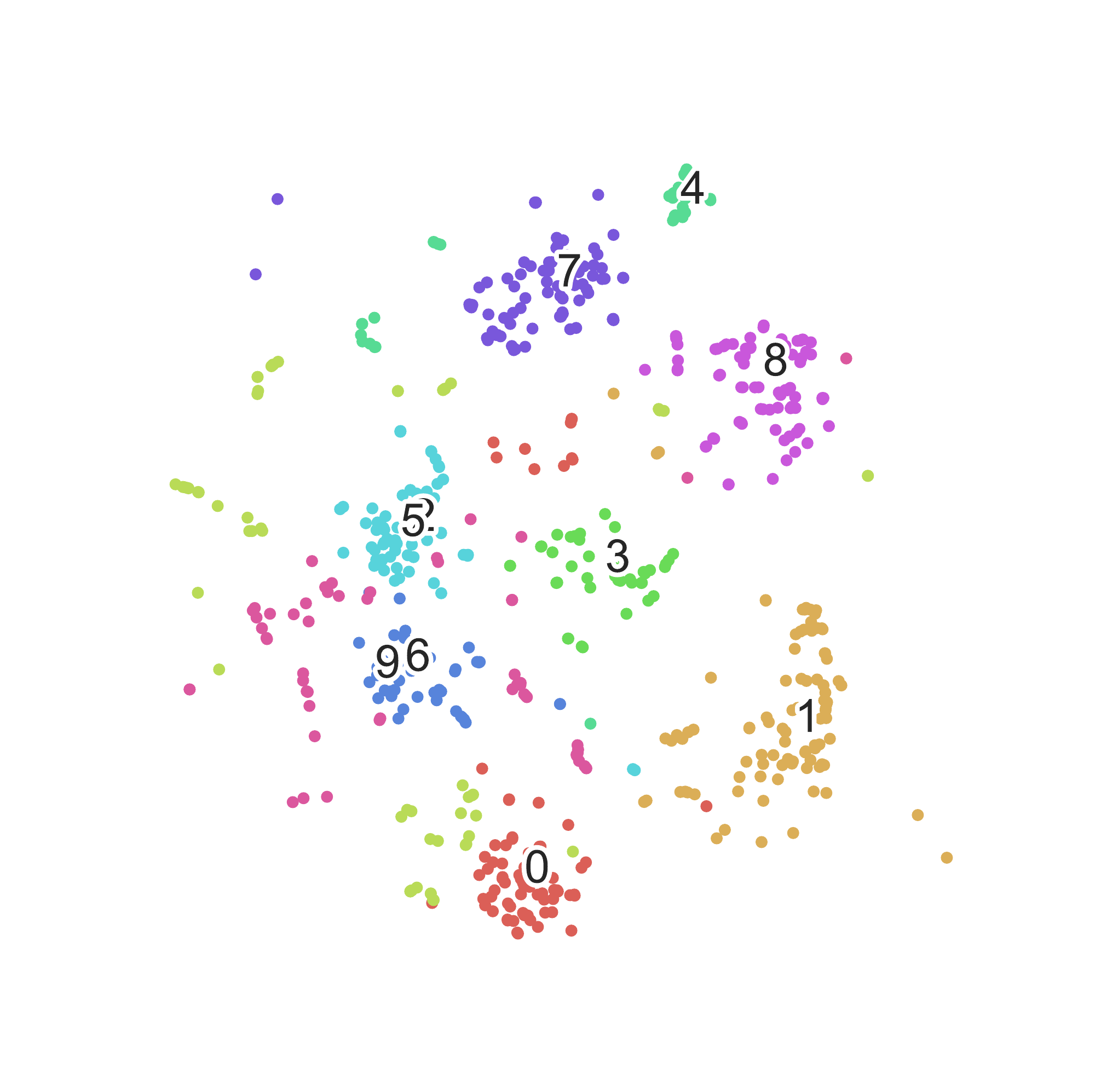}
		}	
		\subfigure[Initial target($Rw$) features]{
			\includegraphics[width=1.5in,height=1.4in]{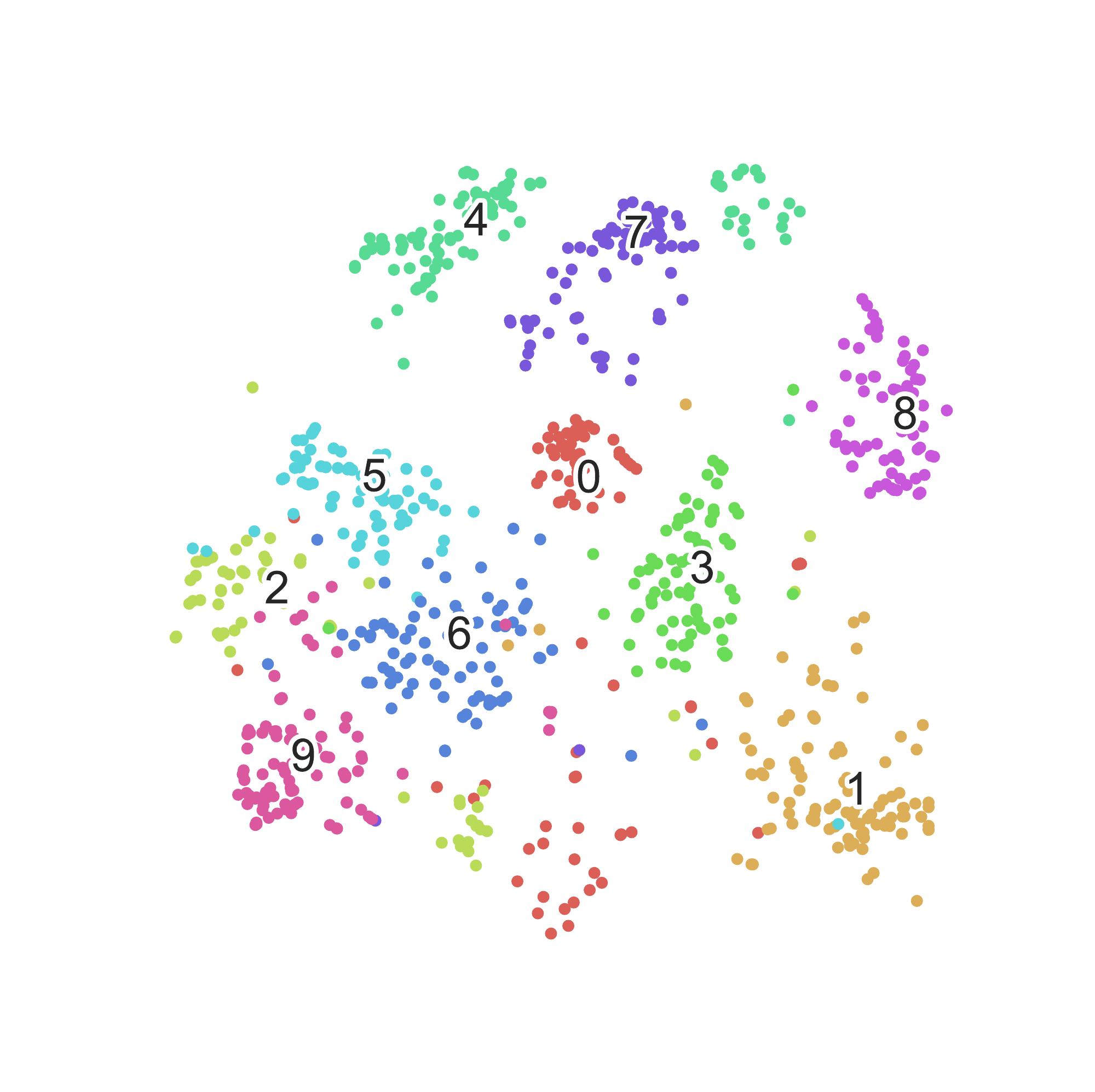}
		}
		\subfigure[Final source($Pr$) features]{
			\includegraphics[width=1.5in,height=1.4in]{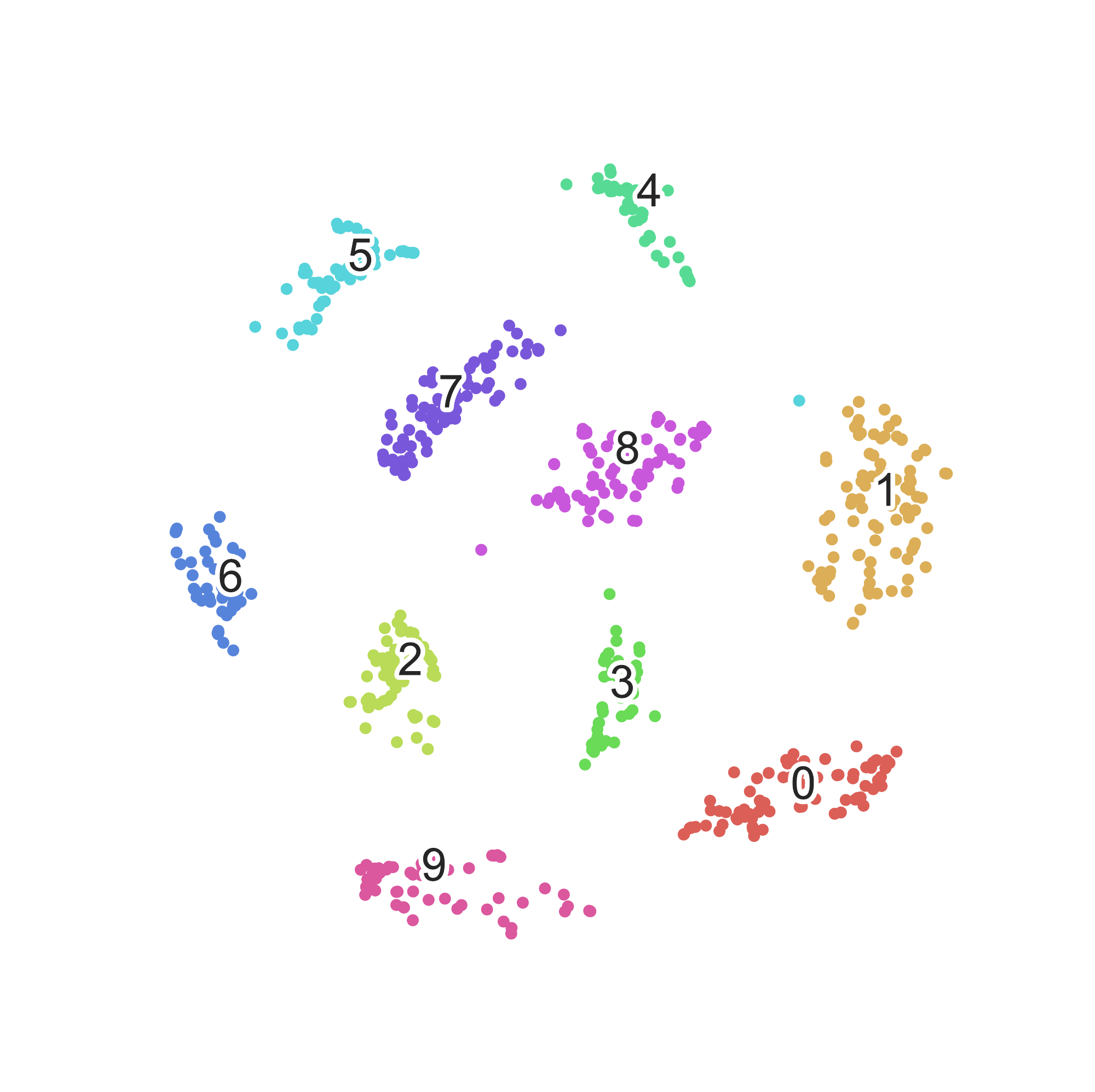}
		}	
		\subfigure[Final target($Rw$) features]{
			\includegraphics[width=1.5in,height=1.4in]{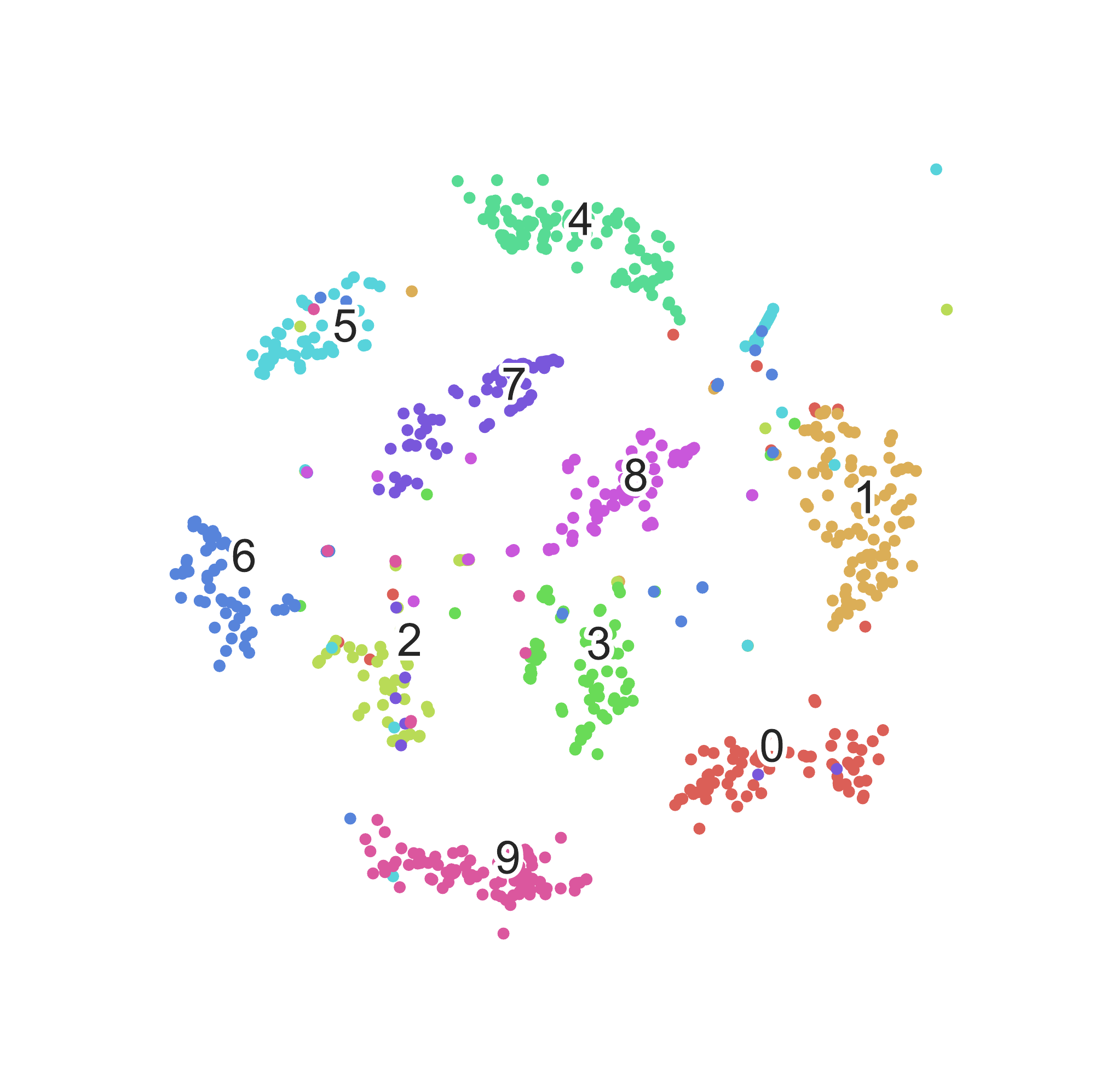}
		} \\
		\subfigure[Initial source($Ar$) features]{
			\includegraphics[width=1.5in,height=1.4in]{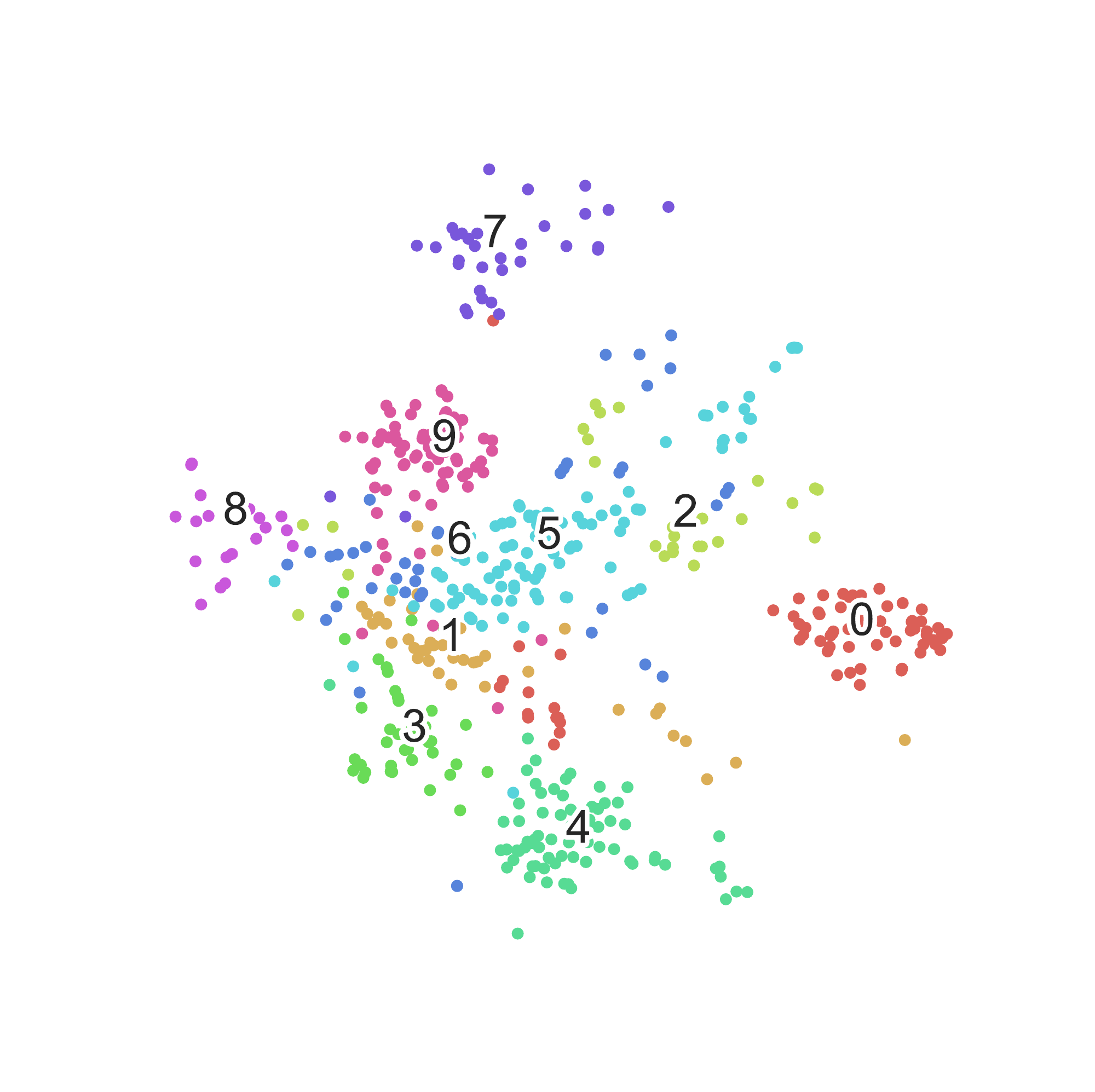}
		}	
		\subfigure[Initial target($Pr$) features]{
			\includegraphics[width=1.5in,height=1.4in]{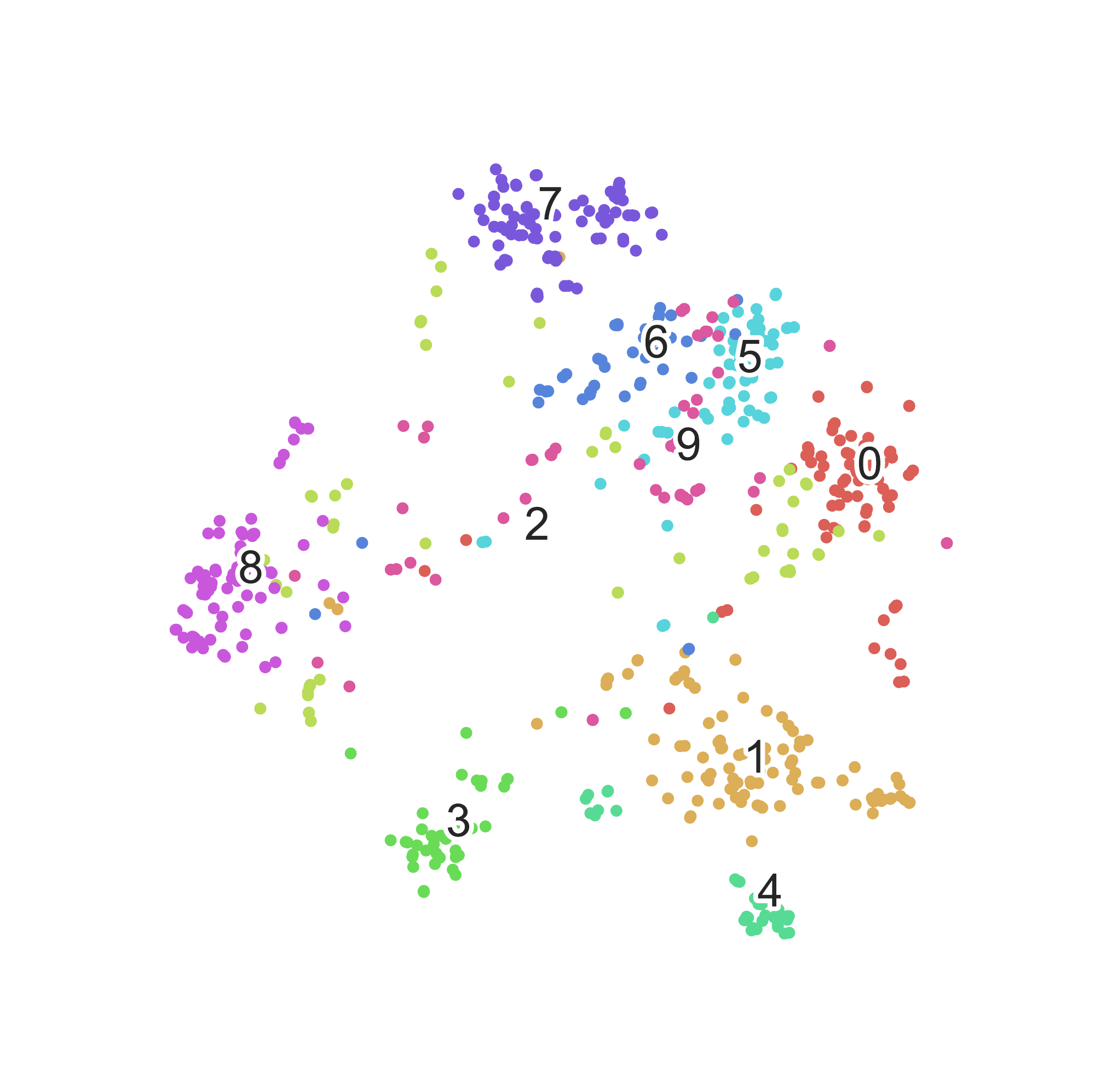}
		}
		\subfigure[Final source($Ar$) features]{
			\includegraphics[width=1.5in,height=1.4in]{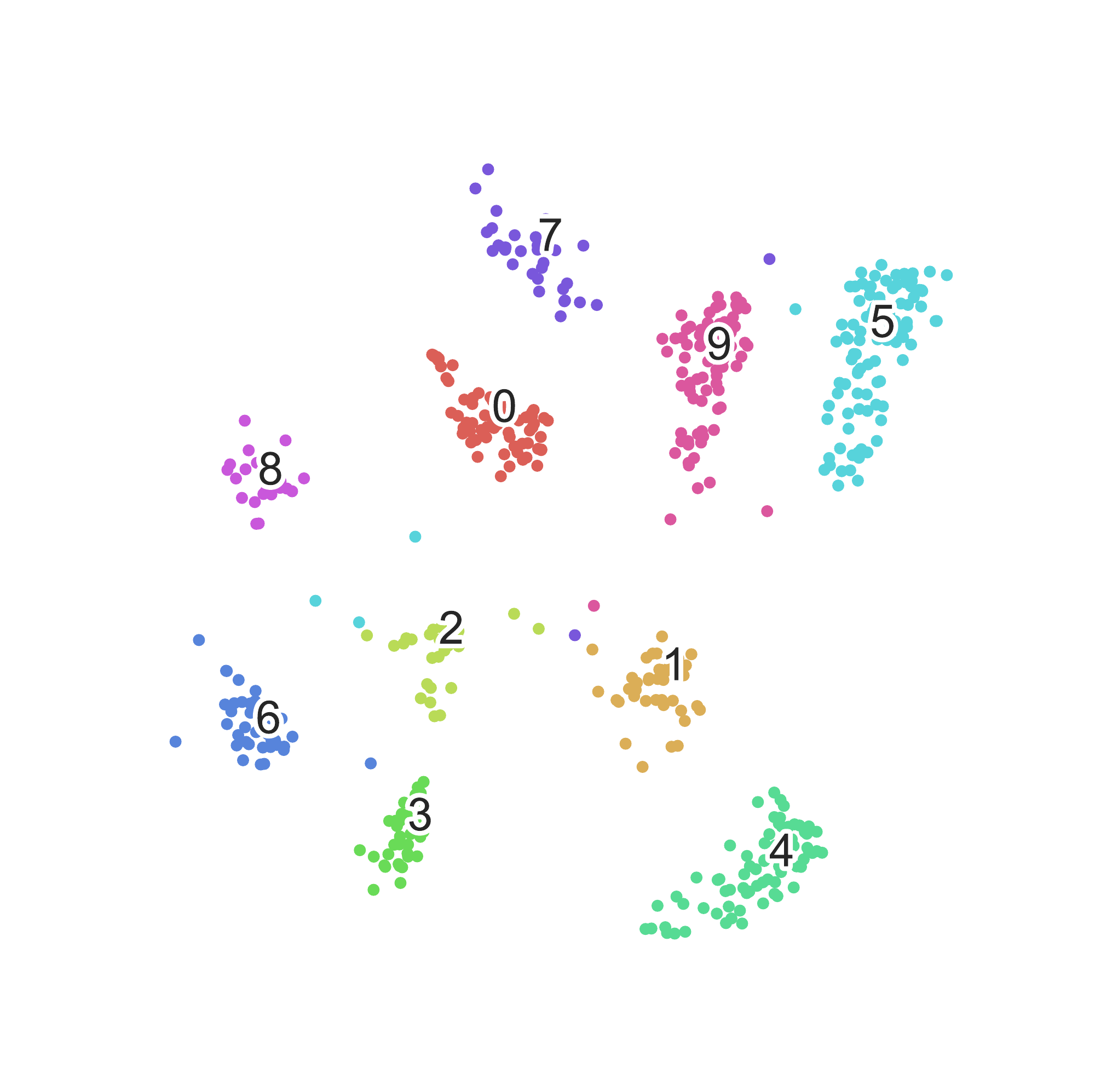}
		}	
		\subfigure[Final target($Pr$) features]{
			\includegraphics[width=1.5in,height=1.4in]{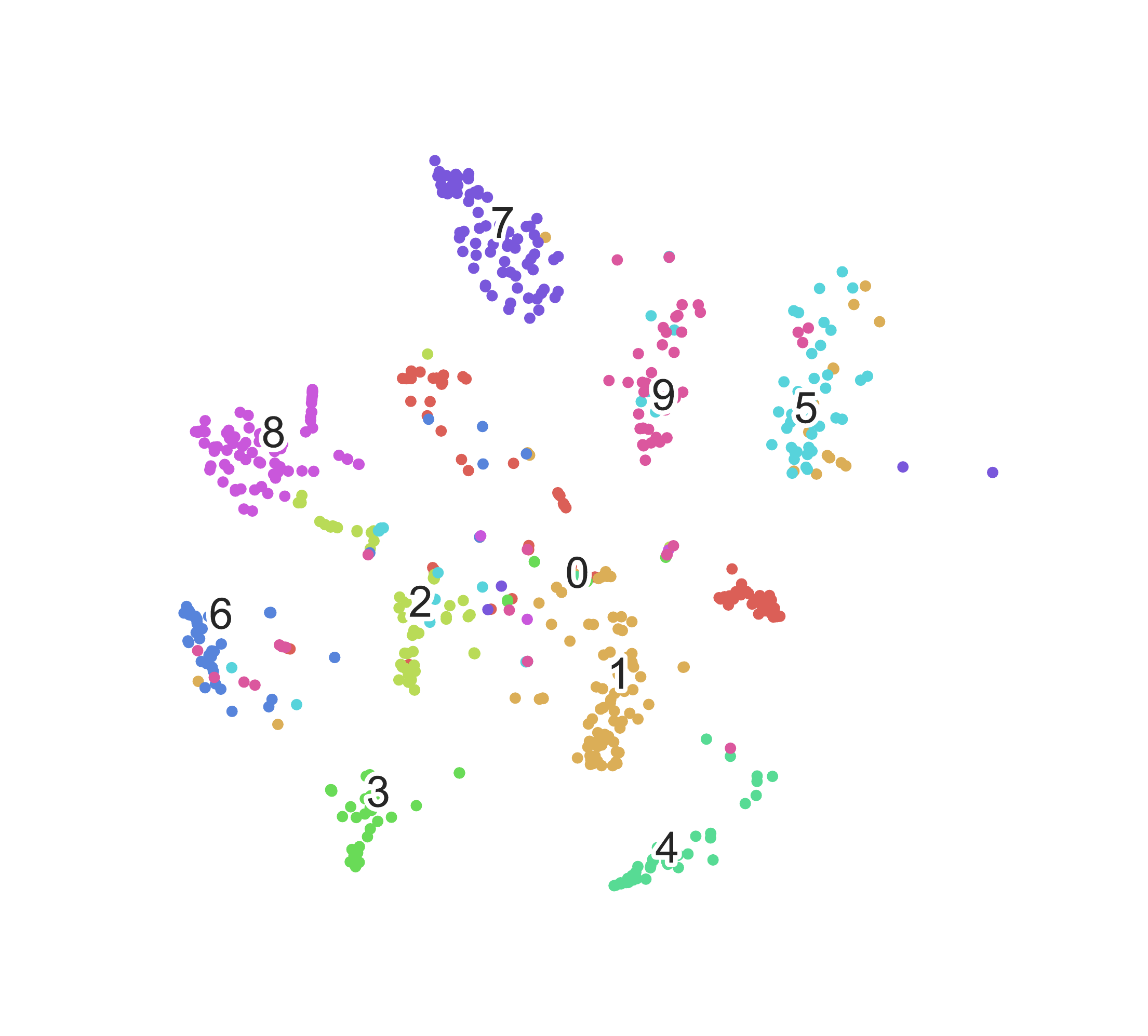}
		} \\
	\end{center}
	\caption{t-SNE embeddings of tasks $A\rightarrow D$, $W\rightarrow A$, $Pr\rightarrow Rw$, and $Ar\rightarrow Pr$ respectively from top to bottom. The first column plots t-SNE embeddings of source deep features extracted by models finetuned on AlexNet. The second column plots t-SNE embeddings of target deep features extracted by models finetuned on AlexNet. The right two columns are t-SNE embeddings of source and target features extracted by the model learned by the proposed approach respectively.}\label{fig:t-SNE}
\end{figure*}

\section{Experiments}
In this section, we firstly introduce the datasets, network architecture, and training process of the experiments. Then, the proposed method is evaluated on classification tasks and the experimental results are compared with those obtained by some state-of-the-art methods in several well-known domain adaptation datasets. Finally, some qualitative evaluations are given.

\subsection{Experimental Setup}
\subsubsection{Datasets}
The \textbf{Office-31} dataset~\cite{saenko2010adapting} contains images originated from three domains: Amazon, Webcam, and DSLR. These three domains consist of the same 31 categories. Amazon(A) contains images downloaded from online merchants\footnote{www.amazon.com}. These images are product shots at medium resolution typically taken in an environment with studio lighting conditions without redundant background. DSLR(D) consists of images that were captured with a digital SLR camera in realistic environments with natural lighting conditions. The images have high resolution and low noise. Webcam(W) consists of images from a webcam. These images are of low resolution and show significant noise, as well as white balance artifacts. These three domains in Office-31 database represents several interesting visual domain shifts as shown in Figure~\ref{fig:office}(a). Using Office-31, we can evaluate the proposed method on 6 transfer tasks: $W\rightarrow D$, $W\rightarrow A$, $D\rightarrow W$, $D\rightarrow A$, $A\rightarrow W$, and $A\rightarrow D$.

The second dataset is the \textbf{Office-Home}\footnote{http://hemanthdv.org/OfficeHome-Dataset/}~\cite{venkateswara2017deep} dataset, consisting of 4 significantly different domains of everyday objects in office and home settings: Artistic images (Ar), Clip Art (Cl), Product images (Pr) and Real-World (Rw). There are 65 categories in each domain and more than 15,000 images in total. Compared with Office-31, it is a more challenging dataset for domain adaptation evaluation because each domain in this dataset contains more categories and images in each category have significant domain shifts visually, as shown in Figure~\ref{fig:office}(b). For this dataset, 12 transfer tasks can be generated for evaluation using all its 4 domains: $Cl\rightarrow Pr$, $Cl\rightarrow Rw$, $Cl\rightarrow Ar$, $Pr\rightarrow Cl$, $Pr\rightarrow Rw$, $Pr\rightarrow Ar$, $Rw\rightarrow Cl$, $Rw\rightarrow Pr$, $Rw\rightarrow Ar$, $Ar\rightarrow Cl$, $Ar\rightarrow Pr$, and $Ar\rightarrow Rw$.

\subsubsection{Network Architecture}
The network in our experiments was built based on the architecture of AlexNet~\cite{krizhevsky2012imagenet}. It contains eight layers with weights. The first five layers are convolutional layers and the last three layers are fully-connected. This architecture requires constant size of inputs, so all the images are rescaled to $227\times 227\times 3$-dimension before being fed as inputs. The number of neurons in the fully-connected (fc) layer fc6 and fc7 are all 4096, and in fc8 it is equal to the number of categories in the dataset. To be fairly compared with other adversarial methods, a bottleneck layer with size 256 is added between fc7 and fc8. The tensor products of the $softmax$ probabilities and the $bottleneck$ layer the outputs are utilized as the inputs of the discriminator. The discriminator used in our experiments consists of three fully connected layers. The size of the first two layers are 1024 followed by ReLU activation layer and dropout layer while the dimension of final outputs is 1.

\begin{table*}
	\centering
	\caption{Comparison of accuracy on Office-31 dataset under conventional unsupervised domain adaptation setting.~\cite{long2018conditional}}
	\small
	\begin{tabular}{p{3.5cm}<{\centering}|p{1.5cm}<{\centering}p{1.5cm}<{\centering}p{1.5cm}<{\centering}p{1.5cm}<{\centering}p{1.5cm}<{\centering}p{1.5cm}<{\centering}p{1.5cm}<{\centering}}
		\hlinew{1.5pt}
		Method & W$\rightarrow$D & W$\rightarrow$A & D$\rightarrow$W & D$\rightarrow$A & A$\rightarrow$W & A$\rightarrow$D & Avg. \\
		\hline
		TCA~\cite{pan2011domain} & 95.2 & 50.9 & 93.2 & 51.6 & 61.0 & 60.8 & 68.8 \\
		GFK~\cite{gong2012geodesic} & 95.0 & 48.1 & 95.6 & 52.4 & 60.4 & 60.6 & 68.7 \\
		AlexNet~\cite{krizhevsky2012imagenet} & 99.0 & 49.8 & 95.1 & 51.1 & 61.6 & 63.8 & 70.1 \\
		RTN~\cite{long2016deep} & 99.6 & 51.0 & 96.8 & 50.5 & 73.3 & 71.0 & 73.7 \\
		DDC~\cite{tzeng2014deep} & 98.5 & 52.2 & 95.0 & 52.1 & 61.8 & 64.4 & 70.7 \\
		DAN~\cite{long2015learning} & 99.0 & 53.1 & 96.0 & 54.0 & 68.5 & 67.0 & 72.9 \\
		JAN~\cite{long2016deep} & 99.5 & 55.0 & 96.6 & 58.3 & 74.9 & 71.8 & 76.0 \\
		ADDA~\cite{tzeng2017adversarial} & 98.8 & 53.5 & 96.2 & 54.6 & 73.5 & 71.6 & 74.7 \\
		DANN~\cite{ganin2016domain} & 99.2 & 51.2 & 96.4 & 53.4 & 73.0 & 72.3 & 74.3 \\
		MJKD~\cite{mao2018deep} & 99.8 & 56.4 & 97.0 & \underline{59.3} & 77.0 & 73.7 & 77.2 \\
		CDAN~\cite{long2018conditional} & \textbf{100} & \underline{57.3} & \underline{97.2} & 57.3 & \textbf{78.3} & \textbf{76.3} & \underline{77.7} \\
		\hline
		Ours & \textbf{100} & \textbf{61.5} & \textbf{98.3} & \textbf{62.4} & \underline{78.2} & \underline{76.0} & \textbf{79.4} \\
		\hlinew{1.5pt}
	\end{tabular}
	\label{tab:fc7_com}
\end{table*}

\begin{table*}[t]
	\setlength{\belowcaptionskip}{0.5cm}
	\centering
	\caption{Comparison of accuracy on Office-Home dataset under conventional unsupervised domain adaptation setting.~\cite{long2018conditional}}%~\cite{long2015learning, long2016deep}
	\small
	\begin{tabular}{p{3.5cm}<{\centering}|p{1.5cm}<{\centering}p{1.5cm}<{\centering}p{1.5cm}<{\centering}p{1.5cm}<{\centering}<{\centering}p{1.5cm}<{\centering}p{1.5cm}<{\centering}p{1.5cm}<{\centering}}
		\hlinew{1.5pt}
		Method & Cl$\rightarrow$Pr & Cl$\rightarrow$Rw & Cl$\rightarrow$Ar & Pr$\rightarrow$Cl & Pr$\rightarrow$Rw &  Pr$\rightarrow$Ar & Avg. \\
		\hline
		AlexNet~\cite{krizhevsky2012imagenet} & 41.7 & 42.1 & 22.1 & 20.3 & 51.1 & 20.5 & 32.9 \\
		DAN~\cite{long2015learning} & 48.6 & 50.8 & 33.8 & 35.1 & 57.7 & 30.1 & 42.7 \\
		DANN~\cite{ganin2015unsupervised} & 51.8 & 55.1 & 35.2 & 39.7 & 59.3 & 31.6 & 45.5 \\
		JAN~\cite{long2016deep} & 53.3 & 54.5 & 36.4 & 40.3 & 60.1 & 33.4 & 46.3 \\				
		CDAN~\cite{long2018conditional} & \underline{56.4} & \underline{57.8} & \textbf{39.7} & \textbf{43.1} & \underline{63.2} & \underline{35.5} & \underline{49.3} \\
		\hline
		Ours & \textbf{59.2} & \textbf{60.5} & \underline{37.8} & \underline{42.5} & \textbf{63.8} & \textbf{36.3} & \textbf{50.0} \\
		\hline
		\hline
		Method & Rw$\rightarrow$Cl & Rw$\rightarrow$Pr & Rw$\rightarrow$Ar & Ar$\rightarrow$Cl & Ar$\rightarrow$Pr & Ar$\rightarrow$Rw & Avg. \\
		\hline
		AlexNet~\cite{krizhevsky2012imagenet} & 27.9 & 54.9 & 31.0 & 26.4 & 32.6 & 41.3 & 35.7 \\
		DAN~\cite{long2015learning} & 39.3 & 63.7 & 44.6 & 31.7 & 43.2 & 55.1 & 46.3 \\
		DANN~\cite{ganin2015unsupervised} & 46.4 & 65.9 & 45.7 & 36.4 & 45.2 & 54.7 & 49.1 \\
		JAN~\cite{long2016deep} & 47.4 & 67.9 & 45.9 & 35.5 & 35.5 & 46.1 & 51.8 \\				
		CDAN~\cite{long2018conditional} & \underline{48.5} & \underline{71.1} & \underline{48.4} & \underline{38.1} & \underline{50.3} & \underline{60.3} & \underline{52.9} \\
		\hline
		Ours & \textbf{49.5} & \textbf{71.8} & \textbf{49.0} & \textbf{42.2} & \textbf{57.1} & \textbf{62.8} & \textbf{55.4} \\
		\hlinew{1.5pt}
	\end{tabular}
	\label{tab:clef}
\end{table*}

\subsubsection{Training Process} 
Models in our experiments were trained using the framework \textbf{Caffe}~\cite{jia2014caffe}. The initial model $\mathcal{M}_0$ was fine-tuned on the source domain using AlexNet pre-trained on ImageNet. Following the standard fine-tuning procedure, the first three convolutional layers were frozen. The learning rates for $conv4-fc7$ were set to a small number to slightly tune the parameters initialized from the pre-trained model. The learning rate for other layers like $bottleneck$ and $fc8$ can be set larger, typically 10 times that of $conv1-fc7$. In order to take each domain into consideration with equal significance in the supervised training part, namely the classifier, labeled data in each mini-batch were half chosen from source domain while the other half were chosen from the proposed select-and-adapt labeled target data $\mathcal{T}_p$ with their pseudo-labels. For instance, in our experiments, the mini-batch size of the labeled data were set to be 64, then 32 of these data were selected from source domain and the other 32 were from $\mathcal{T}^p$. We used the stochastic gradient descent (SGD) update strategy with a momentum of 0.9. The learning rate was initialized from the range 0.0001 to 0.001, and was changed by the following strategy: $base_{lr}\times (1+\gamma \times iter)^{-power}$, where $power$ was set to 0.85 throughout all experiments and $iter$ is the current number of iterations. To make the proposed approach able to directly compare with the other methods, we follow the unsupervised domain adaptation protocol, in which all labeled source domain samples are used for training the initial network, while labels of all target examples are not provided~\cite{long2015learning} during the training process.

\begin{figure}[ht]
	\begin{center}
		\subfigure[$W\rightarrow A$]{
			\includegraphics[width=0.45\linewidth]{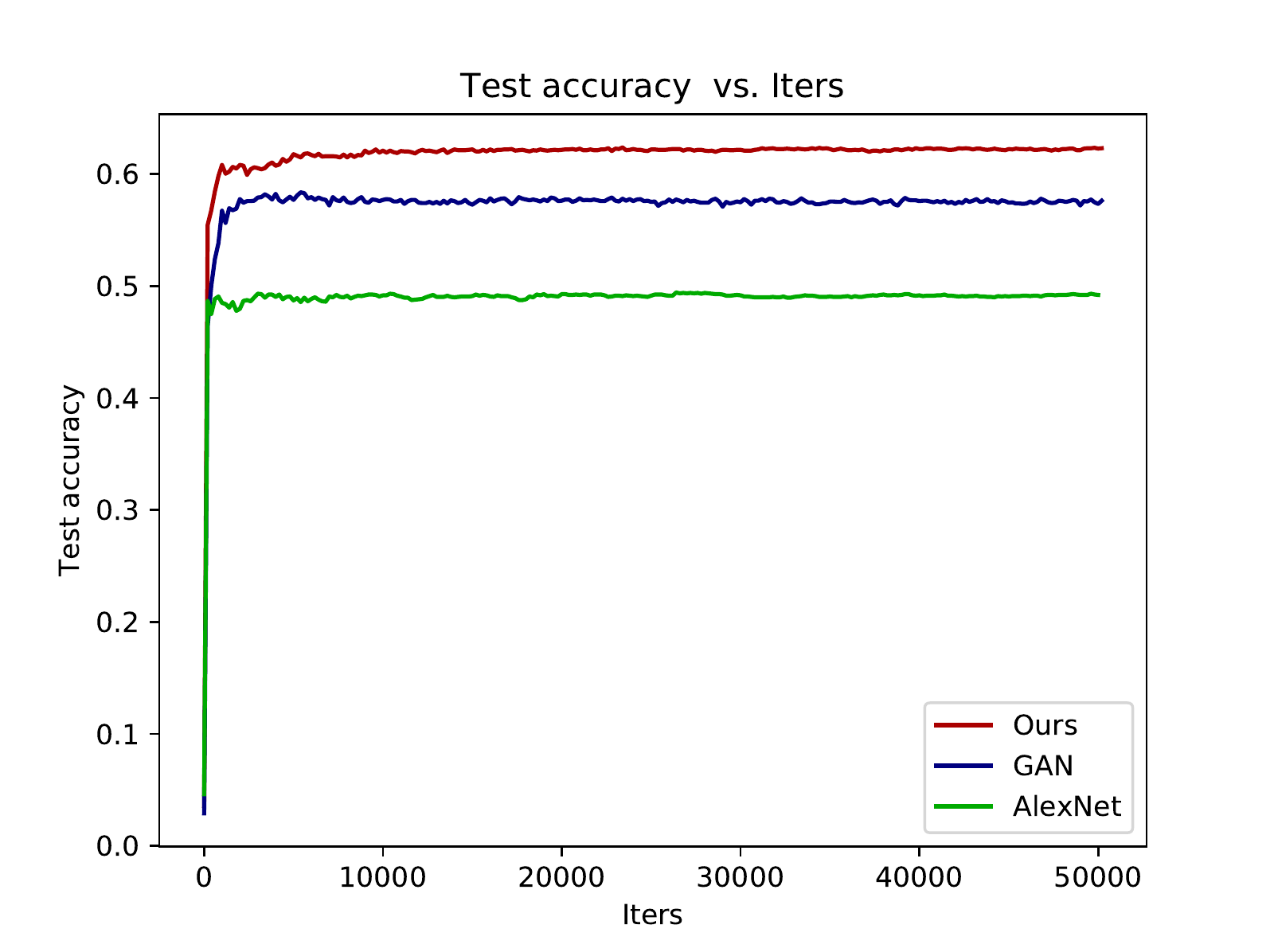}	
		}
		\subfigure[$D\rightarrow W$]{
			\includegraphics[width=0.45\linewidth]{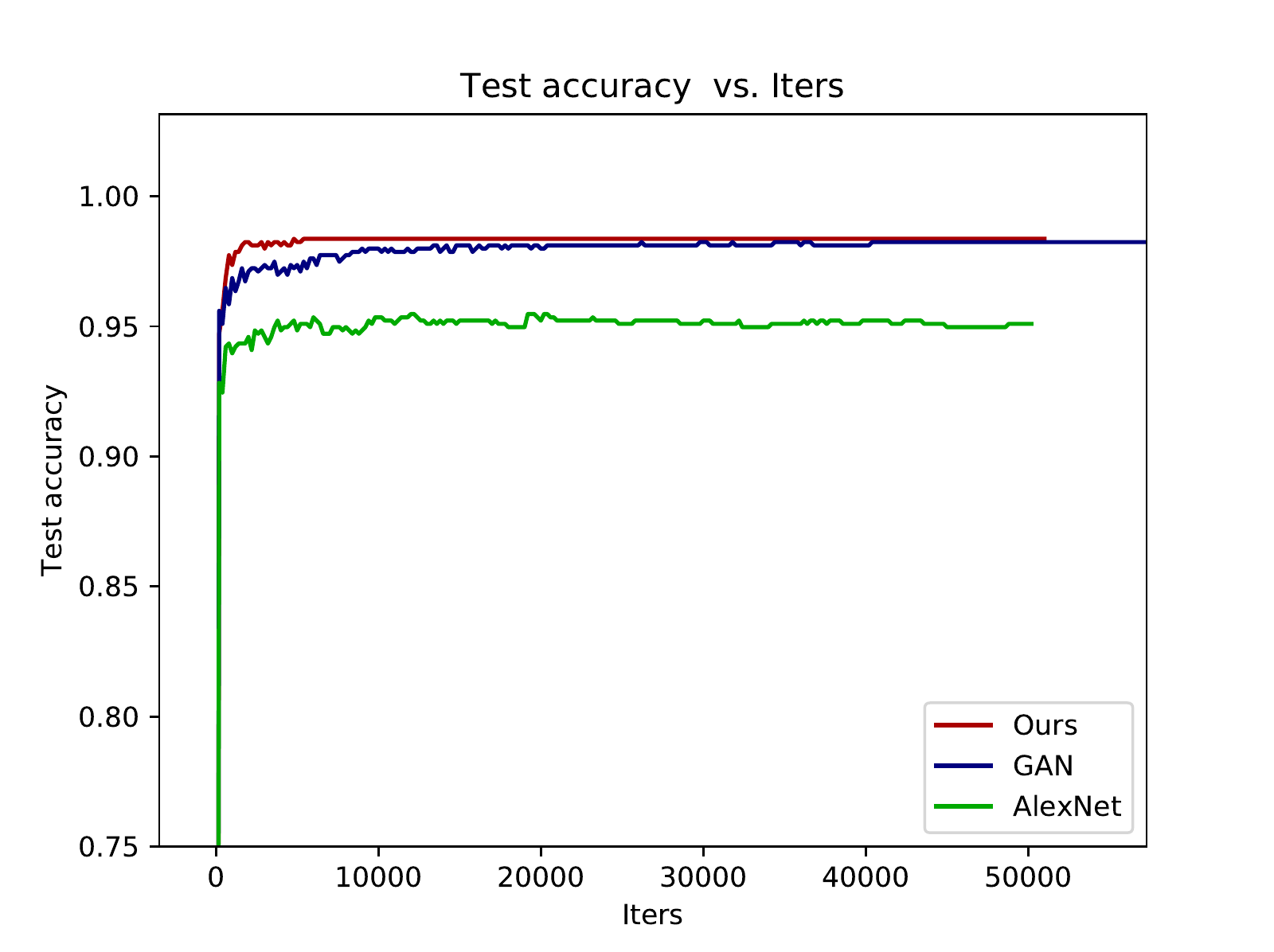}	
		}\\
		\subfigure[$Ar\rightarrow Pr$]{
			\includegraphics[width=0.45\linewidth]{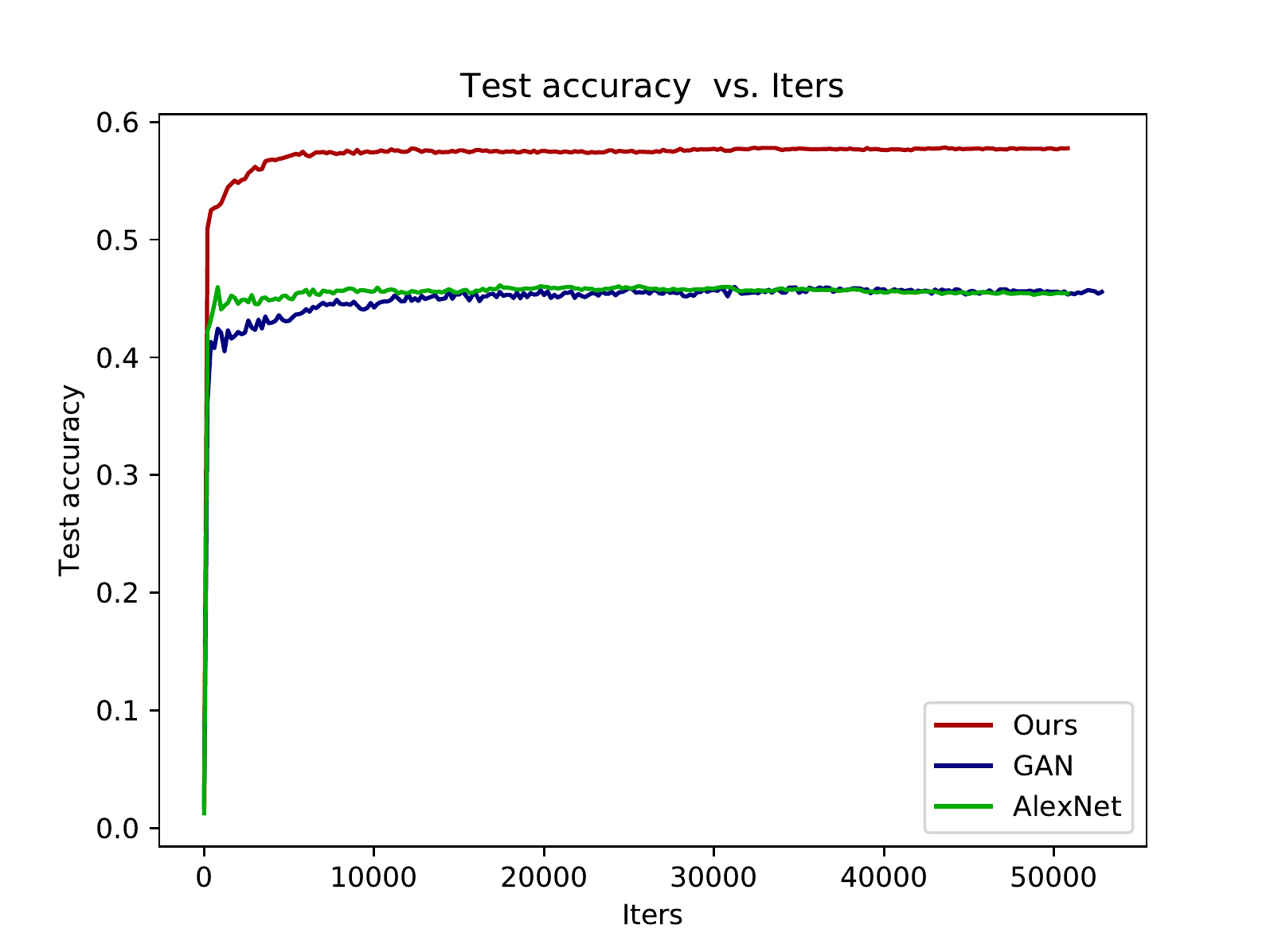}	
		}
		\subfigure[$Pr\rightarrow Rw$]{
			\includegraphics[width=0.45\linewidth]{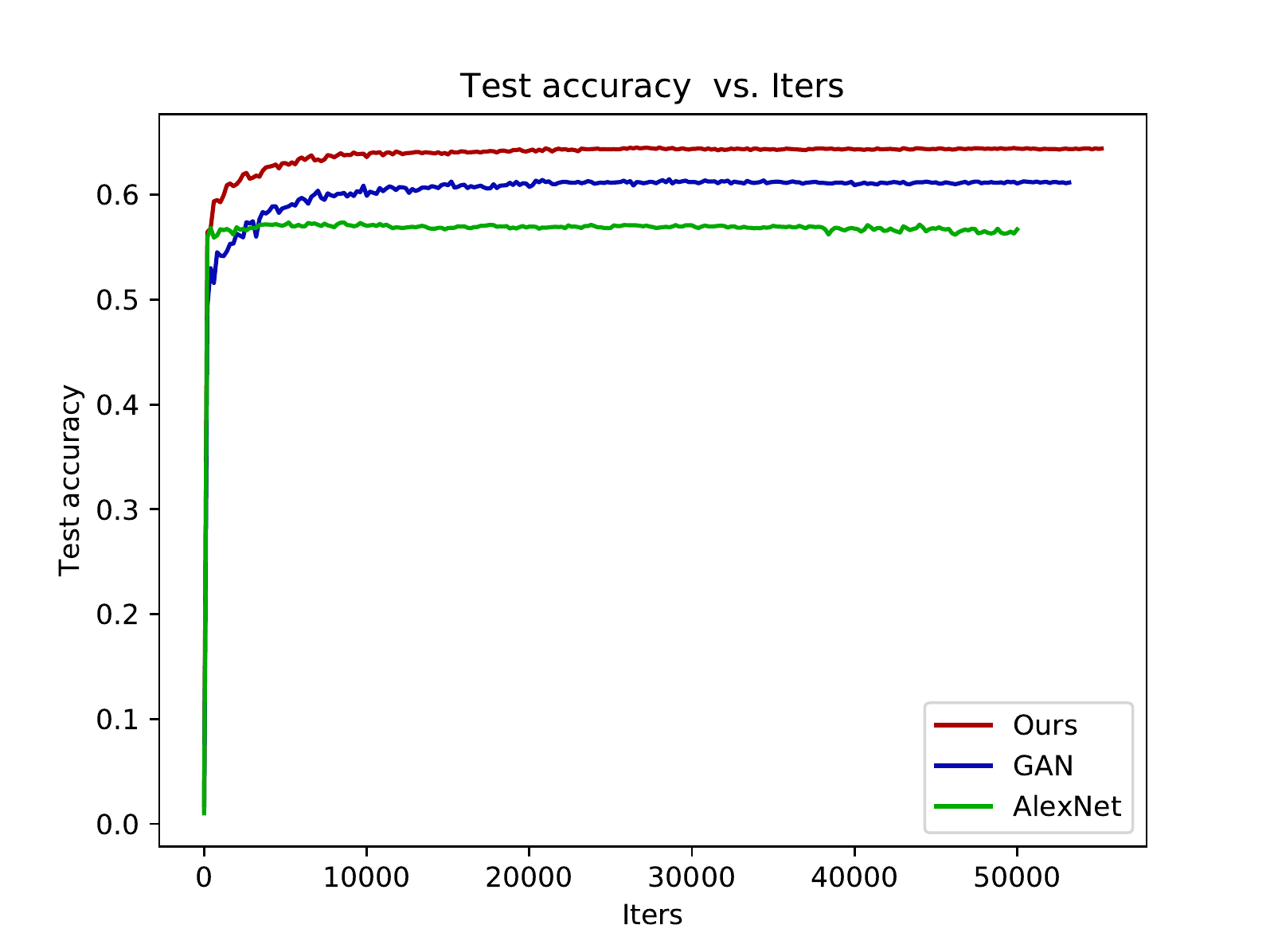}	
		}
	\end{center}
	\caption{(a)\textasciitilde(d) plot the convergence performance of tasks $W\rightarrow A$, $D\rightarrow W$, $Ar\rightarrow Pr$, and $Pr\rightarrow Rw$ respectively. The curves reflect the relationship between the number of iterations and test accuracies. Three models are compared: models finetuned on AlexNet, Domain Adversarial Networks with classifier trained on source data only, and the proposed method, represented by green, blue, and red respectively.}\label{fig:Convergence}
\end{figure}

\begin{figure}[ht]
	\begin{center}
		\subfigure[Parameter Sensitivity]{
			\includegraphics[width=0.45\linewidth]{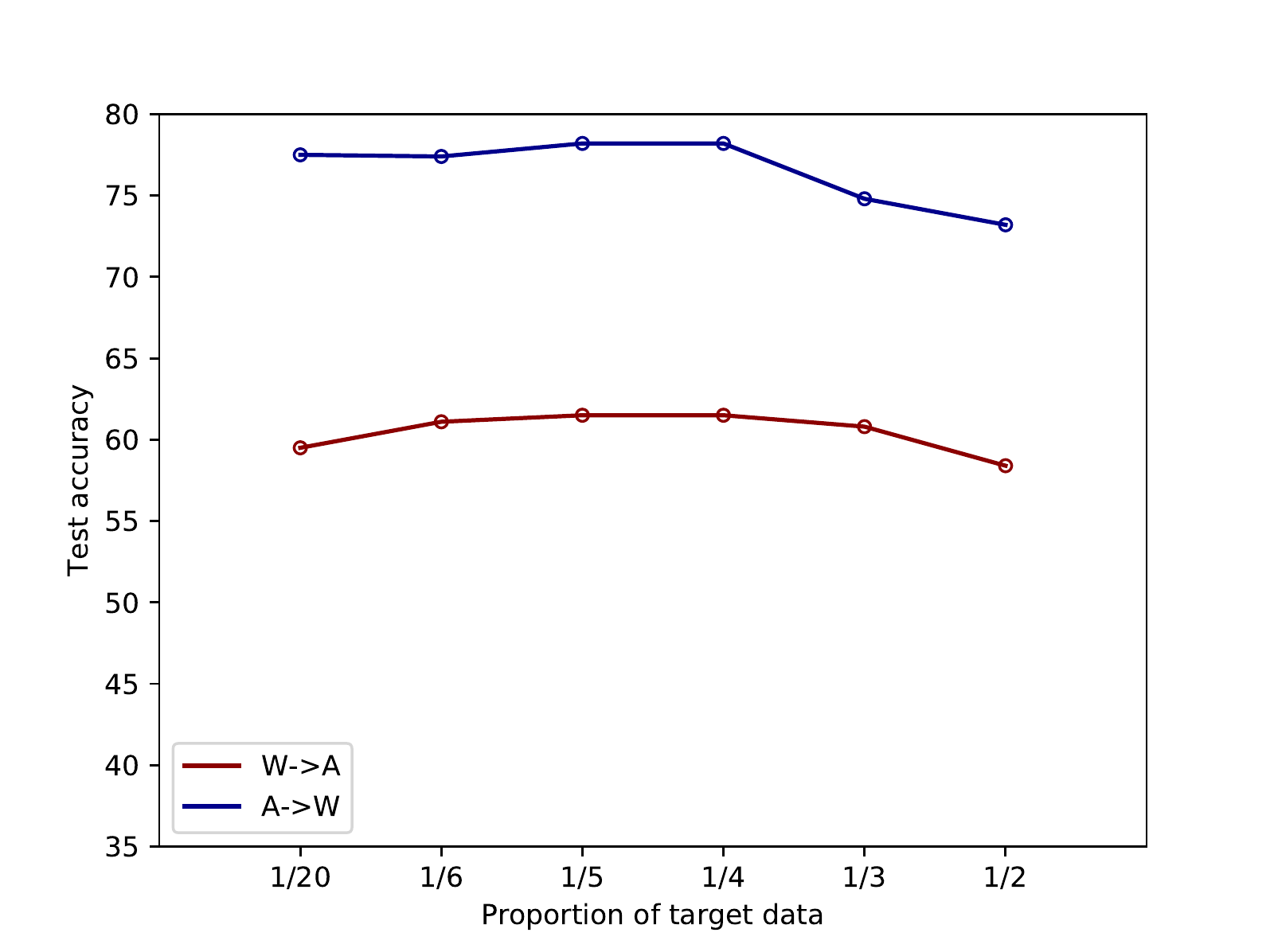}
		}
		\subfigure[Acc. of the selected target data]{
			\includegraphics[width=0.45\linewidth]{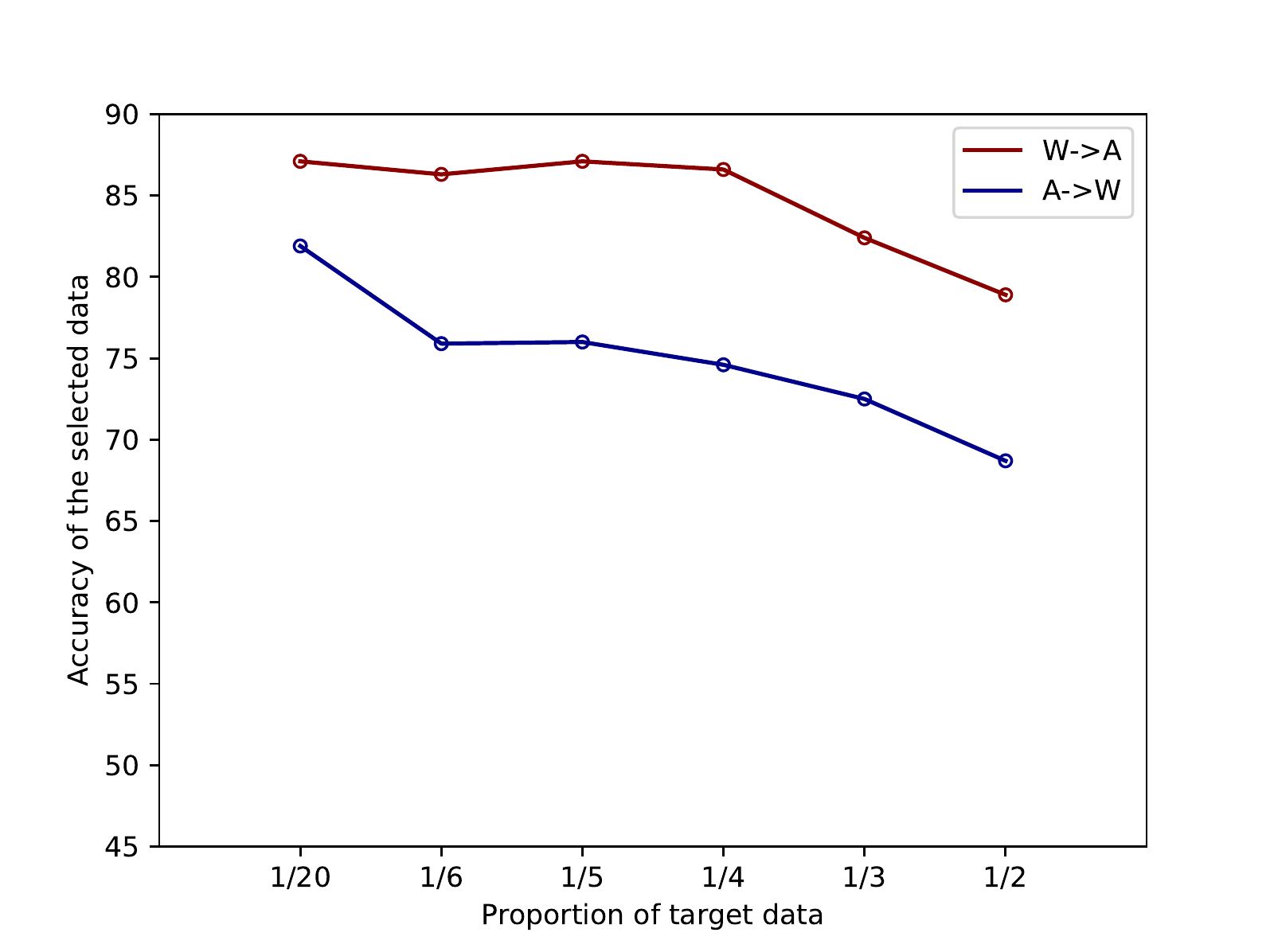}
		}
	\end{center}
	\caption{(a) shows the relationship between the test accuracy and the proportion of selected target data for task $A\rightarrow W$ and task $W\rightarrow A$. (b) shows the relationship between the accuracy of the selected target data and the proportion of selected target data for task $A\rightarrow W$ and task $W\rightarrow A$.}\label{fig:sensi}
\end{figure}

%\begin{figure}[ht]
%	\begin{center}
%		\includegraphics[width=1.0\linewidth]{examples.pdf}
%	\end{center}
%	\caption{Examples of source data, selected target data , and the remaining target data. The first row contains examples of task $A\rightarrow W$. The middle and the last rows are examples of task $Cl \rightarrow Ar$. ``Selected target data'' are those target data considered more likely to be correctly classified and integrated to the source domain.}\label{fig:example}
%\end{figure}

\begin{figure}[ht]
	\begin{center}
		\subfigure[]{
		    \includegraphics[width=0.45\linewidth]{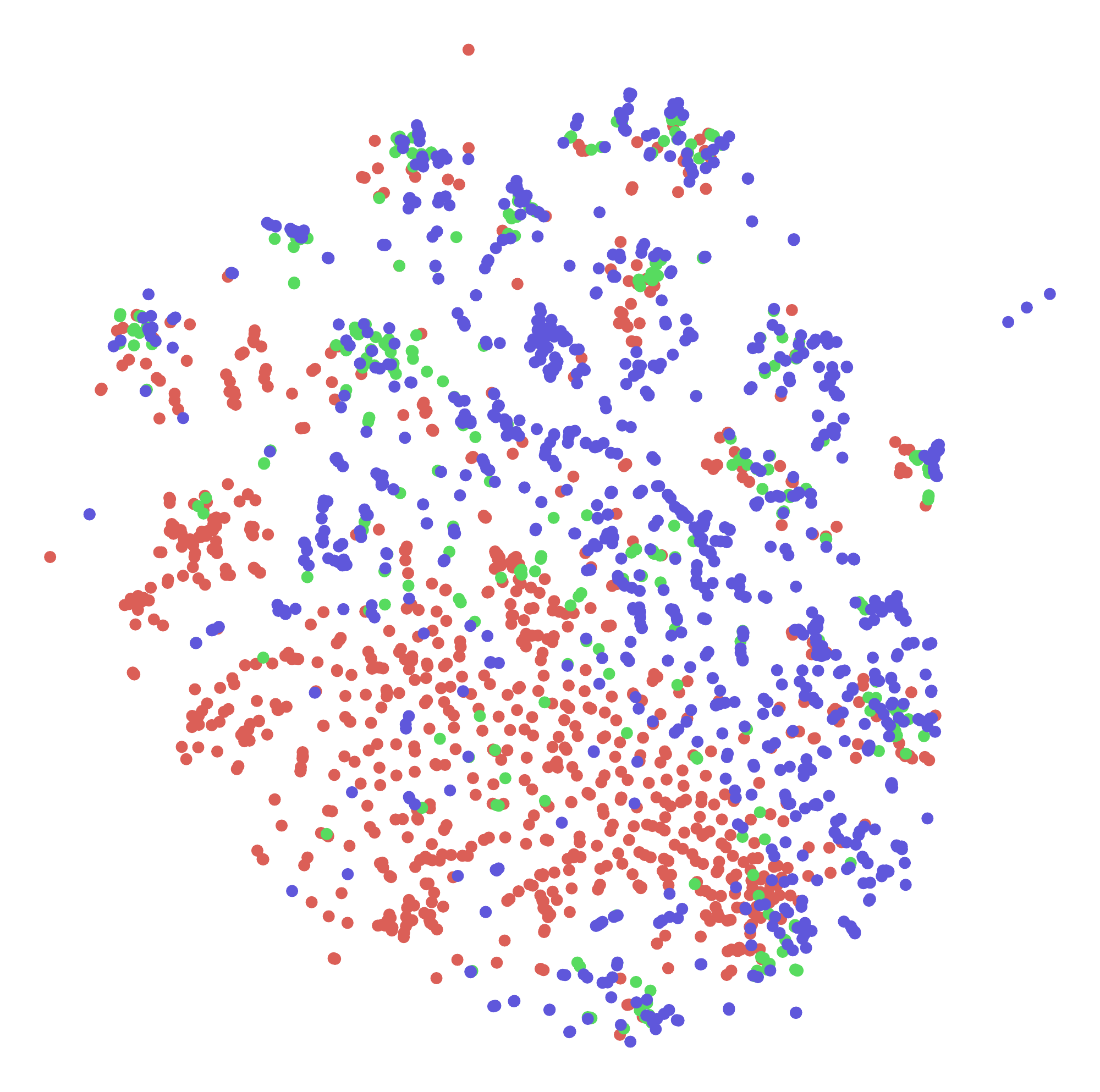}
	    }
        \subfigure[]{
    	    \includegraphics[width=0.45\linewidth]{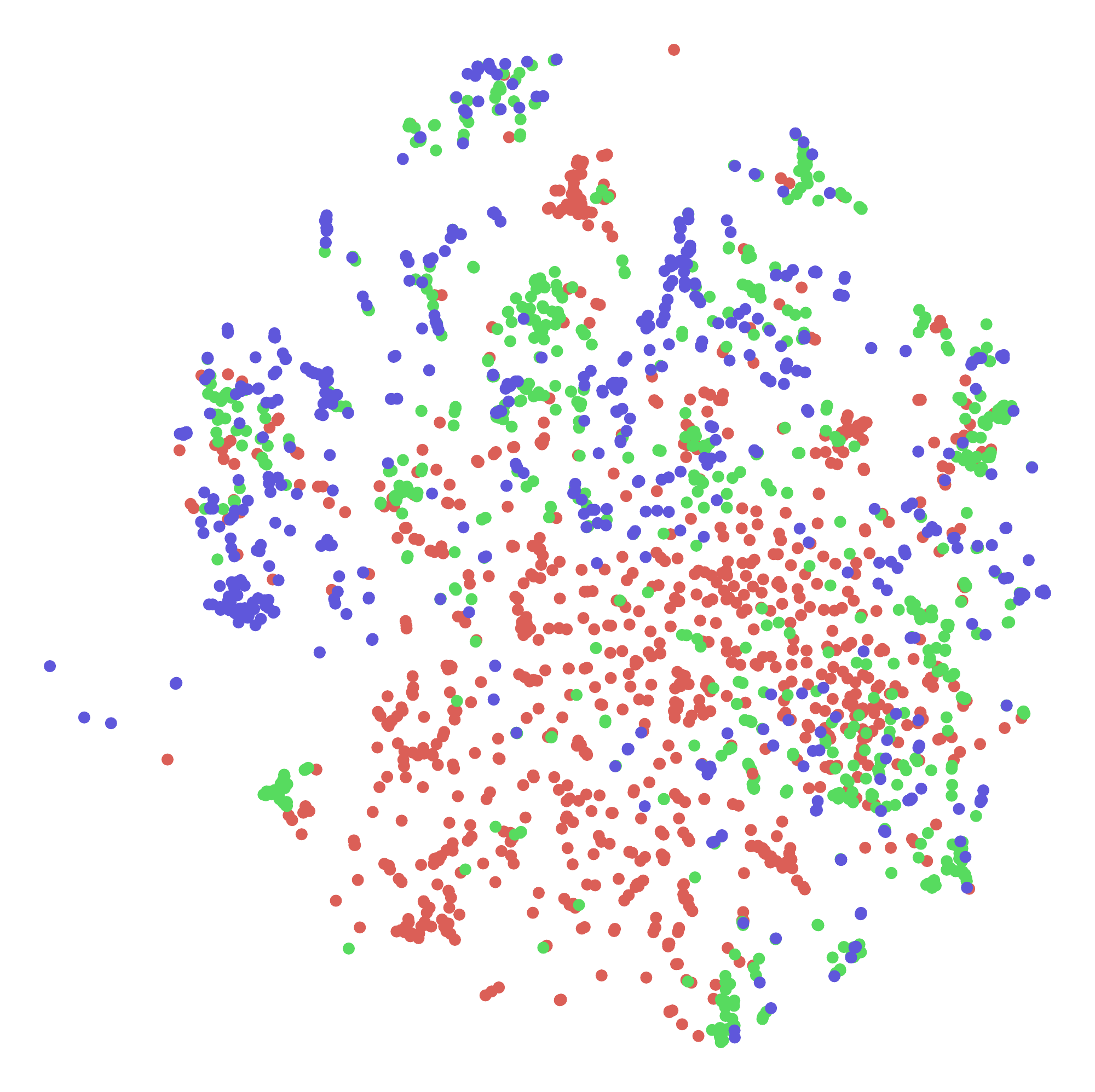}
       }
	\end{center}
	\caption{t-SNE embeddings of features of source data, selected target data $T_a$, and the remaining target data of task $Ar \rightarrow Pr$. The proportions of $T_a$ are $1/4$ and $2/3$ of all the target data in (a) and (b) respectively. Features of the first 20 classes are plotted in this figure. The red points represent the source data, the green points denote $T_a$, and the purple points are the remaining target data. }\label{fig:compare_tsne}
\end{figure}

\subsection{Results}
We compare our results with several existing methods: Transfer Component Analysis (TCA)~\cite{pan2011domain}, Geodesic Flow Kernel (GFK)~\cite{gong2012geodesic}, Deep Adaptation Network (DAN)~\cite{long2015learning}, Residual Transfer Network (RTN)~\cite{long2016deep}, Deep Domain Confusion (DDC)~\cite{tzeng2014deep}, Joint Adaptation Network (JAN)~\cite{long2018conditional}, Adversarial Discriminative Domain Adaptation (ADDA)~\cite{tzeng2017adversarial}, Domain-Adversarial Training of Neural Networks (DANN)~\cite{ganin2016domain}, Multi-layer Joint Kernelized Distance (MJKD)~\cite{mao2018deep}, and Conditional Domain Adversarial Network (CDAN)~\cite{long2018conditional}. The experimental results obtained by models trained using AlexNet~\cite{krizhevsky2012imagenet} architecture are also given as the baseline.

TCA~\cite{pan2011domain} tries to extract transferable components across domains after representing features in a Reproducing Kernel Hilbert Space (RKHS) using Maximum Mean Discrepancy (MMD). GFK~\cite{gong2012geodesic} integrates an infinite number of subspaces that lie on the geodesic flow between the source and target. DAN~\cite{long2015learning} minimizes the maximum distribution discrepancy (MMD) of features from several layers of the deep neural networks after embedding these features in a Reproducing Kernel Hilbert Space (RKHS). As an extension of DAN, JAN~\cite{long2016deep} matches the joint distributions of the deep features and the probability vectors by using their tensor product when computing Joint MMD. RTN~\cite{he2016deep} utilizes residual functions to learn transferable features and the adaptive classifiers jointly. ADDA~\cite{tzeng2017adversarial} is proposed based on adversarial learning by combining discriminative modeling, untied weight sharing, and a GAN loss. Domain-Adversarial DANN~\cite{ganin2016domain} is accomplished through standard back-propagation training by making use of domain adversarial learning~\cite{goodfellow2014generative} by sending features from a single layer of deep networks as the input of a domain discriminator, which maps the source and target features closer by making them indistinguishable. MJKD~\cite{mao2018deep} is a self-training method that selects target most likely to be correctly classified, then add the chosen data to the training set iteratively. CDAN~\cite{long2018conditional} uses the tensor product of deep features and the softmax probability vectors as the inputs of the discriminator and uses a weighted sigmoid loss in the discriminator. Here, we follow an unsupervised domain adaptation protocol, i.e., all labeled source domain samples are used, while labels of all target examples are not provided~\cite{saenko2010adapting}~\cite{ganin2016domain}. The experimental results reported in the paper are the average results after implementing each task 3 times.

The results obtained for the Office-31 dataset are summarized in Table~\ref{tab:fc7_com}. The best results are highlighted in bold and the second best results are underlined. We can observe that JAN which jointly reduces the domain discrepancy by considering the conditional distribution performs better than DAN which only relies on correcting the marginal distribution shifts. Similarly, CDAN which uses the tensor products of features and probabilities as the inputs of the discriminator works better than DANN which only considers the deep features. This reveals that it is important to make use of the classification information in adversarial domain adaptation. The proposed model firstly selects correctly classified target data and then add them to the training set. This can help to make more target data correctly classified during the classification process and can also help to draw more target data close to their correct categories in the adversarial training process, though there may have mis-labeled target data added to the training set. From Table~\ref{tab:fc7_com}, the experimental results show that the proposed method achieves better (e.g., in $W\rightarrow A$, $D\rightarrow A$, $D\rightarrow W$) or comparable accuracies (e.g., in $W\rightarrow D$, $A\rightarrow W$, $A\rightarrow D$) to the previous state-of-the-art methods in each transfer task.  As shown in the last column of Table~\ref{tab:fc7_com}, the average accuracy of the proposed model in this dataset has outperformed all the benchmarking models which are representative state-of-the-art approaches to domain adaptation. 

Results of the Office-Home dataset are recorded in Table~\ref{tab:clef}. Here, several most recently proposed state-of-the-art methods were selected to compare with the proposed one. One might notice that the performance of AlexNet is pretty inferior than the deep domain adaptation approaches, which reflects the difficulty and complexity of Office-Home dataset. For this more challenging domain adaptation dataset which has more categories in each domain and larger discrepancy between different domains, the proposed method is still very distinctive in the domain adaptation performance. Out of the 12 domain adaptation tasks, the proposed one attained the best accuracy in 10. For the remaining two tasks, i.e., $Cl\rightarrow Ar$ and $Pr\rightarrow Cl$, the performance of the proposed method is also pretty comparable, attaining the second best performance. In fact, the performance of the proposed method in this more complicated dataset is even more distinguished than that in the Office-31 dataset.    

In addition, to verify that the proposed method is effective when applied to other framework besides CDAN, we valuated it on DANN. The results of task A$\rightarrow$W, D$\rightarrow$A, Rw$\rightarrow$Pr, and Cl$\rightarrow$Ar are displayed in Table~\ref{tab:DANN}. It shows that based on DANN, which only uses the deep features as the inputs of the discriminator instead of the tensor products, the proposed method can still achieve better performance compared to the original DANN model.

\begin{table}
	\setlength{\belowcaptionskip}{0.5cm}
	\centering
	\caption{Comparison of accuracy based on the architecture of DANN.~\cite{long2018conditional}}
	\small
	\begin{tabular}{p{1.8cm}<{\centering}|p{0.8cm}<{\centering}p{0.8cm}<{\centering}p{0.8cm}<{\centering}p{0.8cm}<{\centering}p{0.8cm}<{\centering}}
		\hlinew{1.5pt}
		Method & A$\rightarrow$W & D$\rightarrow$A & Rw$\rightarrow$Pr & Cl$\rightarrow$Ar & Avg. \\
		\hline
		DANN~\cite{ganin2016domain} & 73.0 & 53.4 & 65.9 & 35.2 & 56.9 \\
		\hline
		Ours & \textbf{78.0} & \textbf{58.6} & \textbf{69.4} & \textbf{35.2} & \textbf{60.3} \\
		\hlinew{1.5pt}
	\end{tabular}
	\label{tab:DANN}
\end{table}

\subsection{Qualitative Analysis}
\subsubsection{t-SNE Embedding} To illustrate how the proposed approach makes the network more discriminative in target domain by mapping target and source data closer, we plot the t-SNE embeddings of deep features extracted by the initial model pre-trained on the source domain and the model obtained from the proposed method in Figure~\ref{fig:t-SNE}. Here, we select 2 tasks from each dataset and plot their deep features: $A\rightarrow D$, $W\rightarrow A$, $Ar\rightarrow Pr$, and $Pr\rightarrow Rw$. The left 2 columns in Figure~\ref{fig:t-SNE} show the source and target deep features extracted by the initial model trained on AlexNet architecture respectively. The right 2 columns show the features extracted by the model learned by the proposed method. We plot the features from ten categories for each domain. Each class is marked with a number as shown in the figures. We can observe that features from the same category become more compact after applying the proposed method. Features from different categories can be discriminated better, and the distribution of each category of source and target data become more coincident.

In addition, in order to visually show what kind of target data are selected to help the domain adaptation, we plot the t-SNE embeddings of the features of the source data, the features of the selected target data $T_a$, and the features of the remaining target data for task $Ar\rightarrow Pr$. Features of the first twenty classes are plotted in Figure~\ref{fig:compare_tsne}. The proportions of $T_a$ are set as $1/4$ and $2/3$ of all the target data in the Figure~\ref{fig:compare_tsne} (a) and Figure~\ref{fig:compare_tsne} (b)  respectively. From the Figure~\ref{fig:compare_tsne} (a), we can observe that the selected target data which is plotted in green, distributed at the edge of the source data (in red) and the remaining target data (in purple). Hence, the selected target data $T_a$ can act as a bridge which helps to map source data and the remaining target data close. Therefore the adaptation performance can be enhanced. From Figure~\ref{fig:compare_tsne} (b), we can observe that the distribution of $T_a$ gets dispersed as the proportion of $T_a$ becomes large (e.g., $2/3$). 

%\subsubsection{Selected Examples}
%To make the results of the proposed method more vividly, some selected target data are displayed in Figure~\ref{fig:example}. In Figure~\ref{fig:example}, examples of task $A \rightarrow W$ and $Cl \rightarrow Ar$ are given. ``Selected target data'' are those target data considered more likely to be correctly classified and integrated to the source domain. We can see that the selected target data are close to the source domain. At the same time, the selected target data are closer to the remaining target data compared with the source data.

\subsubsection{Convergence}
Here, the convergence performances of AlexNet~\cite{krizhevsky2012imagenet}, the original conditioned adversarial domain adaptation~\cite{long2018conditional}, and the proposed method are compared. The convergence curves represent the relationships between the number of iterations and test accuracies are plotted in Figure~\ref{fig:Convergence}. Four tasks are reported here: $W\rightarrow A$, $D\rightarrow W$, $Ar\rightarrow Pr$, and $Pr\rightarrow Rw$. $W\rightarrow A$ and $D\rightarrow W$ are from the Office-31 dataset while $Ar\rightarrow Pr$ and $Pr\rightarrow Rw$ are selected from dataset Office-Home. From Figure~\ref{fig:Convergence}, it can be observed that the proposed method can achieve better results with similar convergence speed.

\subsubsection{Parameter sensitivity}
%\begin{figure}
%	\centering
%	\includegraphics[width=0.7\linewidth]{sensi.pdf}
%\end{figure}
All the experimental results reported here were got under the setting that one fourth of the total number of target data were integrated to the source domain as labeled data. To test the sensitivity of this proportion, we implemented the proposed method by integrating $1/2, 1/3, 1/4, 1/5, 1/6, $ and $1/20$ of the target data respectively for tasks $A\rightarrow W$ and $W\rightarrow A$. The results are plotted in Figure~\ref{fig:sensi} (a). We can observe that when using an appropriate proportion like $1/4, 1/5, 1/6$, the test accuracies are pretty stable. Then we explored the precision of the selected target data under different proportion settings, i.e., $1/2, 1/3, 1/4, 1/5, 1/6, $ and $1/20$. The results are plotted in Figure~\ref{fig:sensi} (b). From Figure~\ref{fig:sensi} (b) one can infer that the precision of the selected target data are much higher than that of all target data. Also, we can observe that with the proportion of the selected data increases, the precisions of the selected data tend to decrease.

\section{Conclusion}
In this paper, we present a new domain adaptation approach which makes use of deep neural networks and the adversarial architecture. Different from previous works, the proposed method makes use of a multi-layer joint kernelized distance to select target data which are more likely to be correctly classified. Then by integrating the source data with the selected target data, the performance of the adversarial domain adaptation performance can be further improved. Analysis is given to show how such an approach can enhance the accuracy of predicting the target labels. Experimental results show that the proposed method can outperform the state-of-the-art methods compared in our experiments. This work can show the value of information provided by the target data, which can provide a new direction of thinking about the domain adaptation problem in the future.
%\section*{Acknowledgement}
%This work was supported by UGC GRF Project No. PolyU 152039/14E.
\section*{Acknowledgement}
This work was supported by GRF, UGC under projects PolyU 152039/14E and PolyU 152228/15E, and PolyU, UGC under project PolyU 152071/17E.

\bibliographystyle{unsrt}
\bibliography{DomainAdp}
\end{document}